\documentclass[preprint,12pt]{elsarticle}

\usepackage{amssymb}
\usepackage{amsmath}
\usepackage{algorithm,algorithmicx,algpseudocode}
\usepackage{subcaption}
\usepackage{booktabs,siunitx}
\usepackage{multirow}
\usepackage{float}
\usepackage{comment}
\usepackage{pifont}
\usepackage{tabularx}   
\usepackage{graphicx}
\usepackage{subcaption}   
\usepackage{caption}
\usepackage{array}
\biboptions{numbers,sort&compress}
\usepackage{lineno}
\usepackage{url}

\newcolumntype{C}{>{\centering\arraybackslash}X} 
\newcolumntype{L}{>{\raggedright\arraybackslash}X}      
\usepackage{makecell}
\newcommand{\cmark}{\checkmark} 

\journal{Nuclear Physics B}

\begin{document}

\begin{frontmatter}



\title{VFM-SDM: A vision foundation model–based framework for training-free, marker-free, and calibration-free structural displacement measurement} 



\author[inst1]{Qingyu Xian\corref{cor1}}
\author[inst2]{Hao Cheng}
\author[inst1]{Berend Jan van der Zwaag} 
\author[inst3]{Rolands Kromanis}
\author[inst1]{Ozlem Durmaz Incel}
\affiliation[inst1]{organization={Pervasive Systems Research Group, Faculty of Electrical Engineering, Mathematics and Computer Science, University of Twente}, 
            city={Enschede},
            country={The Netherlands}
            }
\affiliation[inst2]{organization={Department of Earth Observation Science, Faculty of Geo-Information Science and Earth Observation (ITC), University of Twente},city={Enschede},country={The Netherlands}}
\affiliation[inst3]{organization={Department of Civil Engineering and Management, Faculty of Engineering Technology, University of Twente},city={Enschede},country={The Netherlands}}
\cortext[cor1]{Corresponding author. Email: q.xian@utwente.nl}

\begin{abstract} 
Reliable displacement measurement is fundamental for structural health monitoring and digital engineering workflows, as it provides direct structural response information. 
Vision-based measurement has emerged as a promising approach for low-cost, non-contact displacement monitoring.  
However, its deployment often remains constrained by task-specific model training or on-site preparation, such as marker installation or manual camera calibration.
This study presents a Vision Foundation Model-based framework for Structural Displacement Measurement (VFM-SDM) that integrates VFM-inferred camera parameter estimation and point tracking to reconstruct multi-directional structural displacements via triangulation without task-specific training or on-site preparation, enabling efficient non-contact deployment in real-world applications.
Structural geometry constraints are incorporated to suppress physically implausible deviations and improve estimation consistency. 
A multi-modal field dataset collected from an in-service pedestrian bridge is introduced alongside a unified benchmarking protocol to support reproducible evaluation. 
Representative results show low amplitude errors (NRMSE$_{\text{range}}$: 0.11/0.12), strong temporal agreement (correlation coefficient: 0.86/0.88), and small peak-to-peak amplitude errors (RPPAE: 0.01/0.02) for vertical and lateral displacements, indicating robust performance under real-world conditions. 
The proposed framework advances automated, scalable displacement monitoring and lays the groundwork for VFM-enabled structural response measurements in digital twin and data-centric construction workflows.
\end{abstract}




\end{frontmatter}



\section{Introduction}
\label{sec:intro}

Structural displacement measurements provide direct and interpretable information on structural response and performance, forming a fundamental data source for structural health monitoring and digital engineering workflows \cite{dabous2020condition}. 
Displacement time-series data capture the temporal evolution of structural behavior and enable quantitative analysis of infrastructure systems. 
These data support a range of informatics-driven applications, including structural damage detection and localization \cite{gulgec2019convolutional} and finite element model updating \cite{ereiz2022review}, ultimately enabling data-informed performance assessment and condition evaluation throughout the infrastructure lifecycle \cite{nicoletti2022dynamic}.

Displacement measurement methods are commonly categorized into three groups according to their sensing modality: (1) contact-based approaches \cite{doebling1996damage, moreno2018sensors}, which rely on traditional sensors such as accelerometers and Linear Variable Differential Transformers (LVDTs); (2) non-contact electromagnetic sensing approaches, including laser- and radar-based techniques \cite{garg2019noncontact, xiong2017accurate}; and (3) vision-based approaches \cite{feng2018computer}, which rely on camera-based sensing systems deployed either on fixed monitoring setups or on mobile platforms such as Unmanned Aerial Vehicles (UAVs). 


Contact-based sensors require direct installation on the structure, increasing cost and operational complexity and posing safety risks for in-service infrastructures. Although high-precision sensors such as LVDTs can measure vertical displacement, they typically rely on a stable external reference frame (e.g., ground support), which is often impractical for bridges spanning water or operating under traffic conditions, thereby limiting their field applicability.
Non-contact electromagnetic sensing methods avoid physical installation but can be sensitive to measurement configurations where the sensing direction is not aligned with the principal structural motion. In such cases, reliable estimation of vertical or lateral displacements requires geometric transformation from the measurement direction to the structural coordinate system, which can introduce additional uncertainty and complexity. Their performance further depends on favorable sensing conditions, including clear line-of-sight and minimal electromagnetic interference. In addition, these systems often involve high equipment costs and non-trivial deployment.
As a result, both approaches are generally restricted to sparse measurement locations, limiting their ability to capture spatially distributed structural responses, such as mode shapes and localized deformations.

To address these limitations, vision-based methods \cite{huang2021deep, weng2021homography, bolognini2022vision, li2025advancements, zhang2024automated} have emerged as a compelling alternative. 
Camera-based systems are inherently non-contact, cost-effective, and flexible to deploy, while enabling dense spatial measurements with minimal disruption to structural operation. 
Although vision-based measurements are also affected by imaging geometry, this issue can be explicitly addressed through camera calibration \cite{zhang2002flexible} or camera parameter estimation \cite{wang2023posediffusion}. 
Vision-based displacement measurement involves capturing structural responses using one or multiple cameras and extracting in-plane or three-dimensional (3D) displacements from the recorded image sequences.

A rich body of literature has demonstrated the feasibility of vision-based methods \cite{huang2021deep, weng2021homography, bolognini2022vision, li2025advancements, zhang2024automated} to estimate displacements from multiple views or even a single view.
Many approaches \cite{sun2022vision, marchisotti2023feasibility, zhang2025two} focus on in-plane displacement, typically involving target or point tracking followed by a coordinate transformation from the image pixel coordinate system to the structural coordinate system.
Here, target tracking refers to tracking a predefined physical object or marker \cite{yang2023track}, whereas point tracking aims to track image-level pixels across frames without explicit object semantics \cite{doersch2023tapir}.
However, motion along the normal direction of the infrastructure plane is often non-negligible.
For example, lateral responses of some bridges under traffic loading can become notable and thus represent an important component for displacement monitoring.
Consequently, an increasing number of studies \cite{zhao2022structure, li2025advancements, shao2024out, ruan2025lightweight} have shifted their focus toward the 3D displacement measurement of civil structures, where 3D reconstruction is achieved either through camera parameter estimation or by directly inferring depth information.
With the rapid development of deep learning, numerous methods \cite{sun2022vision, zhang2025two, li2025advancements, shao2024out, shao20243d, shao2025dimmc} incorporate deep learning techniques into their pipelines. 

Nevertheless, three main limitations hinder the wide adoption of vision-based methods in real-world structure monitoring. 
First, many existing approaches \cite{jiao2021displacement, xing2022improving, panigati2025dynamic} rely on artificial markers to ensure robust feature tracking. However, for in-service structures, installing artificial markers is often challenging and may introduce potential safety concerns.
Second, some 3D displacement measurement methods \cite{shao2022target, pan20233d, luo2025structural} require stringent manual camera calibration (e.g., chessboard-based calibration \cite{zhang2002flexible}) to obtain accurate camera parameters and metric scale. This process typically involves capturing multiple images of a calibration target from different viewpoints and estimating camera parameters through optimization, making it labor-intensive and difficult to perform in field monitoring scenarios, particularly under conditions of limited accessibility, large observation distances, and environmental disturbances.
Third, despite the increasing adoption of deep learning–based components in displacement measurement pipelines \cite{sun2022vision, shao20243d, li2025advancements}, 
their practical applicability remains limited by a strong reliance on supervised training or scene-specific fine-tuning.
Such approaches typically assume the availability of labeled data or distributional consistency, which are rarely satisfied at new structural sites.
As a result, models trained on indoor structural models or synthetic simulations often fail to generalize to real-world, in-service infrastructures.
Therefore, these limitations create a gap between laboratory demonstrations and operational monitoring scenarios.

This paper addresses these gaps by introducing a training-free, marker-free, and calibration-free Vision Foundation Model-based Structural Displacement Measurement (VFM-SDM) framework. As a vision-based approach, it enables the measurement of both in-plane and normal-direction displacements in in-service infrastructures such as bridges.
Here, “marker-free” means that natural structural features are directly tracked without installing artificial markers. “Training-free” indicates that the proposed framework relies on inference with a pretrained vision foundation model (VFM) and therefore requires no task-specific model training or fine-tuning. “Calibration-free” means that no on-site manual camera calibration is required, as the necessary camera parameters are implicitly inferred by the VFM, significantly reducing the complexity compared with conventional calibration workflows.

The core idea is to leverage the reasoning capability of a VFM to infer 2D point trajectories and camera parameters from stereo videos, enabling robust 3D displacement reconstruction under real-world conditions. 
Specifically, we adopt VGGT \cite{wang2025vggt}, a VFM pretrained on large-scale and diverse datasets spanning multiple vision tasks, as a core component of the framework. Owing to its strong spatial perception, fine-grained detail awareness, and geometric reasoning capability, VGGT generalizes effectively across domains and environmental conditions without task-specific training or fine-tuning. 
Within the proposed framework, VGGT operates directly on natural structural features to perform stable 2D point tracking and implicit camera parameter estimation \cite{wang2023posediffusion, lin2024relpose++, XIAN2025t-graph}, even under sparse-view conditions.
Finally, the VGGT-inferred 2D point trajectories and camera parameters are combined through stereo triangulation \cite{hartley2003multiple} to reconstruct 3D structural displacement. 

\begin{figure}[t]
  \centering
  \includegraphics[width=1\linewidth]{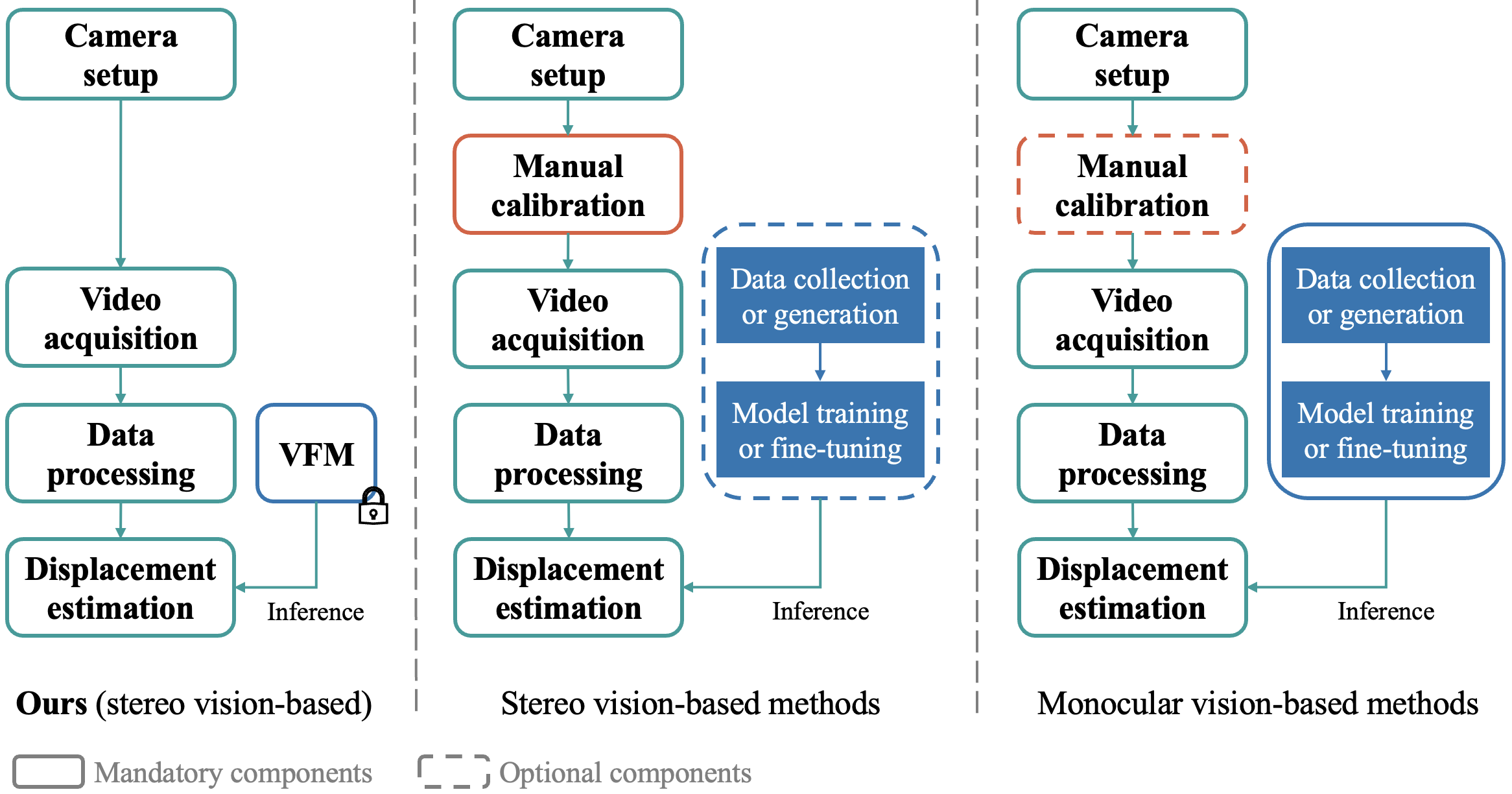}
  \caption{Pipeline comparison of vision-based methods capable of measuring normal-direction structural displacement. Solid boxes denote mandatory components, while dashed boxes indicate optional components. The lock symbol indicates frozen model parameters, meaning that no task-specific training or fine-tuning is required.}
  \label{fig:pipeline-comparison}
\end{figure}

To the best of our knowledge, the proposed framework represents a new pipeline for vision-based structural displacement measurement and constitutes the first attempt to apply VGGT to this task. In this pipeline, the VFM component is not restricted to a specific model. VGGT is adopted as the most suitable VFM currently available, while future advanced vision foundation models can be readily incorporated into the pipeline, further improving the robustness and accuracy of the proposed framework.
Existing vision-based methods capable of measuring normal-direction structural displacement can be broadly categorized into two groups: stereo vision-based methods and monocular vision-based methods. 
A detailed comparison of representative methods is provided in Table~\ref{tab:comparison} in Section~\ref{sec:related}. 
For clearer comparison, Fig.~\ref{fig:pipeline-comparison} illustrates the pipelines of the proposed framework alongside those of the two categories.
Existing stereo vision-based methods typically require manual camera calibration to obtain accurate geometric parameters and may also involve task-specific model training. Monocular vision-based methods typically rely on task-specific model training to estimate depth or scale.
In contrast, the proposed VFM-SDM eliminates both manual camera calibration and task-specific training by exploiting the generalization and spatial reasoning capabilities of the VFM, thereby simplifying the processing pipeline and reducing deployment complexity for in-situ structural monitoring.

To promote physically consistent displacement trajectories, we incorporate a structural geometry refinement (SGR) into the final triangulation stage. This refinement is generally applicable to civil structures where boundary conditions impose strong constraints on specific displacement components. 
Taking bridges as a representative example, longitudinal motion is strongly constrained under normal service conditions, whereas vertical and lateral motions typically dominate the structural response. However, tracking noise caused by defocus or adverse illumination may introduce spurious longitudinal fluctuations after stereo triangulation. 
The proposed SGR enforces structural constraints to suppress such physically implausible deviations, effectively correcting tracking noise-induced artifacts and yielding more physically consistent displacement estimates under challenging imaging conditions.

To facilitate rigorous and reproducible evaluation under realistic operating conditions, we establish a set of field-recorded, multi-modal vibration sequences collected from an in-service steel pedestrian bridge.
To the best of our knowledge, this is the first publicly available multi-modal dataset tailored to real-world structural displacement measurement.
Each sequence consists of synchronized stereo video recordings of the monitored region, reconstructed displacement references, together with associated metadata and a unified evaluation protocol. 
Although modest in scale, the dataset captures representative challenges encountered in field deployments, including defocus, varying illumination, and small-amplitude vibrations, and is specifically tailored for benchmarking structural 3D displacement estimation. 
Evaluation metrics are designed to reflect requirements of the structural displacement measurement and include point-wise displacement value errors, the range-normalized RMSE (NRMSE\textsubscript{range}), the correlation coefficient for assessing temporal consistency between estimated and reference displacement signals, and the relative peak-to-peak amplitude error (RPPAE) for evaluating the ability to recover vibration amplitudes.
Although the proposed framework is theoretically capable of estimating full 3D structural displacements, this study focuses on the vertical and lateral components that are practically observable in real-world bridge environments.

Overall, the proposed framework is designed for practical field applications. The system can be deployed using only two tripods and two cameras placed outside the structure, enabling safe and non-contact displacement monitoring without installing sensors or artificial markers on the structure. The hardware is inexpensive and easy to install, resulting in low equipment and deployment costs. In addition, the data processing follows a unified pipeline that does not require manual calibration procedures or task-specific model training. Field experiments demonstrate that the proposed framework achieves displacement estimation accuracy comparable to accelerometer-based measurements in most tested scenarios. These characteristics make the framework suitable for real-world deployment and efficient structural displacement monitoring, representing a deployment-oriented solution for intelligent structural monitoring.

The major contributions of this paper are summarized as follows:
\begin{enumerate}
\item
We propose the first vision foundation model–based framework (VFM-SDM) for in-situ structural displacement measurement via 3D reconstruction, enabling deployment-oriented monitoring without artificial markers, manual camera calibration, or task-specific training, thereby significantly simplifying real-world deployment.
\item
We introduce a geometry-guided structural refinement that enforces physically plausible displacement constraints during triangulation, correcting tracking noise-induced artifacts and yielding more physically consistent displacement estimates under challenging in-situ imaging conditions.
\item
We establish the first compact, field-recorded multi-modal vibration dataset for in-situ structural displacement measurement, comprising synchronized stereo videos, accelerometer-derived displacement references, and a unified evaluation protocol, providing practical and reproducible benchmarking resources for vision-based structural monitoring.
\item
Experimental results on the field dataset demonstrate accurate and reliable recovery of vertical and lateral displacements, consistent with accelerometer-derived references, while component-level evaluations further validate the effectiveness of the proposed framework.
\end{enumerate}

The remainder of this paper is organized as follows: 
Section~\ref{sec:related} reviews prior work on vision-based in-plane and 3D displacement measurement and summarizes the main approaches to camera parameter estimation.
Section~\ref{sec:method} first introduces the background of vision foundation models (VFMs) and describes the specific VFM adopted in this study, VGGT, then details the proposed VFM-based framework and explains the working mechanism of each component.
Section~\ref{sec:dataset} introduces the data acquisition, dataset overview, and the evaluation protocol. 
Section~\ref{sec:experiments} presents implementation details, quantitative and qualitative results, together with a computational cost analysis and a sensitivity analysis conducted on the constructed field dataset.
Section~\ref{sec:discussion} discusses the reliability of camera parameter estimation using VGGT, limitations, and future directions, and Section~\ref{sec:conclusion} concludes the paper.

\section{Related work}
\label{sec:related}
The related work reviewed in this section focuses primarily on vision-based measurement approaches. Specifically, recent studies are categorized according to the dimensionality of the measured displacement into two groups: (1) in-plane displacement measurement and (2) three-dimensional displacement measurement. In addition, since camera parameter estimation constitutes a core component of the proposed framework, this section also reviews the major technical approaches used for camera parameter estimation.

\subsection{In-plane displacement measurement}
In-plane displacement measurement typically consists of target tracking and a subsequent coordinate transformation from the image pixel coordinate system to the structural coordinate system, which is commonly realized via a homography matrix \cite{hartley2003multiple}.
Several studies have employed artificial markers or engineered patterns to improve measurement precision in vision-based displacement monitoring. 
\citet{jiao2021displacement} proposed a homography-based framework that jointly estimates camera motion and structural response using a RANSAC--ESM hybrid tracker, achieving sub-pixel accuracy under non-stationary camera setups. 
\citet{ji2020vision} developed a monocular target-tracking framework for RC structural tests to estimate planar deformation components and quantify crack patterns from specimen images.
\citet{kromanis2021multiple} developed a camera-independent approach capable of computing structural displacements from videos captured at different viewpoints, obtaining discrepancies of less than 6\% in both laboratory and field tests. 
To further mitigate camera-induced errors, \citet{xing2022improving} introduced a dual-view optical system combining a beam splitter, convex lens, and photomask for drift and vibration compensation, while \citet{marchisotti2023feasibility} extended such setups to UAV-mounted multi-camera systems for vibration measurement and modal analysis.

To enhance robustness and automation, recent studies have trained deep learning models and incorporated them into vision-based displacement measurement frameworks. 
\citet{huang2021deep} introduced a Convolutional Neural Network–Generative Adversarial Network (CNN–GAN) hybrid model for accurate displacement estimation under harsh conditions, and \citet{pan2023vision} coupled YOLO-based detection with Kanade–Lucas–Tomasi (KLT) feature tracking for real-time vibration measurement. 
\citet{jeong2022real} employed a Distractor-Aware Siamese Region Proposal Network (DaSiamRPN) for reference-free tracking under illumination changes and occlusions, while \citet{he2025improving} further improved 2D displacement accuracy by integrating color space fusion and super-resolution reconstruction under varying imaging conditions. 
\citet{weng2021homography} used a neural-network-assisted homography model to correct UAV camera motion, and \citet{sun2022vision} applied a GAN-based super-resolution network to enhance texture details and accuracy in low-resolution surveillance videos. 
In addition, \citet{wan2026improved} proposed an improved SIFT-based method with deep learning-assisted refinement to enhance keypoint detection and matching, enabling accurate target-free 2D displacement estimation.

Parallel research has explored vision-based approaches that directly track natural features without any model training or fine-tuning. 
Early work, such as InnoVision by \citet{luo2018robust}, utilized gradient-based template matching with sub-pixel interpolation for robust tracking under camera vibration. 
Subsequent improvements include distraction-free Siamese tracking \cite{xu2021accurate}, motion magnification \cite{ghandil2021enhanced}, and vision–acceleration fusion via adaptive Kalman filtering \cite{ma2022structural}. 
Multi-camera configurations \cite{wang2022completely} and UAV-based implementations \cite{bolognini2022vision} further extended these techniques to full-scale modal analysis, achieving accuracy comparable to accelerometer measurements.
\citet{zhang2025two} focused on UAV-based measurements, enhancing displacement accuracy through adaptive feature fusion and two-stage motion correction.

However, these methods are inherently limited to in-plane displacement measurement and cannot capture the structural normal-direction response. Consequently, growing research efforts have shifted toward 3D vision-based displacement measurement.

\subsection{3D displacement measurement}
Compared to in-plane displacement measurement, 3D displacement measurement involves additional estimation of spatial geometry. In stereo-vision configurations, this is commonly realized through camera parameter estimation (e.g., chessboard-based calibration \cite{zhang2002flexible}), while in monocular systems it is typically addressed by depth estimation.

To achieve accurate 3D displacement measurement, stereo vision-based methods are a well-established and reliable solution.
Existing stereo vision–based methods can be broadly divided into two categories according to the type of visual features used, namely approaches based on artificial markers and approaches based on natural features.
Some studies introduce artificial markers to facilitate feature tracking. \citet{pan20233d} developed a framework that reconstructs 3D point clouds via Structure-from-Motion (SfM) and multi-view stereo, and integrates a convolutional neural network–based 3D object detection model to identify steel plate structures, enabling millimeter-level quantification of out-of-plane structural displacements. Similarly, \citet{zhang2024automated} proposed automated 2D and 3D vision-based displacement monitoring methods capable of tracking multiple coded targets without manual template selection or repeated calibration, validated through simulations and bridge experiments. To improve large-scale monitoring capability, \citet{cui2025research} proposed a wide-area displacement monitoring system that integrates a rotating platform with binocular vision to expand the field of view. By combining phase-enhanced phase motion estimation with multi-scale depth homography estimation, the method improves robustness to noise and lighting variations for accurate displacement measurement in complex environments. \citet{xie2025new} further proposed a binocular vision–based approach that uses square feature recognition with subpixel enhancement to measure 3D structural displacements and motion trajectories, providing a cost-effective solution for structural health monitoring.
Other studies adopt natural features to enable marker-free 3D displacement measurement. \citet{shao2021computer} introduced a binocular vision system that leverages deep-learning-based keypoint detection and matching to obtain full-field 3D vibration measurements, achieving accuracy comparable to measurements obtained from Linear Variable Differential Transformer (LVDT) and Laser Displacement Sensor (LDS) at lower cost.
\citet{shao2022target} further combined deep-learning-based keypoint tracking with phase-based motion magnification to measure submillimeter 3D vibration displacements in both laboratory and field bridge experiments, showing strong agreement with laser displacement and accelerometer data. 
\citet{wu2025displacement} proposed a stereo vision–based method combining ORB (Oriented FAST and Rotated BRIEF) feature detection and matching with the Lucas–Kanade optical flow tracking method to improve robustness in dynamic environments, enabling multi-point displacement monitoring and modal identification for large-span structures.
\citet{ruan2025lightweight} proposed a lightweight binocular vision framework that programmatically determines camera parameters and employs a hybrid feature point tracking algorithm incorporating unsupervised deep learning to achieve 3D displacement tracking, thereby improving computational efficiency, tracking accuracy, and real-time performance for large-scale structural monitoring. Nevertheless, the framework still relies on the camera's built-in inertial measurement unit to provide pose information for estimating camera parameters.
Overall, stereo vision–based methods generally depend on accurate camera calibration, commonly achieved using chessboard-based calibration procedures \cite{zhang2002flexible} when no auxiliary sensors are available. Although effective in laboratory environments, such calibration procedures are often difficult to implement in field monitoring scenarios.

Some studies have explored monocular vision systems for 3D displacement measurement of civil structures. 
\citet{narazaki2021efficient} introduced a physics-based graphics framework that simulates vision-based 3D displacement measurement to optimize algorithm design and camera configuration, integrating model-informed motion compensation and dense 3D reconstruction, and validating the optimized algorithms on both synthetic and laboratory bridge experiments. 
More recent works employ deep learning with synthetic or laboratory-generated datasets to directly learn structural depth information, thereby avoiding explicit camera calibration. 
\citet{shao2024out} developed a monocular vision system for out-of-plane vibration measurement using large-scale dataset generation and advanced object segmentation to achieve high accuracy and robustness. 
\citet{li2025advancements} proposed a monocular moving-camera framework that combines point tracking for in-plane motion, deep-learning-based depth prediction for out-of-plane motion, and motion compensation through stationary-region registration. 
\citet{luo2025structural} further presented an automatic monocular method that integrates enhanced feature-group tracking with improved depth estimation to enable long-term, accurate 3D displacement measurement by mitigating cumulative and sub-pixel vibration errors.
Several recent studies have innovatively developed learning-based mesh models to bypass camera calibration or feature tracking, directly learning structural pose information or 3D deformation. 
\citet{zhao2022structure} proposed Structure-PoseNet, a monocular vision framework for structures with known geometry that integrates semantic segmentation and pose regression to infer 3D structural poses and dense dynamic displacements from video data, validated on shaking-table experiments and illustrative full-scale cases such as the Tacoma Narrows Bridge.
\citet{shao20243d} introduced a marker-free monocular approach employing a deep neural network to predict 3D structural mesh deformation from a single image, enabling accurate 3D vibration measurement without multiple cameras or key points. 
\citet{shao2025dimmc} further advanced this direction with a ViT-based mesh reconstruction network (e.g., \cite{lin2021end}) and 3D point registration, allowing accurate 3D displacement estimation from moving-camera videos without targets or feature tracking.

Compared with stereo vision–based 3D displacement measurement, monocular systems offer a much simpler workflow. However, most existing methods depend heavily on high-quality training data, often synthetic or collected under controlled laboratory conditions, and on the generalization ability of the deep learning models. Consequently, applying these methods directly to real in-service structures remains challenging.

Vision foundation models (VFMs) have recently emerged as powerful models that learn general-purpose visual representations, offering strong robustness and generalization across diverse tasks.
A growing body of work \cite{li2025advancements, shao2024out, shao20243d, shao2025dimmc} has incorporated VFMs, such as Segment Anything (SAM) \cite{kirillov2023segment} and Track Anything (TAM) \cite{yang2023track}, to enhance robustness and accuracy. However, these VFMs are typically limited to a single output (e.g., segmentation or tracking) and are mainly used as pre-processing tools to isolate structural regions for subsequent tasks, such as training or fine-tuning depth estimation networks.
By contrast, our proposed framework adopts Visual Geometry Grounded Transformer (VGGT) \cite{wang2025vggt}, a 3D perception VFM capable of jointly inferring multiple tasks. We directly utilize the VGGT-predicted camera parameters and point tracking trajectories for displacement reconstruction, which substantially simplifies the overall pipeline and eliminates the need for any task-specific model training or fine-tuning.

\subsection{Camera parameter estimation}
In stereo vision systems, camera parameter estimation involves determining the intrinsic parameters (e.g., focal lengths and principal point) and the extrinsic parameters that relate world coordinates to the camera coordinate frame. 
Existing approaches can be broadly grouped into traditional calibration-based methods and learning-based estimation techniques, which are briefly reviewed below.

\vspace{4pt}
\paragraph{Traditional calibration methods}
Classical camera calibration approaches can also be divided into two categories: target-based calibration and self-calibration (Structure-from-Motion, SfM \cite{sturm2012benchmark, schonberger2016structure}).
Target-based calibration employs known 2D or 3D reference patterns such as checkerboards or AprilTags, placed at various orientations and distances. 
By establishing precise correspondences between detected image corners and their known world coordinates, one can directly solve for the intrinsic parameters and distortion coefficients through linear initialization, as in Zhang’s method \cite{zhang2002flexible}, followed by nonlinear optimization of the reprojection error.
The extrinsic parameters for each view are then obtained as the rigid transformations aligning the observed pattern to the camera frame. 
This approach offers excellent metric accuracy and repeatability, but requires a controlled setup, sufficient illumination, and manual placement of calibration targets, which are often impractical for in-service field deployments.

Self-calibration, by contrast, dispenses with explicit calibration targets and instead estimates camera parameters jointly with 3D structure directly from natural scene imagery. It infers epipolar geometry from point correspondences across multiple views and recovers both camera extrinsics and intrinsics through projective constraints \cite{schonberger2016structure}.
Although self-calibration eliminates the need for artificial patterns, its performance is highly sensitive to the spatial distribution and number of tracked points, the amount of parallax, and the diversity of camera pose. Even when bundle adjustment \cite{triggs1999bundle} is included in the optimization pipeline, the estimation may remain unstable and can fail to converge under sparse or near-degenerate configurations (e.g., small baseline, low texture, or weak pose diversity).

\vspace{4pt}
\paragraph{Learning-based estimation}
Recently, deep learning has enabled a fundamentally different class of learning-based camera parameter estimation methods. 
Unlike traditional approaches that depend on sparse feature correspondences and local geometric constraints, learning-based estimators perceive the entire image content and infer camera parameters by reasoning about the global 3D spatial geometry implicitly encoded in the visual cues. 
Trained on large-scale multi-view datasets, these models \cite{wang2023posediffusion, lin2024relpose++, XIAN2025t-graph, wang2025vggt} learn strong priors over scene layout, camera optics, and even depth distribution, allowing them to generalize to unseen scenes and handle difficult conditions where classical methods fail. 
Once trained, they can directly predict camera parameters from one or a few frames, without explicit feature matching or iterative geometric reconstruction, making them extremely efficient and convenient in real-world applications. 
Although the metric accuracy may be lower than that of traditional calibration under dense viewpoints, their robustness under weak geometry or sparse viewpoints makes them highly attractive for field deployment.

In summary, compared with traditional calibration methods, learning-based estimation methods offer the most balanced compromise between operational convenience and metric accuracy in real infrastructure environments.

\begin{table}[t]
\centering
\fontsize{8pt}{8pt}\selectfont
\caption{Comparison of representative vision-based displacement measurement methods from the cited literature.
“\cmark” indicates that the corresponding component or condition is employed in the method, “—” indicates it is not employed or not reported in the original publication.}
\label{tab:comparison}
\resizebox{\columnwidth}{!}{
\begin{tabular}{@{}lcccccccc@{}}
\toprule
\multirow{2}{*}{\textbf{Method}} &
\multirow{2}{*}{\makecell{\textbf{Normal-}\\\textbf{direction}}} &
\multicolumn{2}{c}{\textbf{Number of Views}} &
\multicolumn{2}{c}{\textbf{Tracked Feature}} &
\multirow{2}{*}{\makecell{\textbf{Field} \\ \textbf{Validation}}} &
\multirow{2}{*}{\makecell{\textbf{Manual} \\ \textbf{Calibration}}} &
\multirow{2}{*}{\makecell{\textbf{Model Training}/ \\ \textbf{Fine-tuning}}} \\
\cmidrule(lr){3-4} \cmidrule(lr){5-6}
& & \textbf{Stereo} & \textbf{Monocular} & \textbf{Artificial} & \textbf{Natural} & & & \\
\midrule

\cite{huang2021deep} & — &  & \cmark & \cmark &  & \cmark & \cmark & \cmark \\
\cite{jeong2022real, weng2021homography, he2025improving} & — &  & \cmark &  & \cmark & \cmark & — & \cmark \\
\cite{pan2023vision} & — &  & \cmark & \cmark &  & — & — & \cmark \\
\cite{bolognini2022vision} & — &  & \cmark &  & \cmark & — & — & — \\
\cite{kromanis2021multiple, panigati2025dynamic} & — &  & \cmark & \cmark &  & \cmark & — & — \\
\cite{jiao2021displacement, xing2022improving, marchisotti2023feasibility} & — &  & \cmark & \cmark &  & — & — & — \\
\makecell[l]{\cite{ma2022structural, xu2021accurate, ghandil2021enhanced},\\
\cite{wang2022completely, luo2018robust, zhang2025two}} & — &  & \cmark &  & \cmark & \cmark & — & — \\
\cite{sun2022vision, wan2026improved} & — &  & \cmark &  & \cmark & — & — & \cmark \\

\cite{narazaki2021efficient} & \cmark &  & \cmark &  & \cmark & — & \makecell{— (from physical\\-based model)} & — \\
\cite{li2025advancements, shao2024out} & \cmark &  & \cmark &  & \cmark & — & — & \cmark \\
\cite{luo2025structural} & \cmark &  & \cmark &  & \cmark & — & \cmark & \cmark \\
\cite{shao20243d, shao2025dimmc} & \cmark &  & \cmark & \multicolumn{2}{c}{—} & — & — & \cmark \\
\cite{zhao2022structure} & \cmark &  & \cmark & \multicolumn{2}{c}{—} & — & \cmark & \cmark \\
\cite{shao2022target} & \cmark & \cmark &  &  & \cmark & \cmark & \cmark & \cmark \\
\cite{pan20233d} & \cmark & \cmark &  & \cmark &  & — & \cmark & \cmark \\
\cite{cui2025research} & \cmark & \cmark &  & \cmark &  & \cmark & \cmark & \cmark \\
\cite{zhang2024automated, xie2025new} & \cmark & \cmark &  & \cmark &  & — & \cmark & — \\
\cite{shao2021computer} & \cmark & \cmark &  &  & \cmark & — & \cmark & \cmark \\
\cite{wu2025displacement} & \cmark & \cmark &  &  & \cmark & — & \cmark & — \\
\cite{ruan2025lightweight} & \cmark & \cmark &  &  & \cmark & — & — (from IMU) & — \\

\midrule
\textbf{Ours} & \cmark & \cmark &  &  & \cmark & \cmark & — & — \\
\bottomrule
\end{tabular}
}
\end{table}

The works discussed above span a wide range of vision-based displacement measurement methods, differing in their sensing configurations, feature extraction strategies, and practical deployment requirements.
In Table~\ref{tab:comparison}, several representative existing methods from recent years are compared across six key aspects: \emph{Normal-direction}, \emph{Number of Views}, \emph{Tracked Feature}, \emph{Field Validation}, \emph{Manual Calibration}, and \emph{Model Training/Fine-tuning}.
It is worth noting that the \emph{Number of Views} refers specifically to the number of camera viewpoints required for performing the displacement measurement task itself, rather than the total number of cameras used in the overall system design.
In some works, multiple cameras are employed for other purposes, such as achieving full-scale structural monitoring.
Regarding \emph{manual Calibration}, this criterion mainly concerns 3D displacement measurement, where most methods rely on manual camera calibration to obtain the intrinsic and extrinsic camera parameters.
In contrast, for in-plane displacement measurement, manual camera calibration is generally unnecessary.
These methods typically establish the mapping between the image plane and the structural plane using a homography matrix or a predefined scale factor, which is relatively simpler to implement.

As shown in Table~\ref{tab:comparison}, compared with existing methods, the proposed framework provides a more concise pipeline for structural displacement measurement by eliminating the need for artificial markers, manual camera calibration, and task-specific model training. This significantly simplifies the workflow and facilitates safer and more practical field deployment. The effectiveness of the framework is validated through field experiments (see Sec.~\ref{sec:experiments}).

\section{Methodology}
\label{sec:method}
This section first introduces Vision Foundation Models (VFMs), with a particular focus on the VGGT model used in this work, providing the necessary background to contextualize the proposed framework.
It then presents the overall VFM-based framework and explains how the VFM is systematically integrated into the structural displacement measurement pipeline. Subsequently, each major component is described in detail, including VFM-based marker-free 2D point tracking, VFM-based sparse-view camera parameter estimation, stereo triangulation, and structural geometry refinement.

\subsection{Preliminaries}
\label{subsec:pre}
VFMs represent a new paradigm in computer vision, characterized by large-scale pretraining on diverse and heterogeneous visual data to learn general and transferable representations. 
Unlike conventional task-specific models that require retraining for each application, VFMs are designed to serve as universal visual backbones, capable of zero-shot or few-shot adaptation, meaning that they can be applied to new tasks without retraining or with only minimal additional supervision, across a wide range of downstream tasks such as detection, segmentation, depth estimation, and 3D reconstruction.
They typically employ transformer-based architectures and self-supervised or multimodal pretraining objectives to capture rich semantic, geometric, and contextual information. 
Representative VFMs include Depth Anything \cite{yang2024depth}, Segment Anything (SAM) \cite{kirillov2023segment}, Track Anything (TAM) \citet{yang2023track} and Visual Geometry Grounded transformer (VGGT) \cite{wang2025vggt} which have demonstrated remarkable generalization across domains and data modalities. 

VGGT is a transformer-based 3D perception vision foundation model that unifies multiple geometric estimation tasks within a single feed-forward framework. 
Pretrained on large-scale and multi-view video datasets, VGGT learns a geometry-aware representation that enables consistent 3D scene perception across single-image, multi-view, and dynamic video inputs without task-specific fine-tuning. 
Serving as a general-purpose 3D backbone, VGGT simultaneously predicts multiple geometric quantities, providing reliable geometric and global semantic priors for downstream applications such as Novel View Synthesis, and 3D reconstruction.

\begin{figure}[t]
  \centering
  \includegraphics[width=1\linewidth]{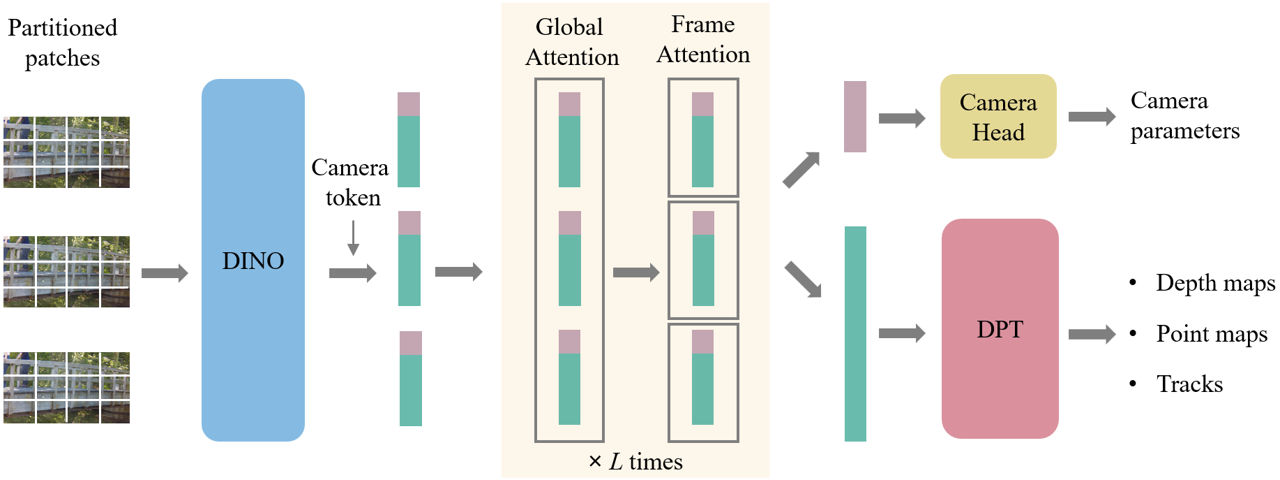}
  \caption{Architecture overview of VGGT.}
  \label{fig:vggt}
\end{figure}

As shown in Fig.~\ref{fig:vggt}, the VGGT network accepts one or multiple RGB frames as input, which are first patchified and embedded into visual tokens using a pretrained DINO \cite{oquab2023dinov2} encoder.
A learnable camera token is then appended to each frame sequence to facilitate camera parameter regression. 
The resulting tokens are processed through a stack of alternating global and frame-wise self-attention blocks, allowing the model to aggregate both inter-frame and intra-frame geometric information and establish long-range correspondences. 
On top of the shared latent representation, VGGT employs two lightweight decoding heads: a camera head, which predicts intrinsic and extrinsic camera parameters, and a Dense Prediction Transformer (DPT) \cite{ranftl2021vision} head, which produces dense geometric outputs including depth maps, point maps, and inter-frame tracks. 
This unified design enables the model to jointly estimate multiple geometric quantities in a single forward pass under multi-task supervision, achieving accurate and robust 3D perception across diverse visual domains.

In particular, VGGT is well suited to our setting as it enables multi-view–consistent geometry estimation and robust pixel-level tracking without task-specific training or manual camera calibration. In our framework, we exploit these capabilities to infer camera parameters from the camera head and inter-frame point tracks from the DPT head.

\subsection{Overall pipeline of VFM-SDM}

The proposed framework is designed to operate on multi-view video observations of a structure. In principle, increasing the number of viewpoints provides richer geometric information and complementary observations, which can improve the accuracy and robustness of displacement reconstruction. However, from a practical perspective of deployment simplicity and cost efficiency, this study focuses on the minimal stereo configuration with two viewpoints, which represents the smallest number of views required for 3D reconstruction. This extreme yet practical setting is adopted to validate the effectiveness of the proposed framework. In real-world applications, the number of viewpoints can be flexibly adjusted according to specific monitoring requirements and deployment constraints.

\begin{figure}[tb]
  \centering
  \includegraphics[width=\linewidth, trim=166 87 158 50, clip]{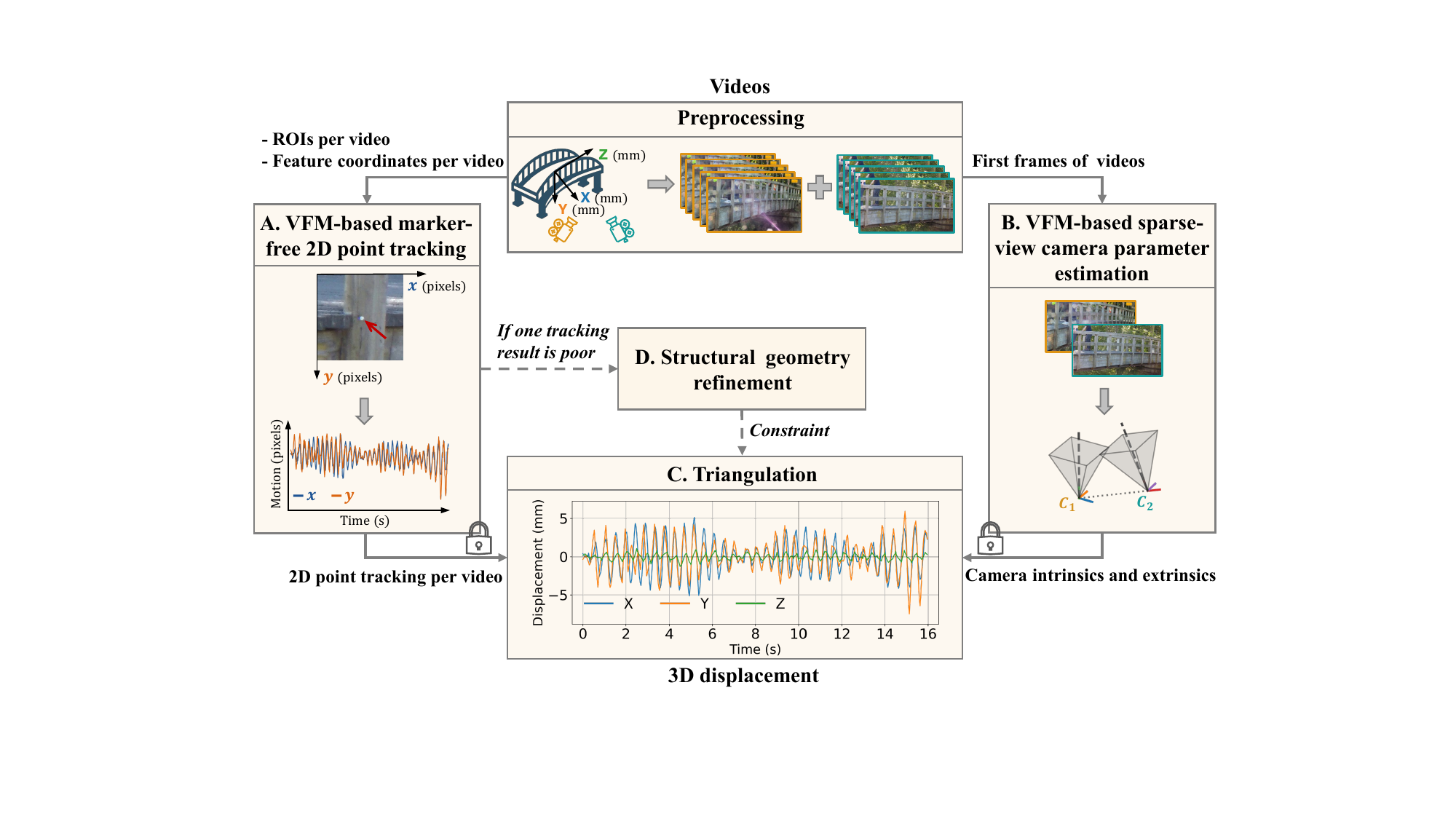}
  \caption{Overall pipeline of VFM-SDM for structural displacement measurement. The lock symbols indicate that the VFM-based modules operate with frozen parameters and require no task-specific training or fine-tuning. ROIs denote the regions of interest cropped from the video frames, which form the input ROI sequences for subsequent point tracking. Uppercase XYZ denote the structural coordinate system, while lowercase $xy$ denote the pixel coordinate system.}
  \label{fig:pipeline}
\end{figure}

As illustrated in Fig.~\ref{fig:pipeline}, the proposed VFM-based framework takes two video sequences of a structure as input, captured from different viewpoints using fixed cameras. After preprocessing, the videos are temporally synchronized and downsampled into corresponding frame pairs. No image enhancement, binarization, or denoising is applied; all frames are processed in their original form.
In the left branch, the synchronized sequences are fed into a VFM-based marker-free 2D point tracking module, which yields pixel-level 2D trajectories for each viewpoint. In parallel, the first frames from all videos are jointly processed by a VFM-based sparse-view camera parameter estimation module to recover the intrinsic and extrinsic parameters.
Both VFM modules operate with all parameters frozen, as indicated by the lock symbol, and are used without any task-specific training or fine-tuning.
The outputs of these branches are combined in the triangulation module to reconstruct the 3D coordinates of the tracked points over time. The reconstructed points are defined in the reference camera coordinate frame and are subsequently transformed into a structure-aligned coordinate frame for structural interpretation.
The orientation between the camera frame and the structural frame is approximated using simple geometric measurements. In this study, the reference camera is mounted on a tripod and leveled using a spirit level to ensure that the camera optical axis is parallel to the bridge deck. Under this setup, both the camera pitch and roll angles can be considered negligible.
Consequently, the transformation between the two coordinate frames can be approximated using only the horizontal angle between the camera optical axis and the bridge longitudinal axis. 
Specifically, two quantities are required: the perpendicular distance from the reference camera to the bridge, measured in situ, and the longitudinal distance from the bridge end to the structural component located at the center of the camera field of view (FOV), which can typically be obtained from bridge drawings.
These two orthogonal distances define a right triangle, from which the horizontal angle between the camera optical axis and the bridge longitudinal axis is estimated and used to perform the coordinate transformation.
Because the estimated camera translation is inherently scale-ambiguous, a known stereo baseline, that is, the fixed physical distance between the two camera centers, is used for metric scale recovery by applying a global scale factor to both camera translations and triangulated 3D points.
To enforce physically meaningful motion reconstruction, a geometry-guided structural refinement is integrated into the triangulation process. In field conditions, degraded video quality (e.g., defocus or glare) can introduce noise in 2D point tracking, which is amplified through triangulation and leads to physically implausible 3D displacement. 
The proposed refinement incorporates structural consistency priors, such as the negligible longitudinal motion of the bridge under normal service conditions, as reconstruction constraints, reflecting the fact that in-service normal infrastructures rarely exhibit notable displacements along all three spatial directions simultaneously. 
This process suppresses spurious longitudinal fluctuations while stabilizing tracking-induced noise in the vertical or lateral directions.

\begin{figure}[tb]
  \centering
  \includegraphics[width=\linewidth, trim=170 80 170 40, clip]{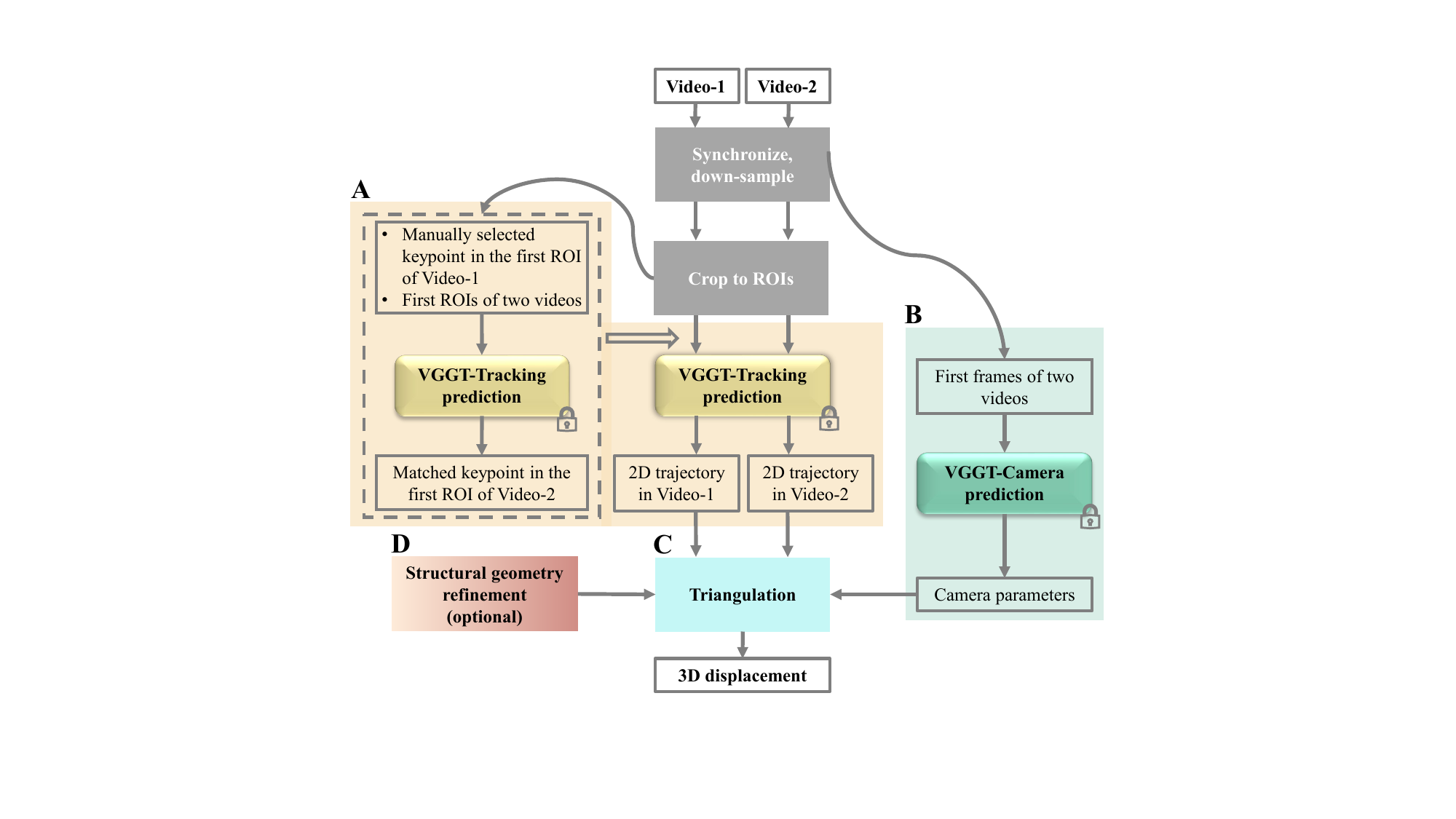}
  \caption{Complete computational pipeline of VFM-SDM. Capital letters A–D correspond to the four modules in Fig.~\ref{fig:pipeline}: (A) VFM-based marker-free 2D point tracking, (B) VFM-based sparse-view camera parameter estimation, (C) Triangulation, and (D) Structural geometry refinement.}
  \label{fig: processing}
\end{figure}

In our implementation, we adopt VGGT~\cite{wang2025vggt}, a recent 3D VFM, whose background and underlying principles are introduced in Sec.~\ref{subsec:pre}. We choose VGGT because, among current state-of-the-art vision foundation models, it is the only one capable of jointly estimating camera parameters and tracking points, making it particularly suited for displacement measurement. Accordingly, leveraging VGGT's robust spatial variation perception and geometric reasoning enables a compact pipeline that requires no training, markers, or calibration.
Next, we detail the operating process of the four main modules: VFM-based marker-free 2D point tracking, VFM-based sparse-view camera parameter estimation, triangulation, and structural geometry refinement. Using a binocular setup as an illustrative example, we describe the workflow and underlying principles of each module and provide the complete computational pipeline, as shown in Fig.~\ref{fig: processing}.

\subsection{VFM-based marker-free 2D point-tracking}
In the VFM-based marker-free 2D point-tracking module, the two input videos are first temporally synchronized and down-sampled. We then crop regions of interest (ROIs) at the monitoring area in both videos. A natural feature is manually selected as the keypoint to be tracked in the first ROI of Video-1, and VGGT’s tracking head is used to automatically locate the corresponding keypoint in the first ROI of Video-2. For each video, the identified keypoint coordinates and its ROI sequence are fed into VGGT’s tracking head, producing an independent pixel-level 2D trajectory for that viewpoint.

\subsection{VFM-based sparse-view camera parameter estimation}
In the VFM-based sparse-view camera parameter estimation module, we use the first frame from each of the two video sequences as a sparse stereo pair. The original full-frame images are directly fed into VGGT’s camera head, without any cropping, so that global scene context and fine-grained structural details are preserved, which enhances VGGT’s 3D geometric reasoning. In a single forward pass, VGGT estimates the intrinsic and extrinsic parameters of both cameras, with the first camera serving as the world-coordinate reference and the second camera’s pose represented relative to it. The resulting camera parameters, together with the 2D trajectories extracted from both videos, are then provided to the triangulation module for subsequent 3D displacement recovery.

\subsection{Triangulation}

In a stereo vision system for structural displacement monitoring, recovering the 3D motion of structural feature points requires knowledge of the camera intrinsics and extrinsics, including the intrinsic matrix $K$ (with $(f_x,f_y)$, $(c_x,c_y)$, skew $s$, and distortion coefficients 
$\boldsymbol{\kappa}$) and the relative pose $(R,t)$ between viewpoints. In our setting, these parameters are not obtained through conventional calibration but are instead estimated directly by a vision foundation model. Under the pinhole projection model, a homogeneous 3D point 
$\tilde{X}=[X^\top,1]^\top$ projects to an image point $\tilde{x}=[x^\top,1]^\top$ as
\begin{equation}
\tilde{x} \;\sim\; K\,[R \mid t]\,\tilde{X},
\qquad
K=\begin{bmatrix}
f_x & s & c_x\\
0 & f_y & c_y\\
0 & 0 & 1
\end{bmatrix},
\label{eq:pinhole}
\end{equation}
optionally followed by a distortion mapping $x \mapsto \mathcal{D}(x;\boldsymbol{\kappa})$.  
Although the parameters are predicted rather than calibrated, their consistency is sufficient to support reliable 3D reconstruction up to a global scale.

For a fixed stereo camera setup observing a structure, the estimated relative pose $(R_{12},t_{12})$ induces the essential matrix
\begin{equation}
x_2^{\top} E x_1 = 0,
\qquad
E = [t_{12}]_\times R_{12},
\label{eq:essential}
\end{equation}
where $x_1$ and $x_2$ denote the normalized image coordinates of tracked structural feature points (e.g., natural texture points on the bridge).
This epipolar geometry constrains valid correspondences and ensures geometric consistency prior to 3D reconstruction.

\begin{figure}[tb]
\centering
\includegraphics[width=0.85\linewidth]{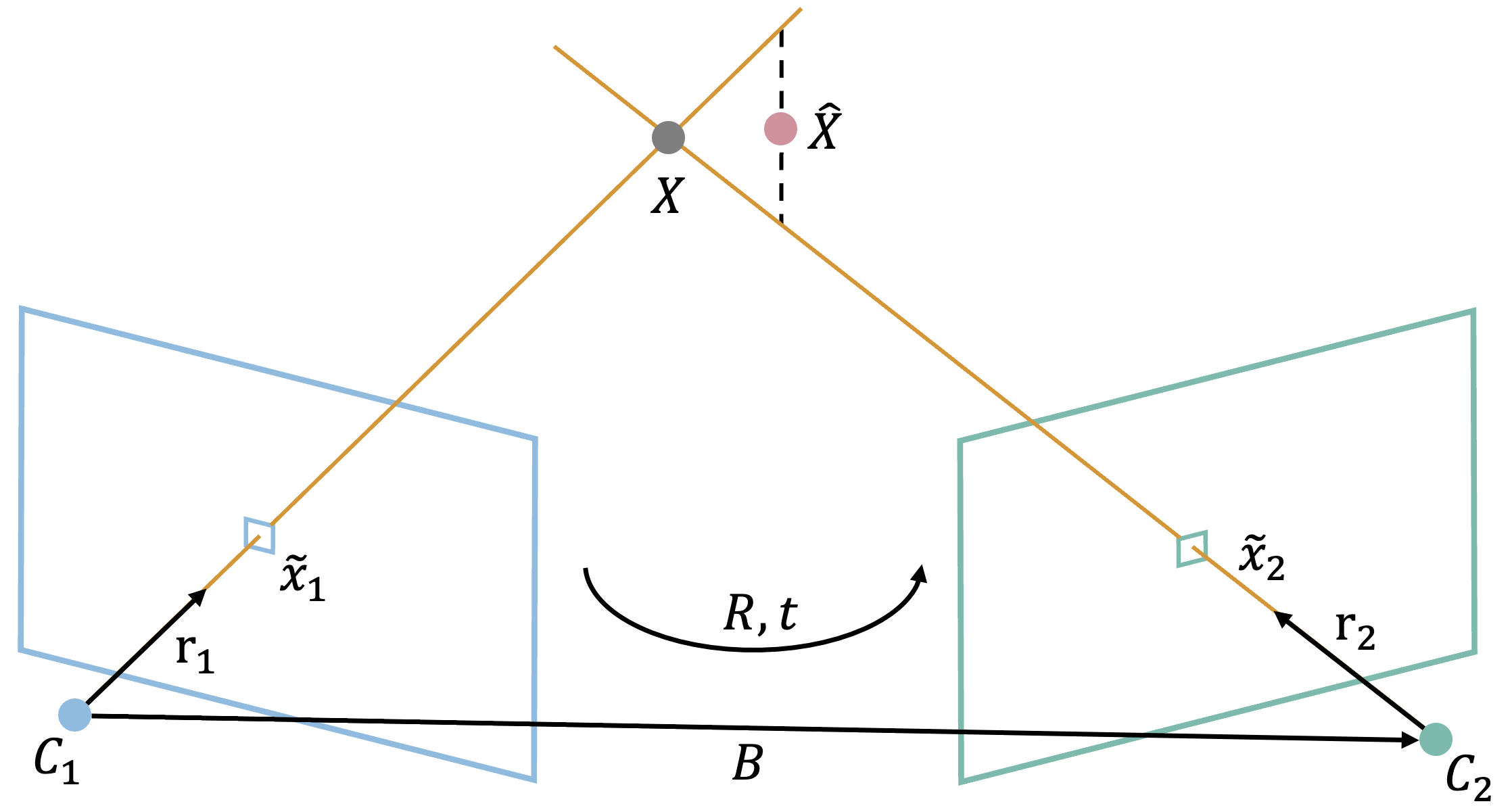}
\caption{
Geometric illustration of triangulation. The 3D point $X$ is observed by two estimated camera models with optical centers $C_1$ and $C_2$. 
Each observation $\tilde{x}_i$ defines a viewing ray $C_i + \lambda_i \mathbf{r}_i$, where $\mathbf{r}_i = R_i^\top K_i^{-1}\tilde{x}_i$. 
Due to noise or prediction error, the rays are skew and do not intersect exactly; the reconstructed point $\hat{X}$ is obtained as the midpoint of the shortest segment between them. The baseline $B$ connects the two camera centers.
}
\label{fig:triangulation}
\end{figure}

Once camera parameters and 2D tracks are available, triangulation reconstructs the 3D position of a structural point as it moves over time. Each image observation back-projects to a spatial ray:
\[
\mathbf{r}_i = R_i^\top K_i^{-1}\tilde{x}_i,
\qquad C_i = -R_i^\top t_i,
\]
where $(C_i,\mathbf{r}_i)$ represent the optical center and viewing direction of camera $i$. Due to tracking noise and prediction inaccuracies, the two rays are generally skew rather than exactly intersecting, and the 3D point is estimated via linear triangulation, as illustrated in Fig.~\ref{fig:triangulation}.

An algebraic triangulation formulation follows directly from Eq.~\eqref{eq:pinhole}.
Specifically, the homogeneous image projection equation can be rewritten as a set of linear constraints by eliminating the unknown projective scale factor. 
Each image observation yields two linearly independent constraints on $\tilde{X}$:
\begin{equation}
(u_i \Pi_i^{(3)} - \Pi_i^{(1)})\tilde{X}=0, \qquad
(v_i \Pi_i^{(3)} - \Pi_i^{(2)})\tilde{X}=0,
\end{equation}
where $(u_i,v_i)$ denote the pixel coordinates of the observed image point in camera $i$, corresponding to the normalized image coordinates $x_i$ defined above, $\Pi_i = K_i[R_i\mid t_i]$ is the projection matrix of camera $i$, and $\Pi_i^{(1)}$, $\Pi_i^{(2)}$, and $\Pi_i^{(3)}$ denote its first, second, and third row vectors, respectively.

Stacking the constraints from both cameras yields
\begin{equation}
A\tilde{X}=0,
\qquad A\in\mathbb{R}^{4\times 4},
\label{eq:triang_lin}
\end{equation}
which is solved via SVD, and the solution $\tilde{X}=(X,Y,Z,W)^\top$ is dehomogenized to obtain Euclidean coordinates.

Because Eq.~\eqref{eq:triang_lin} is homogeneous, the reconstructed 3D points are determined only up to a global scale. In structural monitoring, recovering an accurate metric scale is essential for estimating displacement magnitudes. The physical stereo baseline $B$ provides this scale reference. 
In rectified stereo geometry, depth can be expressed as
\begin{equation}
Z = \frac{f\,B}{d},
\label{eq:depth}
\end{equation}
where $d$ denotes disparity. In our setting, the baseline implied by the predicted camera translations is denoted as $B_{\text{est}}$. If the true physical baseline $B_{\text{true}}$ is known, the metric scale can be recovered through
\[
s = \frac{B_{\text{true}}}{B_{\text{est}}},
\]
which is applied to all reconstructed trajectories.

Triangulation therefore converts pixel correspondences into metrically accurate 3D trajectories of natural feature points on a structure. Its accuracy depends primarily on the consistency of the predicted camera parameters, the stereo baseline, and the reliability of 2D tracking. As such, triangulation forms the geometric core of the proposed VFM-based structural displacement measurement framework.

\subsection{Structural geometry refinement}

This refinement is applicable to the vast majority of infrastructures operating under normal service conditions.
Taking a common civil infrastructure example such as a bridge, in real bridge structures subjected to normal service loads, the displacement response is inherently 3D. However, because both ends of the bridge are constrained by bearings or abutments, the motion primarily occurs in the vertical and lateral directions, while the longitudinal component is typically negligible with only minor temporal variations. In practical image acquisition, video quality may degrade when cameras lose focus or when strong illumination induces glare, leading to noisy or inconsistent 2D point tracking. Such tracking noise can propagate through stereo triangulation and produce reconstructed 3D motions that deviate from these expected structural behaviors. 
To mitigate implausible deviations and to regularize the final displacement trajectories, we incorporate a structural geometry refinement (SGR) step into the triangulation stage to encourage deformation patterns that are consistent with the typical response of bridges under service conditions.

Building on this idea, the SGR module enforces structural and geometric consistency in the reconstructed motion as an optional refinement step integrated into the triangulation process. It is designed for scenarios in which point tracking from one view is relatively reliable, while the other view is affected by higher uncertainty due to video degradation. In such cases, the more reliable view serves as a reference, whereas the uncertain view is refined; if both views exhibit severe tracking errors, the refinement is not applicable due to the lack of a trustworthy reference.
Specifically, the SGR module refines the uncertain view by treating its image coordinates as optimization variables, while keeping the reference view and all camera parameters fixed. 
In practice, only the horizontal pixel coordinates are optimized, while the vertical pixel coordinates remain fixed. In the image coordinate system, the monitored target typically exhibits larger motion in the vertical direction, resulting in a higher signal-to-noise ratio and thus more reliable tracking. In contrast, horizontal motion is usually much smaller, making the corresponding tracking estimates more susceptible to noise introduced during feature detection and matching. The refinement therefore focuses on the horizontal direction while preserving the more stable vertical estimates.
At each iteration, the corrected image coordinates are undistorted, re-triangulated, and transformed into the bridge coordinate frame to ensure temporal coherence of the reconstructed 3D motion. 

The refinement is formulated as an optimization problem guided by three objectives:
(i) suppress longitudinal motion,
(ii) preserve lateral and vertical displacements relative to the baseline triangulation, and
(iii) constrain pixel corrections to remain small.
Let $\mathbf{u}_2(t,p)=[u_{2x}(t,p),u_{2y}(t,p)]^\top$ denote the tracked image coordinate of point $p$ at time $t$ in the right view. 
Only the horizontal coordinate is refined, while the vertical coordinate remains unchanged. 
The corrected observation is defined as

\begin{equation}
\tilde{\mathbf{u}}_2(t,p)=
\begin{bmatrix}
u_{2x}(t,p)+\Delta u_{2x}(t,p)\\
u_{2y}(t,p)
\end{bmatrix}.
\end{equation}

Using the corrected observations, stereo triangulation yields the 3D displacement trajectory $(X(t,p),Y(t,p),Z(t,p))$ in the bridge coordinate system. 
The refinement is obtained by minimizing

\begin{equation}
\min_{\{\Delta u_{2x}(t,p)\}} E ,
\label{eq:sgr_objective}
\end{equation}

where

\begin{equation}
\begin{aligned}
E =
&\; w_{Z,\mathrm{abs}} \sum_{t,p} Z(t,p)^2
+ w_{Z,\mathrm{diff}} \sum_{t,p} \Delta Z(t,p)^2 \\
&+ w_{XY,\mathrm{abs}} \sum_{t,p}
\Big[(X(t,p)-X^{(0)}(t,p))^2+(Y(t,p)-Y^{(0)}(t,p))^2\Big] \\
&+ w_{XY,\mathrm{diff}} \sum_{t,p}
\Big[(\Delta X(t,p)-\Delta X^{(0)}(t,p))^2
+(\Delta Y(t,p)-\Delta Y^{(0)}(t,p))^2\Big] \\
&+ w_{2d}\sum_{t,p}\Delta u_{2x}(t,p)^2 .
\end{aligned}
\label{eq:sgr_expand}
\end{equation}

Here, $\Delta$ denotes the temporal difference operator along the time dimension. $Z$ denotes the longitudinal displacement component, while $X$ and $Y$ denote the lateral and vertical components, respectively. The first two terms suppress the magnitude and temporal variation of longitudinal motion, the next two preserve the lateral and vertical displacements and their temporal differences relative to the baseline triangulation, and the last term regularizes the horizontal pixel correction. The optimization procedure is summarized in Algorithm~\ref{alg:right-refine}, illustrated for the case of right-view optimization.

\begin{algorithm}[H]
\caption{Triangulation with structural geometry refinement (SGR), illustrated for the case of right-view optimization.}
\label{alg:right-refine}
\begin{algorithmic}[1]
\Require Synchronized tracks $u_1(t,p),u_2(t,p)$; camera parameters; relative pose $[R|T]$
\Ensure Corrected right-view tracks $u_2^{\star}(t,p)$ and refined 3D trajectories

\State \textbf{Baseline triangulation:}
undistort $(u_1,u_2)$, triangulate using $[I|0]$ and $[R|T]$, and map the 3D points to the bridge coordinate system
\State Compute baseline trajectories $(X^{(0)},Y^{(0)},Z^{(0)})$ and temporal differences $(\Delta X^{(0)},\Delta Y^{(0)},\Delta Z^{(0)})$
\State Initialize horizontal pixel corrections $\Delta u_{2x} \gets 0$

\Repeat
\State Update right-view observations
\[
\tilde{\mathbf{u}}_2 =
\begin{bmatrix}
u_{2x}+\Delta u_{2x}\\
u_{2y}
\end{bmatrix}
\]
\State Triangulate using $(u_1,\tilde{\mathbf{u}}_2)$ and obtain updated trajectories $(X,Y,Z)$ in the bridge frame
\State Compute temporal differences $(\Delta X,\Delta Y,\Delta Z)$
\State Form residuals corresponding to Eq.~\eqref{eq:sgr_expand}:
\Statex \quad (i) longitudinal suppression $\bigl(Z,\Delta Z\bigr)$
\Statex \quad (ii) lateral/vertical fidelity $\bigl(X-X^{(0)},\,Y-Y^{(0)},\,\Delta X-\Delta X^{(0)},\,\Delta Y-\Delta Y^{(0)}\bigr)$
\Statex \quad (iii) horizontal pixel regularization $\bigl(\Delta u_{2x}\bigr)$
\State Update $\Delta u_{2x}$ by least-squares optimization of Eq.~\eqref{eq:sgr_objective}
\Until convergence or maximum iterations

\end{algorithmic}
\end{algorithm}

\section{Data acquisition and dataset overview}
\label{sec:dataset}
\begin{figure*}[t]
\centering
\includegraphics[height=4.0cm,keepaspectratio]{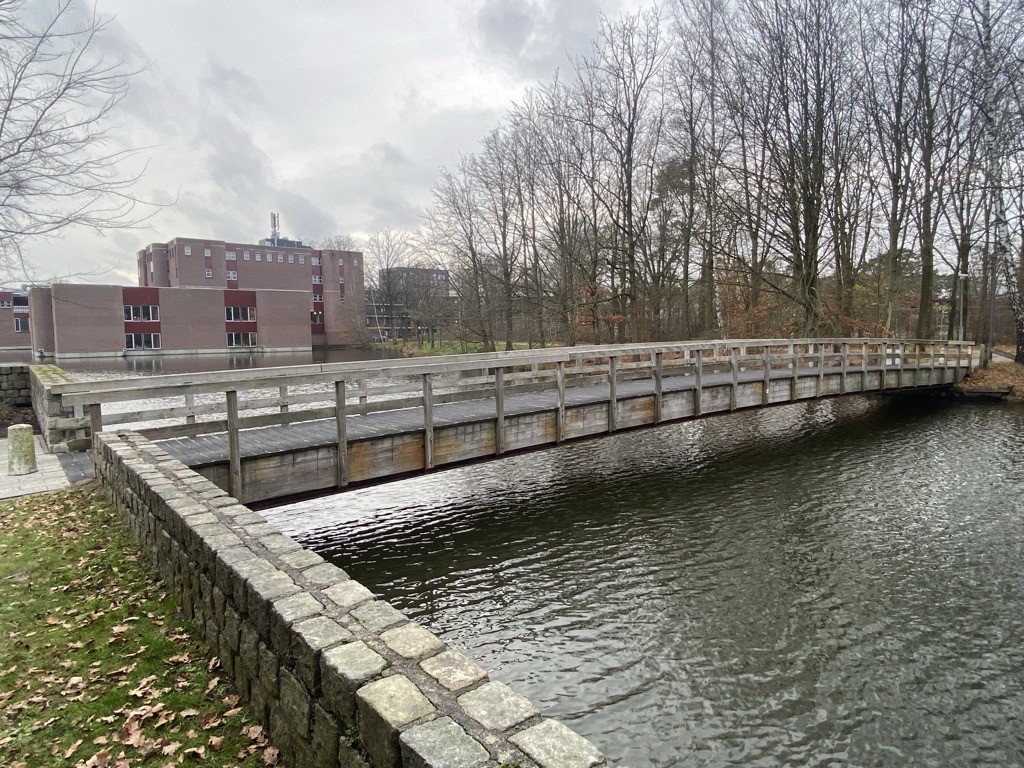}
\hspace{0.1cm}
\includegraphics[height=4.0cm,keepaspectratio]{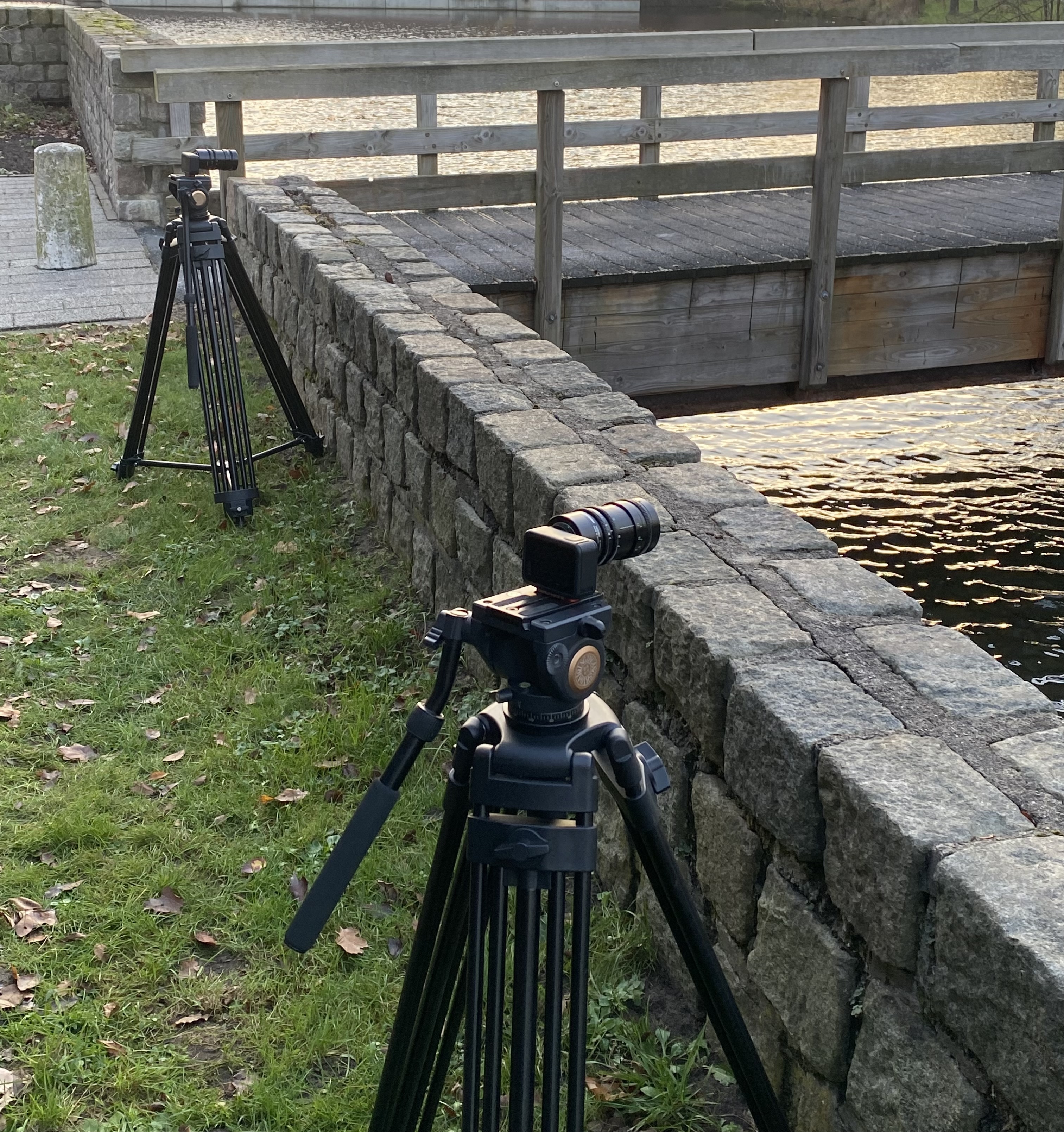}
\hspace{0.1cm}
\includegraphics[height=4.0cm,keepaspectratio]{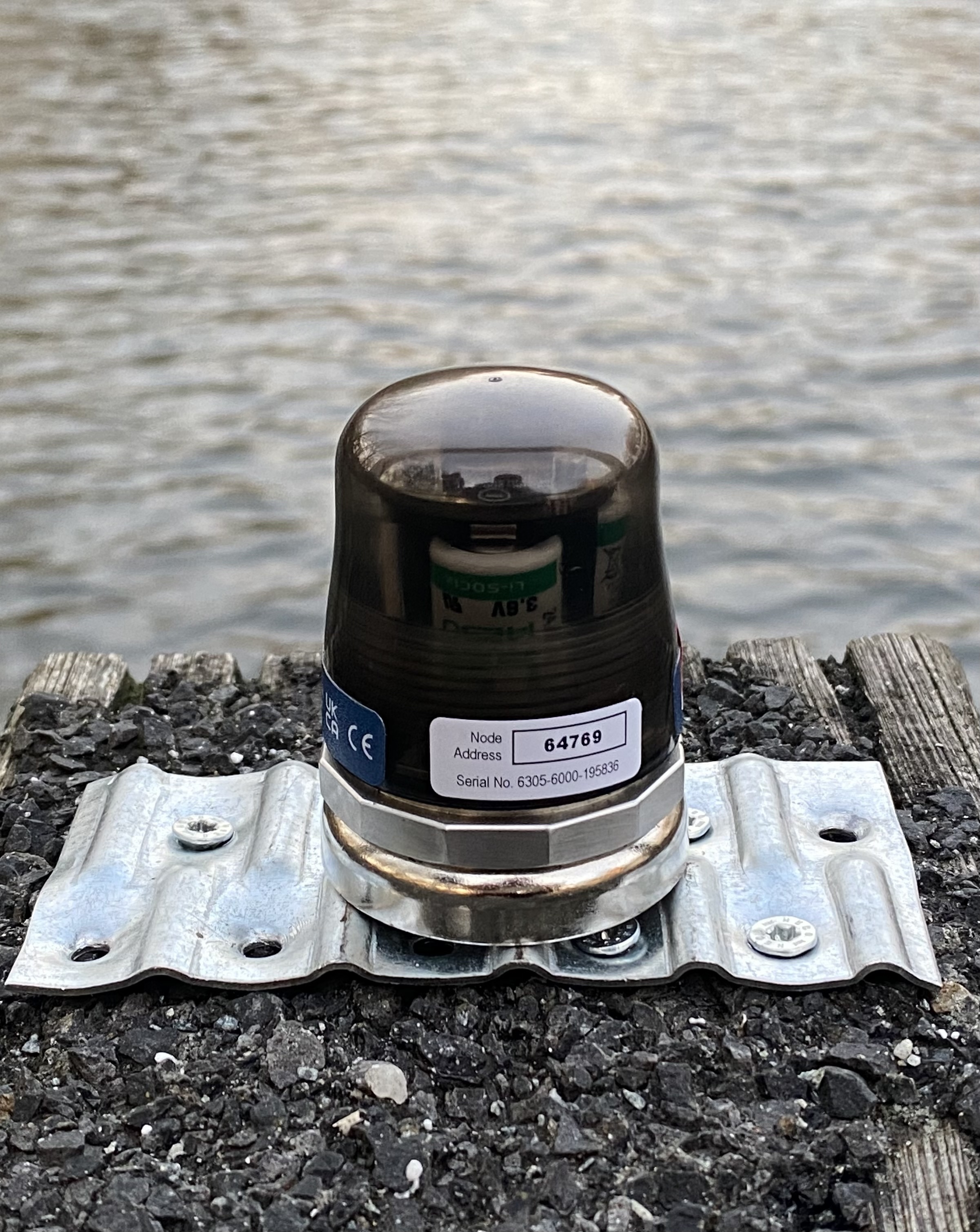}

\caption{Overview of the data acquisition setup. From left to right: bridge overview, stereo vision configuration, and accelerometer installation on the bridge.}
\label{fig:overall_setup}
\end{figure*}

All experiments were conducted on an in-service steel girder pedestrian bridge located on the University of Twente campus, which is covered with a wooden deck and therefore appears visually as a wooden bridge (Fig.~\ref{fig:overall_setup}). 
All data were collected under real outdoor conditions, with uncontrolled lighting and wind. 
Accordingly, this section presents the experimental setup, data collection procedure, dataset summary, and evaluation metrics in sequence.

\subsection{Experimental setup}
A stereo vision system was constructed using two modified GoPro cameras equipped with conventional 8--48 mm and telephoto 25--135 mm lenses, respectively. The two cameras were mounted on separate tripods on the same side of the bridge. 
Due to site constraints, the range of variation in camera positions and orientations was limited; however, the cameras were reinstalled and adjusted from scratch in each experiment without using any predefined reference setups, resulting in similar yet distinct camera positions and orientations across different experiments.
Both cameras shared an overlapping field of view covering the target monitoring region. 
All videos were recorded in 4K resolution at 60 fps, ensuring high spatial and temporal fidelity for displacement extraction. This configuration produced a stable yet flexible stereo setup for capturing bridge displacement under different geometric baselines, i.e., the distance between the two camera centers, and viewpoints (see Fig.~\ref{fig:overall_setup}).
In our experimental setup, the two cameras were mounted on tripods at approximately the same height, and the baseline was measured directly as the distance between the tripod centers on the ground using a tape measure. In scenarios where the cameras are placed farther apart or at different heights, the baseline can be obtained using simple and portable tools, such as handheld laser distance meters.

\begin{figure}[H]
\centering
\begin{subfigure}{\textwidth}
    \centering
    \includegraphics[width=0.48\textwidth]{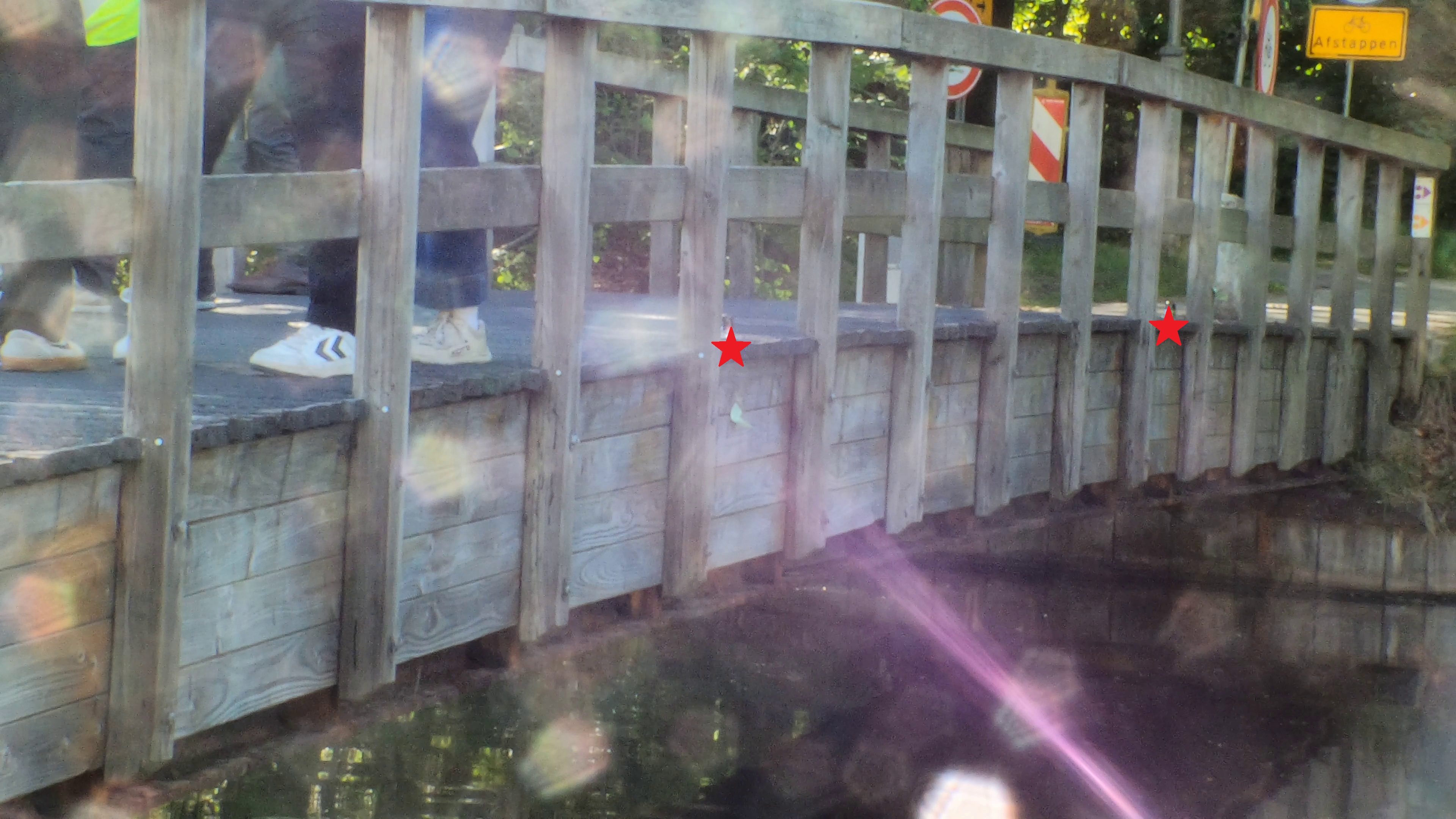}%
    \hfill
    \includegraphics[width=0.48\textwidth]{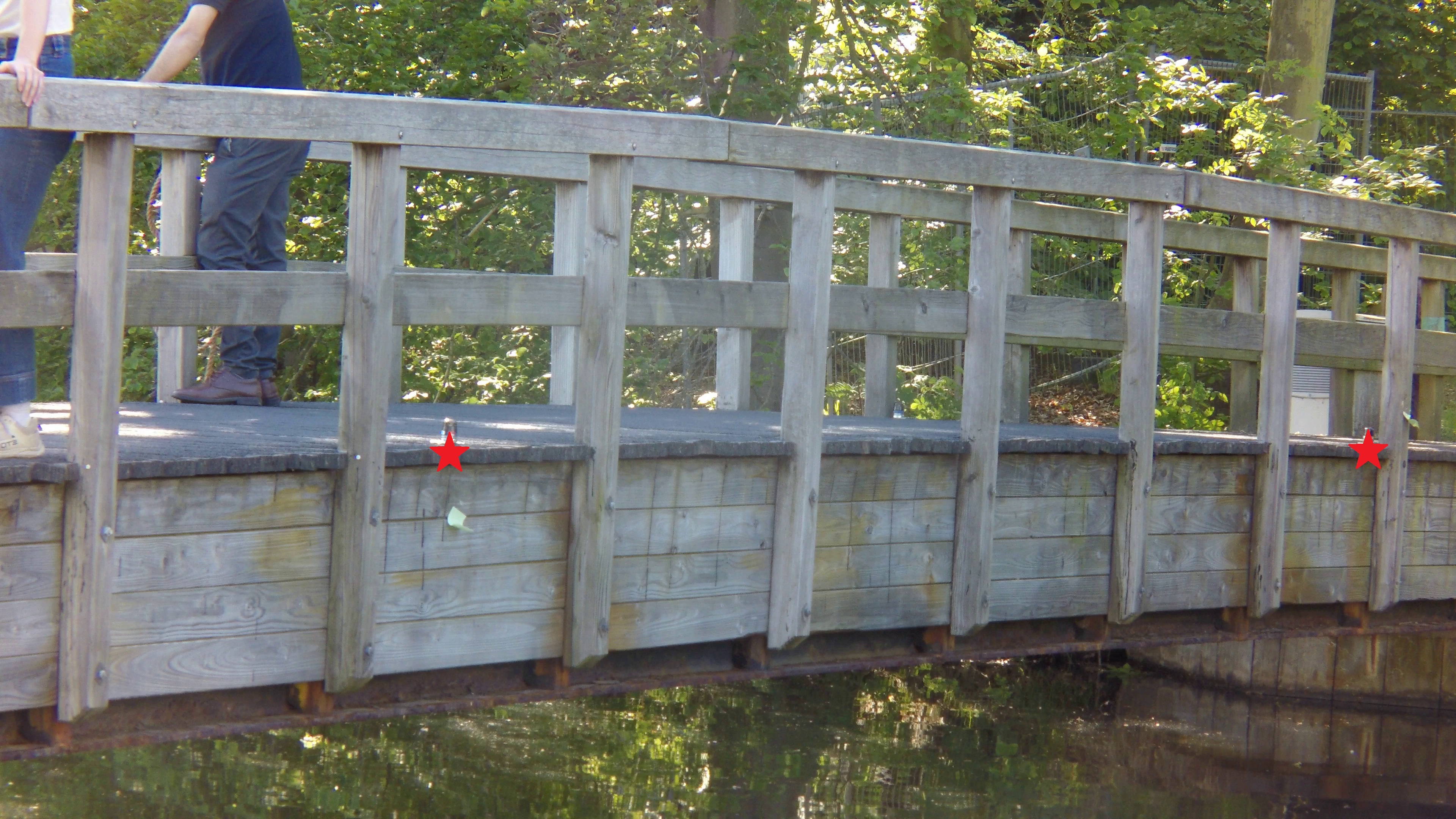}
    \caption{}
    \label{fig:data1_stereo}
\end{subfigure}

\vspace{2pt}

\begin{subfigure}{\textwidth}
    \centering
    \includegraphics[width=0.48\textwidth]{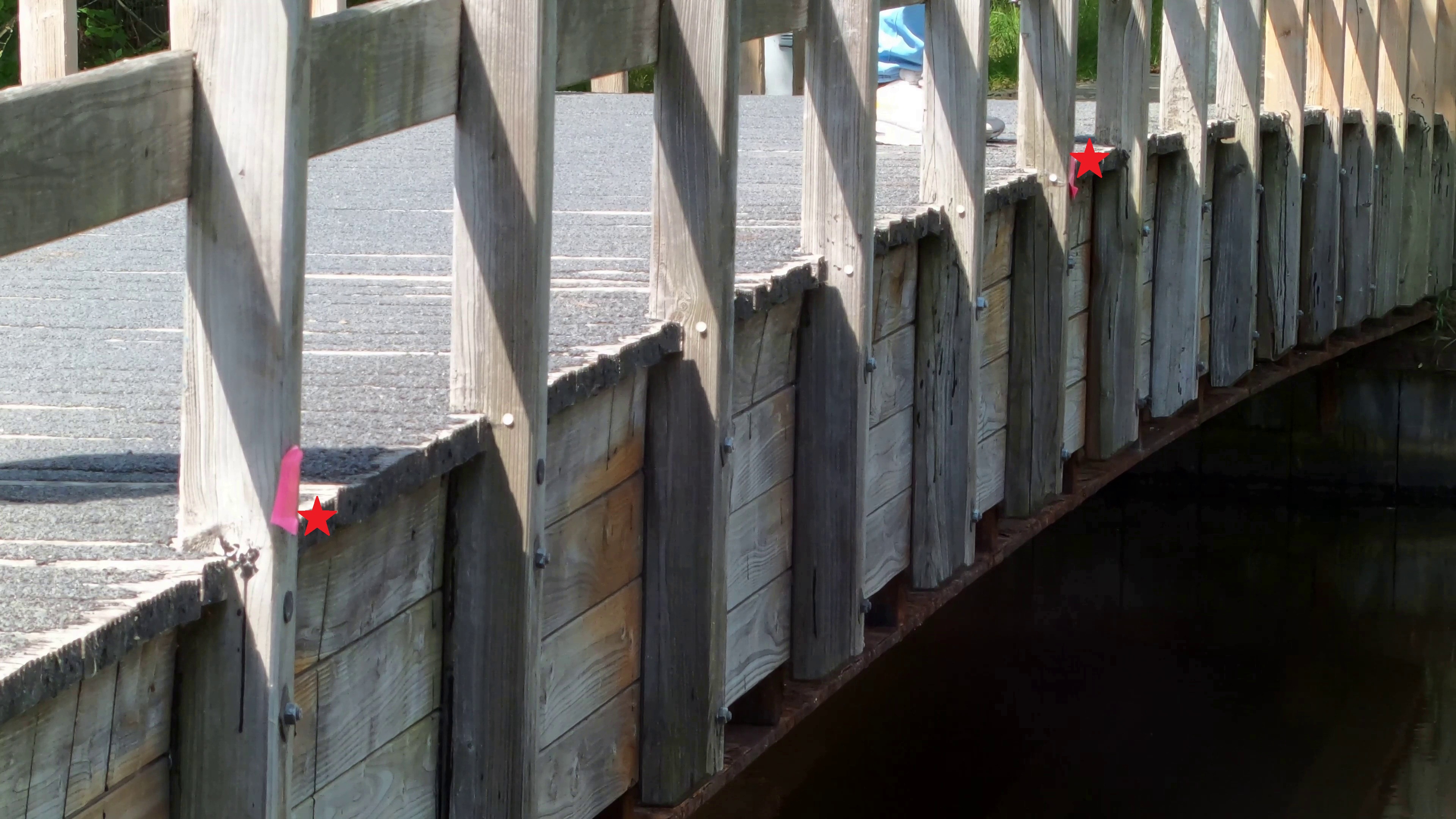}%
    \hfill
    \includegraphics[width=0.48\textwidth]{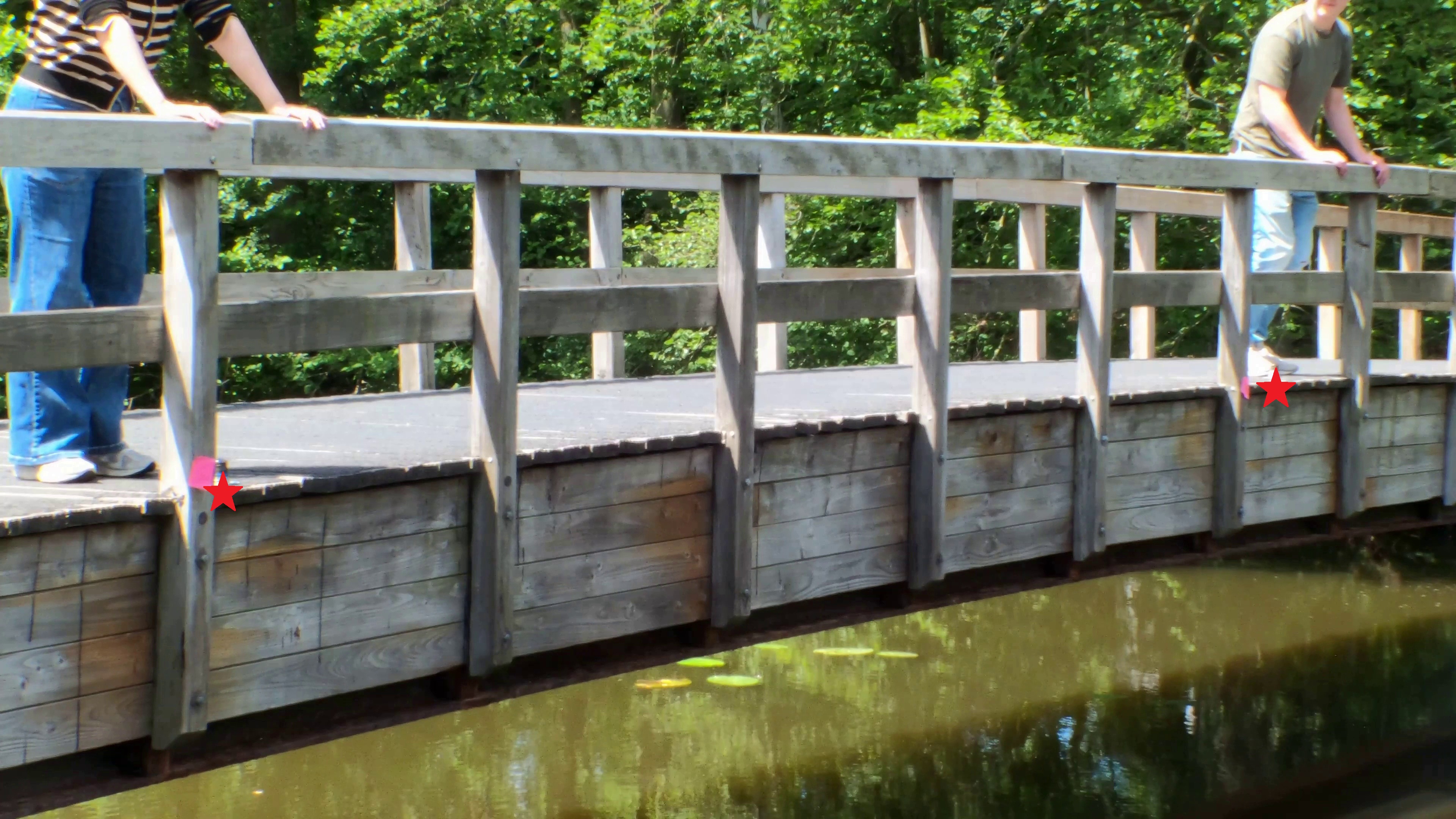}
    \caption{}
    \label{fig:data2_stereo}
\end{subfigure}

\vspace{2pt}

\begin{subfigure}{\textwidth}
    \centering
    \includegraphics[width=0.48\textwidth]{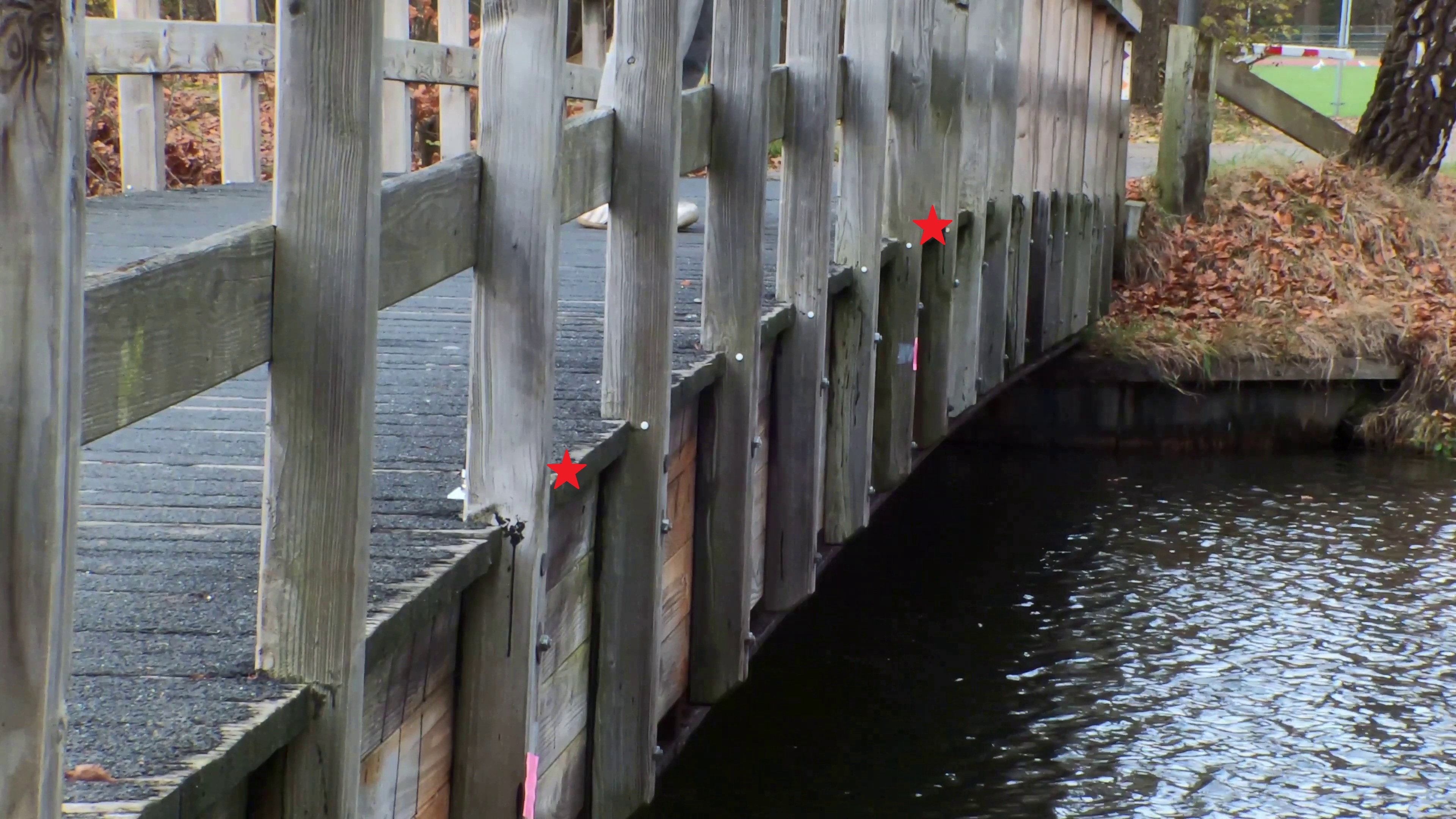}%
    \hfill
    \includegraphics[width=0.48\textwidth]{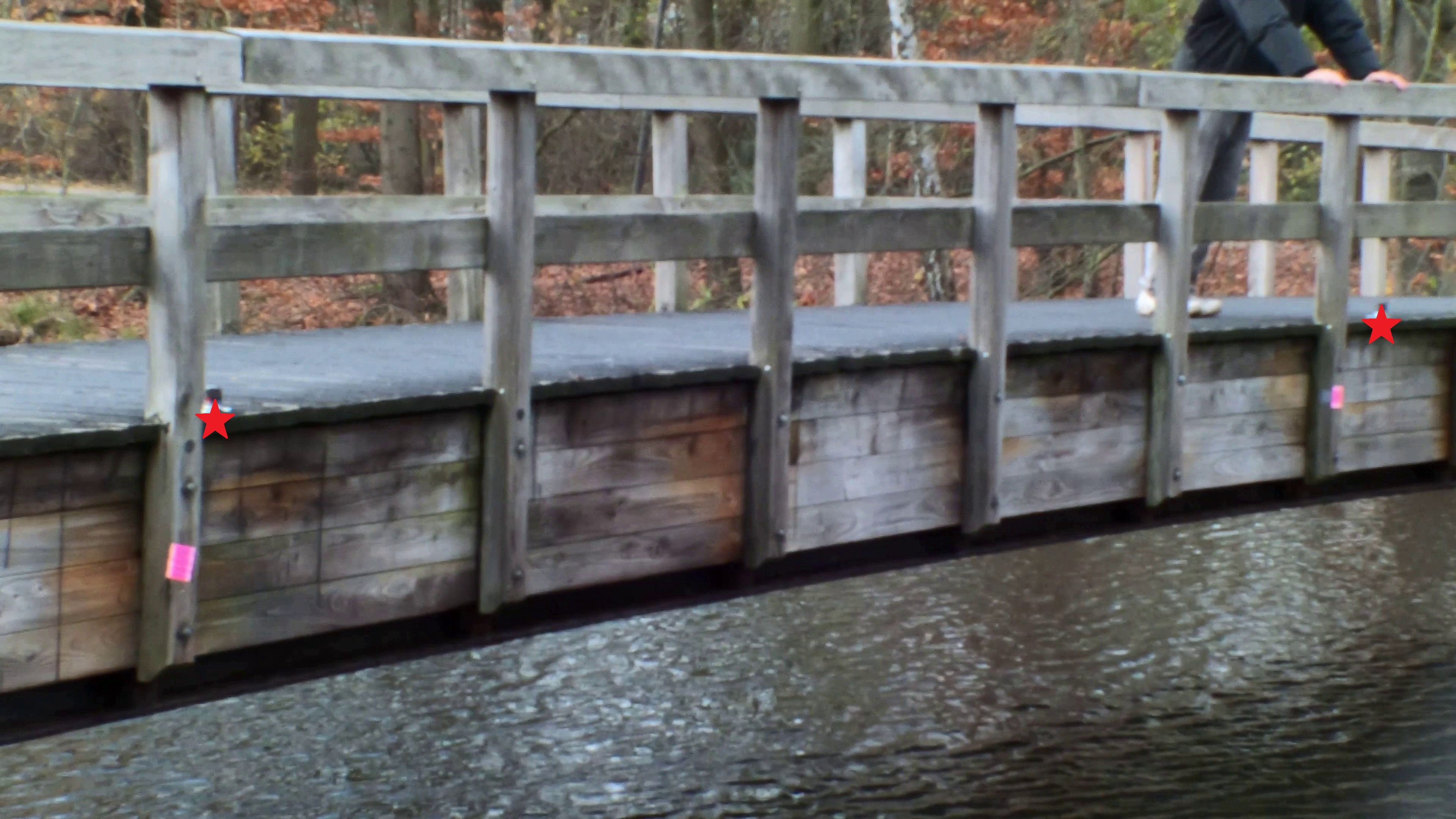}
    \caption{}
    \label{fig:data3_stereo}
\end{subfigure}

\caption{
Overview of the three stereo sequences in the proposed dataset. Each row shows the first frames from the left (camera~1) and right (camera~2) views of one stereo sequence, ordered from top to bottom as (a) Data-1, (b) Data-2, and (c) Data-3. The sticky notes placed on the bridge are used solely to expedite setup by facilitating localization of the monitored region within the FoV, rather than for tracking. The red stars indicate the locations of the accelerometers.}
\label{fig:dataset_stereo_frames}
\end{figure}

To obtain an effective reference for comparison with our vision-based displacement measurements, two battery-powered G-Link-200 tri-axial wireless accelerometers (64\,Hz sampling rate), synchronized in time, were rigidly attached to two locations on the bridge deck (see Fig.~\ref{fig:overall_setup} and Fig.~\ref{fig:dataset_stereo_frames}). Both accelerometers were installed on the same side of the bridge, corresponding to the side closer to the cameras, and were aligned with the two monitored regions within the camera fields of view.
The corresponding accelerometer-derived displacement reference was obtained by integrating the recorded acceleration signals.

In indoor and some outdoor experiments, laser displacement sensors and Linear Variable Differential Transformer (LVDT) sensors are often regarded as reliable reference instruments for structural displacement measurement. However, their deployment becomes considerably more challenging in outdoor field conditions. In our case, the bridge spans over water, making it difficult to establish an appropriate measurement configuration with a stable reference frame beneath the structure for vertical displacement measurement. 
As a result, laser sensors may need to be deployed in oblique configurations due to geometric constraints, yielding indirect measurements that require geometric projection to recover vertical or lateral displacement components in the structural coordinate system, thereby introducing additional uncertainty.
LVDT sensors require direct contact and fixed support points, which are typically infeasible in such scenarios.
Compared with accelerometers, these sensors are generally more costly and less flexible to deploy.
By contrast, accelerometers are relatively inexpensive, easy to install, and widely used in field structural monitoring \cite{huang2021deep, panigati2025dynamic, shao2022target, ma2022structural, ghandil2021enhanced, wang2022completely, ponsi2024vision, wu2025displacement, bolognini2022vision}. Several prior studies have employed accelerometer measurements, combined with numerical integration, to derive displacement responses and serve as reference signals for validating vision-based methods \cite{huang2021deep, ghandil2021enhanced, ponsi2024vision, wu2025displacement}.
Following established practice, the acceleration signals in our experiments are processed using a unified pipeline and integrated to obtain displacement responses, ensuring consistent and reliable reference measurements across all datasets.

\subsection{Data collection procedure}

Bridge was excited by pedestrians running or jumping on its deck. As this study focuses on the accurate extraction of structural displacement rather than the relationship between applied loads and structural response, pedestrian weights and excitation were not deliberately controlled or recorded. In the first sequence, multiple pedestrians randomly applied excitation by running or jumping on the bridge, whereas in the remaining two sequences, excitation was randomly induced by two pedestrians running or jumping for approximately 20\,s. For each trial, the camera position and orientation, environmental conditions, and pedestrian excitation patterns differed, resulting in three distinct displacement scenarios.

Because strict hardware synchronization between the two GoPro cameras is difficult to achieve in outdoor conditions, stereo videos are synchronized in post-processing using response-based temporal alignment. The temporal offset between viewpoints is estimated by matching the 2D trajectories extracted independently from each view. The accelerometers provide absolute world-time timestamps, enabling an initial coarse alignment with the video streams, followed by finer synchronization by aligning accelerometer-based and vision-based vibration trajectories. 
This procedure yields temporally aligned stereo videos and accelerometer-derived displacement references for each sequence.

\subsection{Dataset summary}
Each of the three sequences includes:
\begin{itemize}
    \item synchronized left--right stereo videos (each with a duration of 16\,s),
    \item temporally aligned accelerometer-derived displacement references,
    \item acquisition metadata (acceleration signals, sensor positions, etc.).
\end{itemize}

\begin{table}[b]
\centering
\fontsize{8pt}{8pt}\selectfont
\caption{Summary of the three GoPro-based vibration sequences collected on a real pedestrian bridge.
The labels 1/4 span, 1/2 span, and 3/4 span denote the quarter-span, mid-span, and three-quarter-span locations, respectively, measured from the bridge end closer to the cameras and defined as the zero reference.}
\label{tab:data_summary_main}
\resizebox{\columnwidth}{!}{
\begin{tabular}{lcccc>{\raggedright\arraybackslash}m{4.6cm}}
\toprule
\textbf{Dataset} & \textbf{Date} & \textbf{Monitored Points} & \textbf{Devices} & \textbf{Baseline (cm)} & \textbf{Environmental Condition} \\
\midrule
Data-1 & 2025-05-14 & 1/2 span, 3/4 span & 2 GoPros & 396 & Sunny; glare on the left perspective \\
Data-2 & 2025-06-19 & 1/4 span, 1/2 span & 2 GoPros & 400 & Sunny; blur in the left perspective \\
Data-3 & 2025-11-29 & 1/4 span, 1/2 span & 2 GoPros & 497 & Cloudy; blur and dark illumination in the right perspective \\
\bottomrule
\end{tabular}
}
\end{table}

Table~\ref{tab:data_summary_main} summarizes the key properties of the collected sequences, where 1/4 span, 1/2 span, and 3/4 span denote locations measured from the bridge end closer to the cameras (defined as the zero reference), corresponding to the quarter-span, mid-span, and three-quarter-span positions of the bridge, respectively; the same notation applies hereafter.
The vibration sequences, captured using GoPro cameras, exhibit varying degrees of imperfect imaging quality.
This is primarily due to uncontrollable outdoor factors such as weather conditions and illumination variations, as well as occasional misfocus during image acquisition, as illustrated in Fig.~\ref{fig:dataset_stereo_frames}.
To facilitate rapid field deployment under such conditions, small sticky notes were temporarily placed on the bridge to assist in visually localizing the monitored region within the camera fields of view during setup. These markers were not used for feature tracking or displacement estimation and do not play any role in the proposed framework.
Consequently, the acquired videos do not represent ideal laboratory conditions but instead reflect the practical challenges commonly encountered in real-world field data acquisition. Experiments conducted on such data therefore offer a more realistic evaluation of the effectiveness and practical robustness of the proposed framework.
In addition to the GoPro-based stereo experiments (Data-1, Data-2, and Data-3), we also collected a smartphone-based stereo sequence using an iPhone~11 and an iPhone~15 Pro Max, with synchronized accelerometer measurements (Data-4); further details are provided in the \ref{app:phone}.

\subsection{Evaluation metrics}
To quantitatively assess the accuracy of vision-based displacement measurements, we adopt three complementary metrics that jointly evaluate waveform fidelity, displacement magnitude consistency, and overall agreement with accelerometer-derived displacement references. Following prior studies in vision-based structural displacement measurement (e.g., \cite{zhang2024automated}), we employ the range-normalized RMSE \cite{hyndman2006another} (NRMSE\textsubscript{range}) and the correlation coefficient \cite{lee1988thirteen} as primary metrics to quantify the similarity between the estimated and reference displacement time series. These two indices capture both amplitude-normalized waveform deviations and phase-aligned temporal consistency.

In addition to these measures, we introduce the Relative Peak-to-Peak Amplitude Error (RPPAE) \cite{kat2012validation} as a supplementary metric to specifically assess the agreement in vibration magnitude. While NRMSE\textsubscript{range} reflects overall waveform accuracy, it does not explicitly isolate errors in displacement amplitude, which are of particular relevance in structural vibration analysis. 
RPPAE therefore provides a direct and interpretable measure of amplitude consistency, complementing NRMSE\textsubscript{range} and correlation metrics that primarily characterize waveform similarity and temporal alignment.

\textbf{Normalized Root Mean Square Error (NRMSE\(_{\text{range}}\)).}
The RMSE measures the average pointwise deviation between the predicted displacement $\hat{d}(t)$ and the reference signal $d(t)$. To account for different vibration amplitudes across experiments, we normalize RMSE by the dynamic range of the reference displacement:
\begin{equation}
    \mathrm{NRMSE}_{\mathrm{range}} 
    = \frac{\sqrt{\frac{1}{N}\sum_{t=1}^N \left( \hat{d}(t) - d(t) \right)^2 }}
    {\max_t d(t) - \min_t d(t)}.
\end{equation}
A lower NRMSE\(_{\text{range}}\) indicates better agreement in the absolute displacement values, and is widely used in structural dynamics to quantify amplitude-consistent signal reconstruction.

\textbf{Correlation coefficient.}
The Pearson correlation coefficient evaluates the similarity of waveform shape regardless of scale:
\begin{equation}
    \rho = \frac{
        \sum_{t=1}^N \left( \hat{d}(t)-\overline{\hat{d}} \right)
        \left( d(t)-\overline{d} \right)
    }{
        \sqrt{\sum_{t=1}^N \left( \hat{d}(t)-\overline{\hat{d}} \right)^2 }
        \sqrt{\sum_{t=1}^N \left( d(t)-\overline{d} \right)^2 }
    }.
\end{equation}
This metric reflects the accuracy with which the method captures the temporal evolution and oscillatory characteristics of the structural response. A higher correlation ($\rho \rightarrow 1$) indicates better agreement in waveform shape and temporal consistency between the estimated and reference displacement signals.

\textbf{Relative Peak-to-Peak Amplitude Error (RPPAE).}
For vibration monitoring, peak-to-peak amplitude is a physically meaningful measure of structural response severity. We compute the global peak-to-peak amplitudes of the predicted and reference signals,
\begin{equation}
    A_{\mathrm{pred}} = \max_t \hat{d}(t) - \min_t \hat{d}(t), \quad
    A_{\mathrm{refer}}    = \max_t d(t) - \min_t d(t),
\end{equation}
and define the relative error as
\begin{equation}
    \mathrm{RPPAE} = 
    \frac{|A_{\mathrm{pred}} - A_{\mathrm{refer}}|}{A_{\mathrm{refer}}}.
\end{equation}
This metric directly evaluates the method's ability to recover the correct vibration amplitude, which is critical for structural condition assessment and safety evaluation.

Together, these three metrics provide a comprehensive evaluation of displacement measurement performance, capturing amplitude fidelity, temporal coherence, and physically meaningful vibration magnitude.

\section{Experiments}
\label{sec:experiments}
Due to the absence of publicly available benchmark datasets for structural displacement measurement under real-world conditions, the experiments are conducted on the vibration sequences collected in our proposed dataset. 
This section first describes the implementation details, followed by quantitative and qualitative evaluations on the proposed dataset. A computational cost analysis is then provided to assess the efficiency of the proposed framework, and finally a sensitivity analysis is carried out to further analyze its robustness.

\subsection{Implementation details}
The key implementation settings include temporal resolution alignment, VGGT inference settings, accelerometer signal processing, SGR optimization settings, and the evaluation scope, ensuring reproducibility.

\vspace{4pt}
\paragraph{Temporal resolution alignment}
Because the video recordings were captured at 60~fps whereas the accelerometers sampled at 64~Hz, a unified temporal resolution is required for fair comparison. 
Considering the trade-off between VGGT inference cost and reconstruction accuracy, the original 60~fps videos were temporally downsampled to 30~fps before point tracking. Inference at 60~fps would require substantially higher GPU memory and longer computation time, whereas 30~fps provides a practical and efficient compromise.
To maintain consistency, displacement time series were first obtained by integrating the accelerometer signals and were subsequently resampled to 30~Hz.

\vspace{4pt}
\paragraph{VGGT inference settings}
According to the official VGGT repository, only one model architecture and pretrained weight set are publicly available at the time of our experiments, which we directly use without modification.
For 2D point trajectory estimation, the tracked point is manually selected in the first frame. 
A $294 \times 294$ patch approximately centered on this point is cropped and placed at the top-left corner of a $294 \times 518$ canvas, with the remaining pixels filled with zeros. 
This preprocessing ensures compatibility with the VGGT input preprocessing routine and prevents additional resizing, cropping, or padding. 
The processed patches and the corresponding point coordinates are then fed to the VGGT point-tracking branch.
For camera parameter estimation, the first frames from both viewpoints are directly provided at full resolution to the VGGT camera-parameter branch, which outputs the intrinsic and extrinsic camera parameters.

\vspace{4pt}
\paragraph{Accelerometer signal processing}
Tri-axial acceleration signals were converted from $g$ to $m/s^2$, and the motion onset was detected using the rolling standard deviation of the combined acceleration magnitude. 
A Hampel filter (window: $0.75\,\mathrm{s}$, threshold: $3.5$ MAD) was applied to remove impulsive outliers, followed by mean removal and a fourth-order Butterworth band-pass filter ($1$--$10\,\mathrm{Hz}$) applied to the detected motion segment. 
Displacement was obtained by double trapezoidal integration starting from the motion onset with zero initial velocity and displacement. 
To suppress integration drift, the intermediate velocity signal was linearly detrended before the second integration, and the final displacement was expressed in millimeters. 
The resulting displacement time series was subsequently resampled to $30$~Hz to match the temporal resolution used in the visual measurements. 
All acceleration signals were processed using the same parameter settings, without dataset-specific tuning.

\vspace{4pt}
\paragraph{SGR optimization settings}
The weight parameters in Eq.~\ref{eq:sgr_expand} are set to
$w_{Z,\mathrm{abs}}=4.0$, $w_{Z,\mathrm{diff}}=8.0$, 
$w_{XY,\mathrm{abs}}=6.0$, $w_{XY,\mathrm{diff}}=12.0$, 
and $w_{2d}=0.05$. 
Residual terms are normalized by their characteristic motion scales so that the weights control the relative influence of the constraint terms. 
The parameter configuration was determined through controlled experiments on the Data-1 sequence at the mid-span location (1/2 span), where multiple weight combinations were examined. 
The selected configuration was subsequently applied to the remaining five monitoring locations, where consistent improvements were observed across all locations. 
The refinement behavior remains stable under moderate variations of these parameters.

\vspace{4pt}
\paragraph{Evaluation scope}
Under normal service conditions, including pedestrian-induced excitation, the longitudinal displacement of the monitored bridge is theoretically negligible. Consequently, the experimental evaluation focuses on the lateral (X) and vertical (Y) components, where both the structural response and the reference signals are sufficiently informative. 
This choice is consistent with most existing studies on 3D displacement estimation. In addition, comprehensive validation along all three spatial directions is rarely feasible for in-service infrastructures under normal operating conditions. Accordingly, following prior work (e.g., \citet{zhang2024automated}), the evaluation in this study concentrates on displacement components that are expected to exhibit pronounced motion in real-world structures.

\subsection{Quantitative results}

This section presents the results of four sets of comparative experiments. First, we compare the proposed framework with a traditional monocular target-tracking baseline following Ji et al.~\cite{ji2020vision} and Kromanis and Kripakaran~\cite{kromanis2021multiple}, which serves as the main comparison with existing vision-based displacement measurement practice.
Second, we evaluate the impact of the structural geometry refinement (SGR) module by comparing the performance of the proposed framework with and without this component (see Table~\ref{tab:disp_quantities} and Table~\ref{tab:metric_results}). In addition, we adopt a rigorous component-level evaluation strategy to further validate the proposed framework. Specifically, the two core modules of our framework, the VGGT-based camera parameter estimation module and the VGGT-based point tracking module, are individually replaced with representative state-of-the-art methods from their respective domains. The camera parameter estimation module is evaluated against PoseDiffusion~\cite{wang2023posediffusion}, with quantitative results reported in Table~\ref{tab:ablation_camera_module}. Similarly, the point tracking module is benchmarked against BootsTAPIR~\cite{doersch2024bootstap}, with performance comparisons presented in Table~\ref{tab:ablation_tracking_module}. This modular evaluation helps elucidate the role of each component and their respective contributions to the overall 3D displacement reconstruction performance.

For quantitative comparison, we implemented a representative target-tracking baseline following the vision-based measurement pipelines in \citet{ji2020vision} and \citet{kromanis2021multiple}. 
These methods represent a well-established traditional pipeline for monocular planar displacement measurement, where structural targets are tracked within predefined regions of interest (ROIs) in consecutive image frames to a resolution of 1/16 pixel, and the resulting pixel-domain displacement histories are converted into engineering units using a scale factor.
Other recent methods, including several 3D vision-based displacement measurement approaches, could not be directly evaluated in our setting because their implementations are not publicly available, they require task-specific model training, or they rely on additional on-site preparation. Therefore, we compare our method with this widely adopted traditional target-tracking baseline.

In the baseline evaluation, we used only the left-view video of each test, as the baseline relies on a monocular vision system for planar displacement measurement.
The tracked vertical image-plane displacements were converted to millimetres, temporally aligned with the acceleration-derived reference signal, and evaluated along the structural vertical (Y) direction. 
The quantitative comparison is summarized in Table~\ref{tab:traditional_baseline_comparison}, where bold values indicate the better result between the two methods; dashes indicate directions that cannot be measured by the traditional monocular baseline.

\begin{table}[t]
\centering
\fontsize{8pt}{8pt}\selectfont
\caption{Quantitative comparison with a traditional monocular target-tracking baseline. Our method reports both lateral (X) and vertical (Y) bridge displacement components, while the baseline provides only vertical displacement estimates. All metrics are computed with respect to accelerometer-derived displacement references: range-normalized RMSE (NRMSE\textsubscript{range}, $\downarrow$), correlation coefficient ($\uparrow$), and relative peak-to-peak amplitude error (RPPAE, $\downarrow$). \textbf{Bold} numbers indicate better performance when both methods are available.}
\label{tab:traditional_baseline_comparison}
\resizebox{\textwidth}{!}{
\begin{tabular}{@{}llccccccc@{}}
\toprule
\multirow{2}{*}{Dataset} & \multirow{2}{*}{Location} & \multirow{2}{*}{Axis} &
\multicolumn{2}{c}{NRMSE\textsubscript{range} ($\downarrow$)} &
\multicolumn{2}{c}{Correlation coefficient ($\uparrow$)} &
\multicolumn{2}{c}{RPPAE ($\downarrow$)} \\
\cmidrule(lr){4-5}\cmidrule(lr){6-7}\cmidrule(lr){8-9}
 &  &  & Ours & \makecell{Target tracking\\\cite{ji2020vision,kromanis2021multiple}}
 & Ours & \makecell{Target tracking\\\cite{ji2020vision,kromanis2021multiple}}
 & Ours & \makecell{Target tracking\\\cite{ji2020vision,kromanis2021multiple}} \\
\midrule
\multirow{4}{*}{Data-1}
 & 1/2 span & Y & \textbf{0.15} & 0.17 & \textbf{0.66} & 0.52 & 0.14 & \textbf{0.05} \\
 &          & X & \textbf{0.15} & --   & \textbf{0.77} & --   & \textbf{0.04} & -- \\
 & 3/4 span & Y & \textbf{0.17} & 0.20 & \textbf{0.51} & 0.39 & \textbf{0.08} & 0.18 \\
 &          & X & \textbf{0.18} & --   & \textbf{0.58} & --   & \textbf{0.07} & -- \\
\midrule
\multirow{4}{*}{Data-2}
 & 1/4 span & Y & \textbf{0.11} & 0.18 & \textbf{0.88} & 0.62 & \textbf{0.03} & 0.07 \\
 &          & X & \textbf{0.15} & --   & \textbf{0.76} & --   & \textbf{0.03} & -- \\
 & 1/2 span & Y & \textbf{0.11} & 0.16 & \textbf{0.86} & 0.64 & \textbf{0.01} & 0.45 \\
 &          & X & \textbf{0.12} & --   & \textbf{0.88} & --   & \textbf{0.02} & -- \\
\midrule
\multirow{4}{*}{Data-3}
 & 1/4 span & Y & \textbf{0.07} & 0.09 & \textbf{0.93} & 0.88 & 0.03 & \textbf{0.02} \\
 &          & X & \textbf{0.19} & --   & \textbf{0.53} & --   & \textbf{0.07} & -- \\
 & 1/2 span & Y & \textbf{0.06} & 0.10 & \textbf{0.95} & 0.87 & \textbf{0.10} & 0.37 \\
 &          & X & \textbf{0.30} & --   & \textbf{0.40} & --   & \textbf{0.71} & -- \\
\bottomrule
\end{tabular}
}
\end{table}

Overall, our method provides a broader measurement capability by recovering both X and Y displacement components, while the traditional baseline is limited to the vertical component in this comparison. For the common Y direction, our method outperforms the baseline on most metrics and target locations. In particular, it consistently achieves lower $\mathrm{NRMSE}_{\mathrm{range}}$ and higher correlation coefficients, indicating more accurate pointwise displacement estimates and better preservation of the temporal vibration waveform. The only two cases where our method is not the best are the RPPAE values at the Data-1 1/2-span and Data-3 1/4-span locations. This behaviour is plausible because RPPAE depends only on the global maximum and minimum of the recovered displacement history and is therefore more sensitive to small errors at isolated extrema than the other two metrics. In our pipeline, the displacement is estimated from two views and depends on both tracking results and VGGT-inferred camera parameters; small point-tracking errors from the two views, together with camera-parameter inference uncertainty, can propagate to the reconstructed vertical displacement and slightly over- or under-estimate the peak-to-peak amplitude. By contrast, the monocular planar baseline directly scales a single image-plane displacement, which may occasionally preserve the global amplitude more closely even when its waveform agreement and pointwise accuracy are worse. Thus, the comparison suggests that our method provides stronger overall displacement reconstruction accuracy and additional directional observability, while peak-amplitude estimation remains sensitive to the combined effects of multi-view tracking and camera-parameter uncertainty.

\begin{table}[t]
\centering
\fontsize{8pt}{8pt}\selectfont
\caption{Displacement-derived global peak-to-peak amplitudes obtained from our proposed framework (with and without structural geometry refinement, SGR) and accelerometer-derived displacement references (acc). These quantities provide descriptive information about the measured vibration response.}
\label{tab:disp_quantities}
\resizebox{\textwidth}{!}{
\begin{tabular}{@{}llccccc@{}}
\toprule
\multirow{2}{*}{Dataset} & \multirow{2}{*}{Location} & \multirow{2}{*}{Axis} &
\multicolumn{3}{c}{Global peak-to-peak amplitude (mm)} \\
\cmidrule(lr){4-6}
 &  &  & Ours (w/o SGR) & Ours (w/ SGR) & Acc \\
\midrule
\multirow{4}{*}{Data-1}
 & 1/2 span & Y & 13.71 & 13.66 & 11.94 \\
 &           & X &  8.66 &  9.34 &  9.70 \\
 & 3/4 span & Y &  9.19 &  8.97 &  8.30 \\
 &           & X &  9.94 &  7.27 &  7.79 \\
\midrule
\multirow{4}{*}{Data-2}
 & 1/4 span & Y & 12.57 & 12.49 & 12.15 \\
 &           & X &  4.71 &  4.18 &  4.30 \\
 & 1/2 span & Y & 17.50 & 17.46 & 17.37 \\
 &           & X &  8.63 &  7.42 &  7.28 \\
\midrule
\multirow{4}{*}{Data-3}
 & 1/4 span & Y &  7.78 &  7.79 &  7.54 \\
 &           & X &  1.97 &  1.65 &  1.54 \\
 & 1/2 span & Y & 11.98 & 12.06 & 10.95 \\
 &           & X &  4.09 &  3.83 &  2.24 \\
\bottomrule
\end{tabular}
}
\end{table}

Table~\ref{tab:disp_quantities} summarizes the displacement-derived global peak-to-peak amplitudes obtained from our proposed VFM-based framework, with and without structural geometry refinement (SGR), alongside accelerometer-derived displacement references.
The peak-to-peak amplitudes characterize the overall vibration magnitude at each monitored location.
Across three field datasets and both measurement axes, the proposed VFM-based framework, with or without SGR, yields peak-to-peak displacement amplitudes that closely agree with accelerometer-derived references. This agreement is particularly strong in high-amplitude vibration cases, for example in the Data-2 vertical direction, where the estimated amplitude of 17.46 mm closely matches the reference value of 17.37 mm. In contrast, larger discrepancies are observed in low-amplitude scenarios, such as the lateral direction of Data-3, where performance is limited by the low signal-to-noise ratio.

Table~\ref{tab:metric_results} presents a comprehensive quantitative comparison of the proposed VFM-based framework with and without SGR, evaluated against accelerometer-derived displacement references based on three complementary metrics: range-normalized RMSE (NRMSE$_\text{range}$), correlation coefficient, and relative peak-to-peak amplitude error (RPPAE). Across three field datasets and both the X and Y directions, the proposed VFM-based framework shows consistently agreement with the reference.
The mean and standard deviation across all measurement cases are also reported at the bottom of the table to provide an overall comparison of accuracy and stability between the two variants. Overall, our proposed framework with SGR performs slightly better than the variant without SGR and demonstrates improved stability. This improvement is particularly evident in the RPPAE metric, which decreases from 0.18 to 0.11, indicating a notable gain brought by the incorporation of SGR.

\begin{table}[t]
\centering
\fontsize{8pt}{8pt}\selectfont
\caption{Quantitative comparison between the proposed framework with and without structural geometry refinement (SGR). All metrics are computed with respect to accelerometer-derived displacement references: range-normalized RMSE (NRMSE\textsubscript{range}, $\downarrow$), correlation coefficient ($\uparrow$), and relative peak-to-peak amplitude error (RPPAE, $\downarrow$). \textbf{Bold} numbers indicate better performance between the two variants for each entry. The mean and standard deviation are computed across all measurement instances, corresponding to three datasets, two monitored locations per dataset, and two motion directions.}
\label{tab:metric_results}
\resizebox{\textwidth}{!}{
\begin{tabular}{@{}llcccccccc@{}}
\toprule
\multirow{2}{*}{Dataset} & \multirow{2}{*}{Location} & \multirow{2}{*}{Axis} &
\multicolumn{2}{c}{NRMSE\textsubscript{range} ($\downarrow$)} &
\multicolumn{2}{c}{Correlation coefficient ($\uparrow$)} &
\multicolumn{2}{c}{RPPAE ($\downarrow$)} \\
\cmidrule(lr){4-5}\cmidrule(lr){6-7}\cmidrule(lr){8-9}
 &  &  & w/o SGR & w/ SGR & w/o SGR & w/ SGR & w/o SGR & w/ SGR \\
\midrule
\multirow{4}{*}{Data-1}
 & 1/2 span & Y & \textbf{0.15} & \textbf{0.15} & \textbf{0.66} & \textbf{0.66} & 0.15 & \textbf{0.14} \\
 &           & X & \textbf{0.15} & \textbf{0.15} & \textbf{0.78} & 0.77          & 0.11 & \textbf{0.04} \\
 & 3/4 span  & Y & 0.18 & \textbf{0.17} & 0.50          & \textbf{0.51} & 0.11 & \textbf{0.08} \\
 &           & X & 0.22          & \textbf{0.18} & 0.50          & \textbf{0.58} & 0.28 & \textbf{0.07} \\
\midrule
\multirow{4}{*}{Data-2}
 & 1/4 span & Y & 0.12          & \textbf{0.11} & \textbf{0.88} & \textbf{0.88} & \textbf{0.03} & \textbf{0.03} \\
 &          & X & 0.17          & \textbf{0.15} & 0.73          & \textbf{0.76} & 0.10          & \textbf{0.03} \\
 & 1/2 span & Y & \textbf{0.11} & \textbf{0.11} & \textbf{0.86} & \textbf{0.86} & \textbf{0.01} & \textbf{0.01} \\
 &          & X & 0.15          & \textbf{0.12} & 0.86          & \textbf{0.88} & 0.19          & \textbf{0.02} \\
\midrule
\multirow{4}{*}{Data-3}
 & 1/4 span & Y & 0.08          & \textbf{0.07} & \textbf{0.93} & \textbf{0.93} & \textbf{0.03} & \textbf{0.03} \\
 &          & X & 0.23          & \textbf{0.19} & 0.44          & \textbf{0.53} & 0.28          & \textbf{0.07} \\
 & 1/2 span & Y & \textbf{0.06} & \textbf{0.06} & \textbf{0.95} & \textbf{0.95} & \textbf{0.09} & 0.10          \\
 &          & X & 0.31          & \textbf{0.30} & \textbf{0.40} & \textbf{0.40} & 0.82          & \textbf{0.71} \\
\midrule
Mean & -- & -- & 0.16 & 0.15 & 0.71 & 0.73 & 0.18 & 0.11 \\
Std  & -- & -- & 0.063 & 0.063 & 0.192 & 0.176 & 0.210 & 0.184 \\
\bottomrule
\end{tabular}
}
\end{table}

A clear trend is that SGR leads to more notable improvements along the lateral (X) direction than along the vertical (Y) direction.
This behavior is physically intuitive: the bridge exhibits significantly larger vertical motion under human-induced excitation.
Table~\ref{tab:disp_quantities} confirms this disparity; for example, in Data-3, the vertical amplitude (7.54~mm) is nearly five times larger than the lateral amplitude (1.54~mm).  
Such large vertical amplitudes make feature tracking in the Y direction inherently more stable, whereas tracking small-amplitude lateral motion is more susceptible to pixel-level noise.
By enforcing geometric consistency in the reconstructed 3D trajectories, SGR effectively suppresses such noise, leading to lower NRMSE$_{\text{range}}$, higher correlation, and notably reduced amplitude error in the X direction. This effect is further corroborated by the qualitative results in Section~\ref{subsec:qualitative}, where the refined trajectories exhibit visibly improved smoothness and physical plausibility.

We further observe that the quality of the acquired videos has a direct impact on the final displacement reconstruction accuracy. For Data-1, the presence of strong glare in the left-view camera noticeably degraded the tracking performance, especially for the more distant point located at the 3/4 span, leading to reduced measurement accuracy. In Data-2, although the right-view image is slightly blurred, the tracking remained sufficiently stable, resulting in relatively accurate displacement estimates. 
For Data-3, the measurement accuracy is high in the vertical (Y) direction but considerably lower in the lateral (X) direction. This discrepancy arises because the right-view image suffers from blur and low illumination, and the lateral vibration amplitude is inherently small (approximately 2~mm), making the X-direction tracking highly sensitive to noise.
Despite this, Table~\ref{tab:metric_results} shows that SGR still reduced the RPPAE in this challenging case (e.g., from 0.28 to 0.07), preventing the more severe drift observed in the baseline method.

Table~\ref{tab:ablation_camera_module} shows that replacing VGGT-based camera parameter estimation module with PoseDiffusion \cite{wang2023posediffusion} substantially degrades accuracy, with particularly large increases in RPPAE and NRMSE\textsubscript{range}, indicating unstable amplitude recovery. Although PoseDiffusion occasionally achieves slightly higher correlation on the X-axis at a few locations, these gains do not translate into improved error metrics, suggesting that the predicted signals may follow the overall trend while exhibiting pronounced scale or amplitude bias. Overall, the results validate the effectiveness and robustness of the VGGT-based camera module within the proposed framework.
The mean and standard deviation across all measurement cases, reported at the bottom of the table, further highlight the overall superiority and stability of the VGGT-based implementation.

Table~\ref{tab:ablation_tracking_module} shows that the VGGT-based tracker and BootsTAPIR \cite{doersch2024bootstap} achieve broadly comparable performance across all datasets, locations, and axes, with no consistent dominance across the three evaluation metrics. In some cases, BootsTAPIR attains slightly lower NRMSE\textsubscript{range} or RPPAE and marginally higher correlation, while in others the VGGT-based tracker performs better. Overall, these results indicate that both tracking strategies can provide reliable point trajectories within the proposed framework.
The mean and standard deviation across all cases, reported at the bottom of the table, further confirm the comparable overall performance of the two tracking strategies.

\begin{table}[t]
\centering
\fontsize{8pt}{8pt}\selectfont
\caption{Quantitative comparison of camera parameter estimation strategies within the proposed framework. The default implementation of the proposed framework is VGGT-based and is compared with an alternative PoseDiffusion-based implementation \cite{wang2023posediffusion}. All metrics are computed with respect to accelerometer-derived displacement references: range-normalized RMSE (NRMSE\textsubscript{range}, $\downarrow$), correlation coefficient ($\uparrow$), and relative peak-to-peak amplitude error (RPPAE, $\downarrow$). \textbf{Bold} numbers indicate better performance between the two variants for each entry. The mean and standard deviation are computed across all measurement instances, corresponding to three datasets, two monitored locations per dataset, and two motion directions.}
\label{tab:ablation_camera_module}
\resizebox{\textwidth}{!}{
\begin{tabular}{@{}llcccccccc@{}}
\toprule
\multirow{2}{*}{Dataset} & \multirow{2}{*}{Location} & \multirow{2}{*}{Axis} &
\multicolumn{2}{c}{NRMSE\textsubscript{range} ($\downarrow$)} &
\multicolumn{2}{c}{Correlation coefficient ($\uparrow$)} &
\multicolumn{2}{c}{RPPAE ($\downarrow$)} \\
\cmidrule(lr){4-5}\cmidrule(lr){6-7}\cmidrule(lr){8-9}
 &  &  & VGGT-based & PoseDiffusion-based & VGGT-based & PoseDiffusion-based & VGGT-based & PoseDiffusion-based \\
\midrule
\multirow{4}{*}{Data-1}
 & 1/2 span & Y & \textbf{0.15} & 0.32 & \textbf{0.66} & 0.38 & \textbf{0.15} & 1.03 \\
 &           & X & \textbf{0.15} & 0.26 & 0.78 & \textbf{0.80} & \textbf{0.11} & 1.01 \\
 & 3/4 span  & Y & \textbf{0.18} & 0.35 & \textbf{0.50} & 0.19 & \textbf{0.11} & 1.15 \\
 &           & X & \textbf{0.22} & 0.26 & 0.50 & \textbf{0.66} & \textbf{0.28} & 0.79 \\
\midrule
\multirow{4}{*}{Data-2}
 & 1/4 span & Y & \textbf{0.12} & 0.74 & \textbf{0.88} & 0.62 & \textbf{0.03} & 2.78 \\
 &          & X & \textbf{0.17} & 1.98 & \textbf{0.73} & 0.23 & \textbf{0.10} & 7.79 \\
 & 1/2 span & Y & \textbf{0.11} & 0.26 & \textbf{0.86} & 0.53 & \textbf{0.01} & 0.51 \\
 &          & X & \textbf{0.15} & 0.37 & \textbf{0.86} & 0.51 & \textbf{0.19} & 0.81 \\
\midrule
\multirow{4}{*}{Data-3}
 & 1/4 span & Y & \textbf{0.08} & 0.49 & \textbf{0.93} & 0.82 & \textbf{0.03} & 2.21 \\
 &          & X & \textbf{0.23} & 1.21 & 0.44 & \textbf{0.46} & \textbf{0.28} & 5.63 \\
 & 1/2 span & Y & \textbf{0.06} & 1.55 & \textbf{0.95} & 0.85 & \textbf{0.09} & 8.85 \\
 &          & X & \textbf{0.31} & 3.37 & \textbf{0.40} & 0.33 & \textbf{0.82} & 16.41 \\
\midrule
Mean & -- & -- & 0.16 & 0.93 & 0.71 & 0.53 & 0.18 & 4.08 \\
Std  & -- & -- & 0.063 & 0.918 & 0.192 & 0.217 & 0.210 & 4.634 \\
\bottomrule
\end{tabular}
}
\end{table}

\begin{table}[t]
\centering
\fontsize{8pt}{8pt}\selectfont
\caption{Quantitative comparison of point tracking strategies within the proposed framework. The default implementation of the proposed framework is VGGT-based and is compared with an alternative BootsTAPIR-based implementation \cite{doersch2024bootstap}. All metrics are computed with respect to accelerometer-derived displacement references: range-normalized RMSE (NRMSE\textsubscript{range}, $\downarrow$), correlation coefficient ($\uparrow$), and relative peak-to-peak amplitude error (RPPAE, $\downarrow$). \textbf{Bold} numbers indicate better performance between the two variants for each entry. The mean and standard deviation are computed across all measurement instances, corresponding to three datasets, two monitored locations per dataset, and two motion directions.}
\label{tab:ablation_tracking_module}
\resizebox{\textwidth}{!}{
\begin{tabular}{@{}llcccccccc@{}}
\toprule
\multirow{2}{*}{Dataset} & \multirow{2}{*}{Location} & \multirow{2}{*}{Axis} &
\multicolumn{2}{c}{NRMSE\textsubscript{range} ($\downarrow$)} &
\multicolumn{2}{c}{Correlation coefficient ($\uparrow$)} &
\multicolumn{2}{c}{RPPAE ($\downarrow$)} \\
\cmidrule(lr){4-5}\cmidrule(lr){6-7}\cmidrule(lr){8-9}
 &  &  & VGGT-based & BootsTAPIR-based & VGGT-based & BootsTAPIR-based & VGGT-based & BootsTAPIR-based \\
\midrule
\multirow{4}{*}{Data-1}
 & 1/2 span & Y & \textbf{0.15} & 0.17 & \textbf{0.66} & 0.58 & 0.15 & \textbf{0.12} \\
 &           & X & 0.15 & \textbf{0.13} & 0.78 & \textbf{0.83} & \textbf{0.11} & 0.18 \\
 & 3/4 span  & Y & \textbf{0.18} & 0.19 & \textbf{0.50} & 0.36 & 0.11 & \textbf{0.01} \\
 &           & X & 0.22 & \textbf{0.17} & 0.50 & \textbf{0.63} & 0.28 & \textbf{0.04} \\
\midrule
\multirow{4}{*}{Data-2}
 & 1/4 span & Y & 0.12 & \textbf{0.11} & \textbf{0.88} & 0.87 & 0.03 & \textbf{0.00} \\
 &          & X & \textbf{0.17} & \textbf{0.17} & \textbf{0.73} & \textbf{0.73} & 0.10 & \textbf{0.08} \\
 & 1/2 span & Y & \textbf{0.11} & 0.13 & \textbf{0.86} & 0.84 & \textbf{0.01} & 0.06 \\
 &          & X & 0.15 & \textbf{0.12} & 0.86 & \textbf{0.87} & 0.19 & \textbf{0.09} \\
\midrule
\multirow{4}{*}{Data-3}
 & 1/4 span & Y & \textbf{0.08} & 0.12 & \textbf{0.93} & 0.91 & \textbf{0.03} & 0.05 \\
 &          & X & 0.23 & \textbf{0.20} & 0.44 & \textbf{0.49} & 0.28 & \textbf{0.24} \\
 & 1/2 span & Y & \textbf{0.06} & 0.07 & \textbf{0.95} & 0.94 & \textbf{0.09} & 0.10 \\
 &          & X & 0.31 & \textbf{0.29} & \textbf{0.40} & \textbf{0.40} & \textbf{0.82} & 0.85 \\
 \midrule
Mean & -- & -- & 0.16 & 0.16 & 0.71 & 0.70 & 0.18 & 0.15 \\
Std  & -- & -- & 0.063 & 0.055 & 0.192 & 0.184 & 0.210 & 0.239 \\
\bottomrule
\end{tabular}
}
\end{table}

Based on the two comparative experiments reported in Table~\ref{tab:ablation_camera_module} and Table~\ref{tab:ablation_tracking_module}, we can further infer that the inferior performance of PoseDiffusion on our dataset is primarily attributable to its training on a single dataset for camera parameter estimation. In contrast, VGGT is a vision foundation model trained on multiple datasets and optimized for multiple vision tasks, which endows it with substantially stronger generalization ability and robustness when applied to unseen real-world scenarios. Meanwhile, although BootsTAPIR is also a task-specific deep learning model focusing on point tracking, its training on large-scale, multi-scene datasets enables performance that is comparable to VGGT for this particular task. Compared with PoseDiffusion and BootsTAPIR, VGGT provides stable predictions across multiple tasks within a unified model, thereby simplifying the overall system configuration and processing pipeline for vision-based displacement measurement and aligning naturally with the design philosophy of our proposed framework.

\subsection{Qualitative results}
\label{subsec:qualitative}
To complement the quantitative evaluation, we present representative qualitative results using the 1/4-span location from Data-2. Although six visualization sets are available (three datasets and two monitoring points each), only one is shown here for brevity. As illustrated in Fig.~\ref{fig:data2_stereo}, both the 1/4-span and 1/2-span points lie within the fields of view of the two cameras. 

\begin{figure}[t]
  \centering
  \begin{minipage}{0.48\linewidth}
    \centering
    \includegraphics[width=\linewidth]{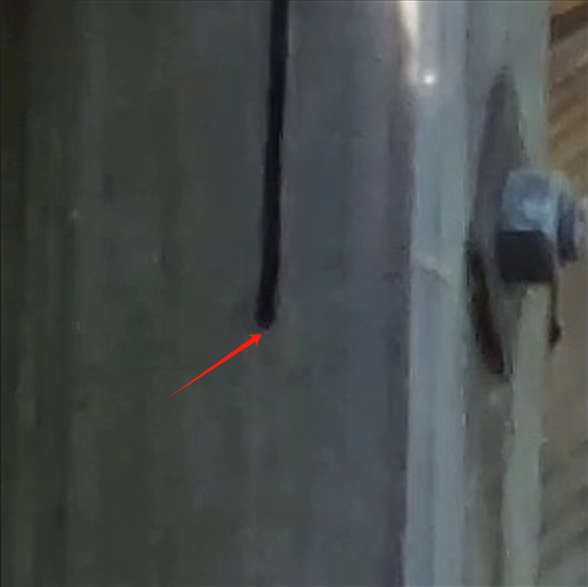}
  \end{minipage}\hfill
  \begin{minipage}{0.48\linewidth}
    \centering
    \includegraphics[width=\linewidth]{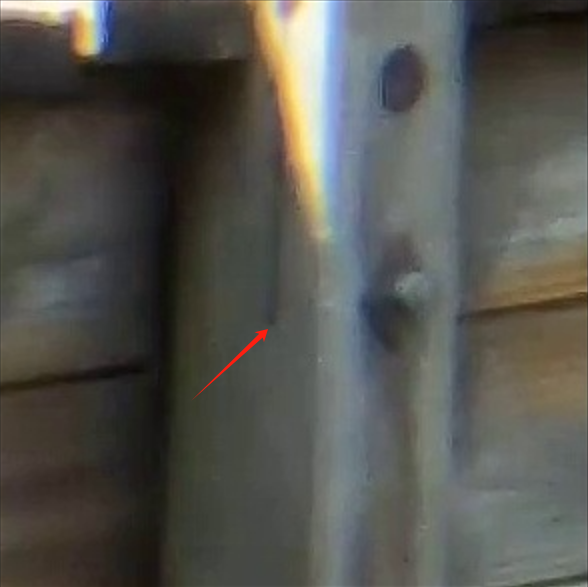}
  \end{minipage}
  \caption{The first ROIs at the 1/4-span location of the bridge in Video-1 (left) and Video-2 (right) from Data-2. The red arrow indicates the tracked point.}
  \label{fig:roi}
\end{figure}

As illustrated in Fig.~\ref{fig: processing} of the Sec.~\ref{sec:method}, we manually select a distinctive natural feature within the first region of interest (ROI) of the Video-1, such as a bolt, a sharp texture corner, or another structurally meaningful feature point as the tracking target.
As shown in Fig.~\ref{fig:roi}, a point was manually selected in the first ROI of Video-1, and its corresponding point in the first ROI of Video-2 was identified using VGGT-Tracking.
This strategy enables fully artificial marker-free displacement measurement.

\begin{figure}[H]
  \centering

  \begin{minipage}{0.9\linewidth}  
    \centering
    \includegraphics[width=\linewidth]{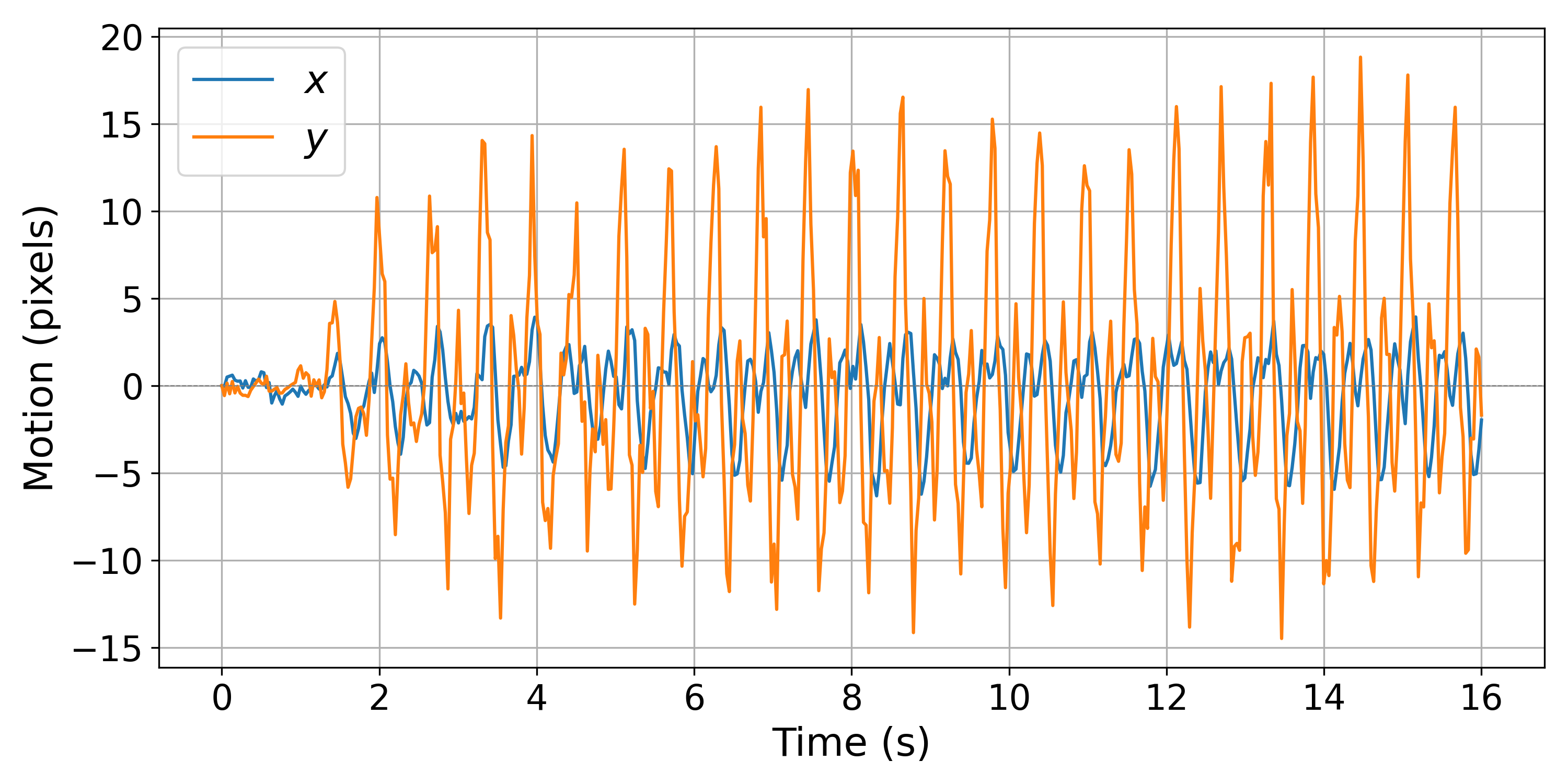}
    \caption*{(a)}
  \end{minipage}


  \begin{minipage}{0.9\linewidth}
    \centering
    \includegraphics[width=\linewidth]{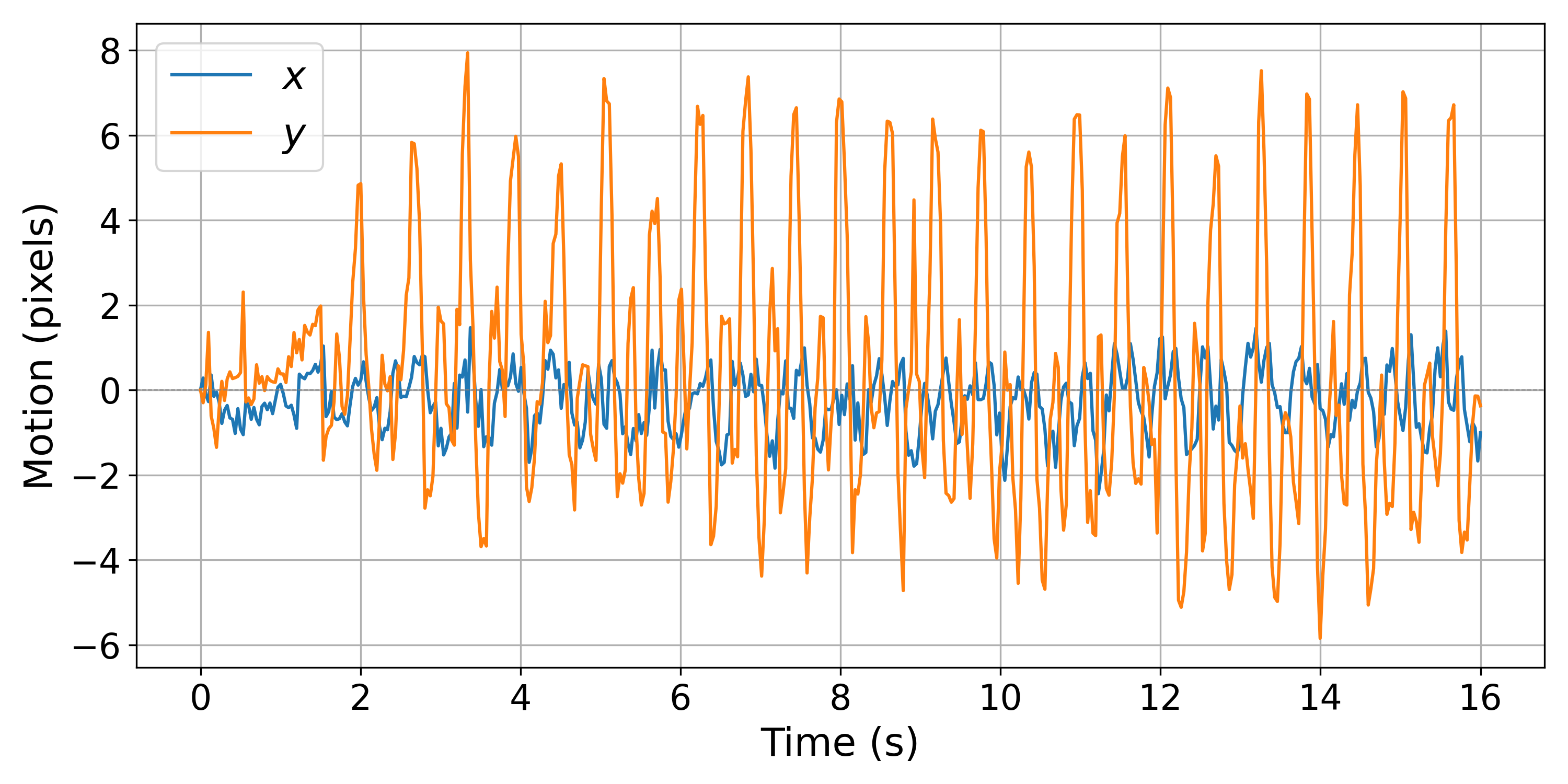}
    \caption*{(b)}
  \end{minipage}

  \caption{
  VGGT-based tracking results of the 1/4-span point in the stereo video pair (Data-2) in the pixel coordinate system, showing tracking in (a) the first video and (b) the second video, where positive $y$ values correspond to downward motion following the standard image coordinate convention.}
  \label{fig:tracking_pair_2}
\end{figure}

Fig.~\ref{fig:tracking_pair_2} displays the tracking trajectories obtained from the two videos using VGGT-based tracking. The first view exhibits highly stable trajectories, whereas the second view contains noticeable noise, particularly along the $x$-axis in the pixel coordinate system. This is attributed to the slightly blurred appearance of the second video and the inherently small $x$-axis vibration amplitude, which makes point tracking more susceptible to noise.

\begin{figure}[H]
  \centering

  \begin{minipage}{0.9\linewidth}
    \centering
    \includegraphics[width=\linewidth]{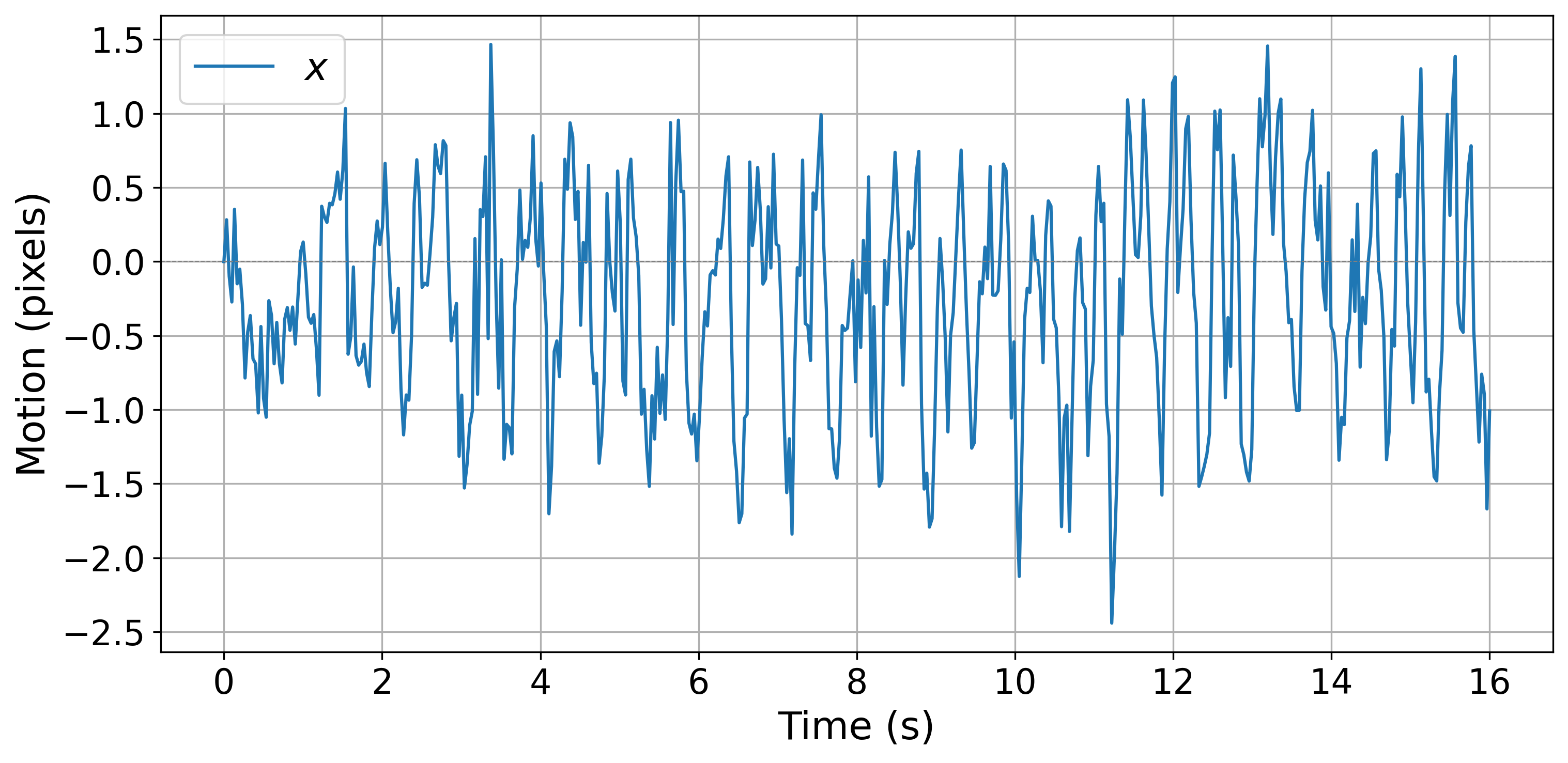}
    \caption*{(a)}
  \end{minipage}


  \begin{minipage}{0.9\linewidth}
    \centering
    \includegraphics[width=\linewidth]{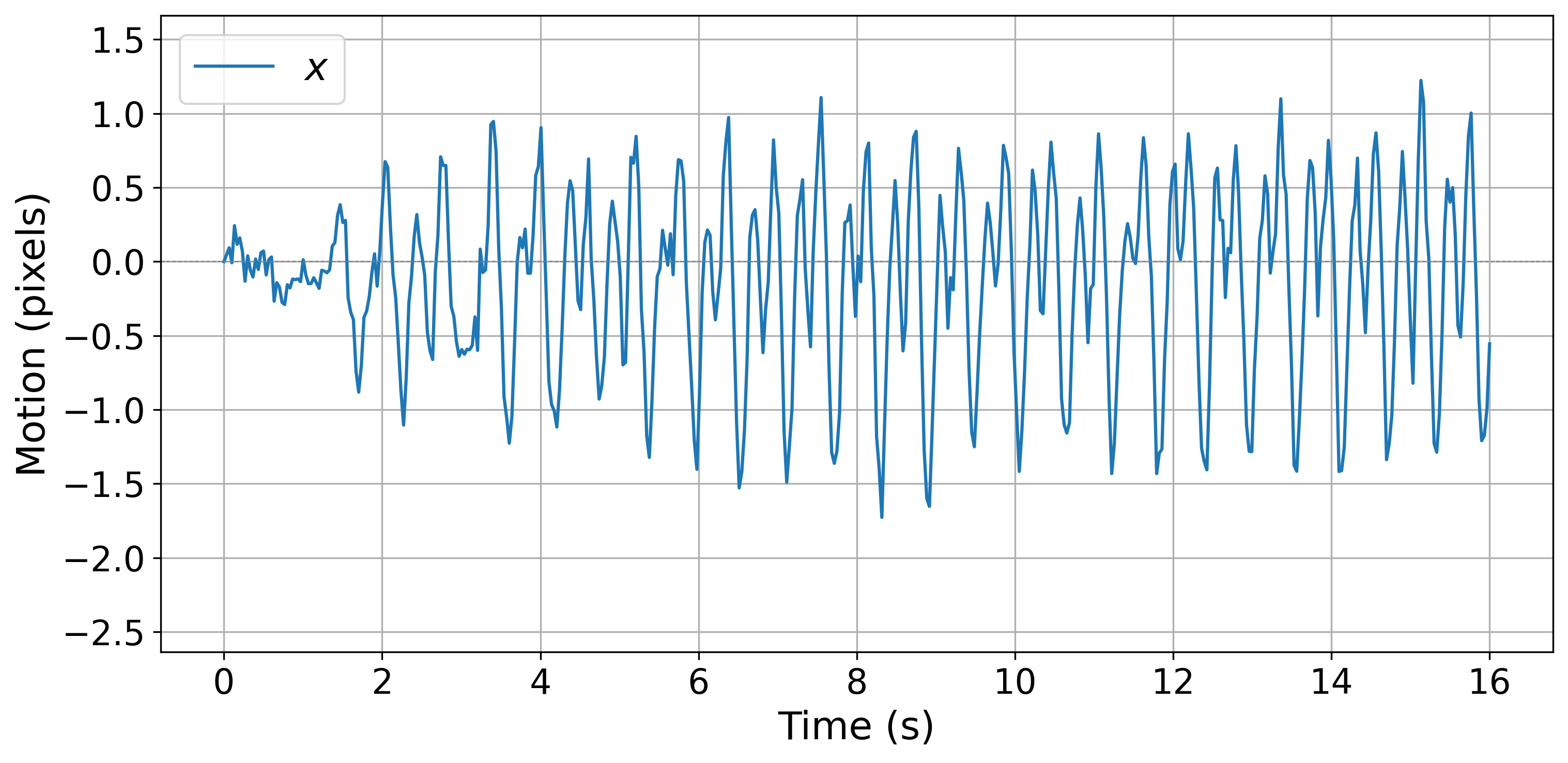}
    \caption*{(b)}
  \end{minipage}

  \caption{
  VGGT-based tracking results of the 1/4-span point in the second video (Data-2) along the $x$-axis in the pixel coordinate system, shown (a) before SGR and (b) after SGR.}
  \label{fig:tracking_compare_2}
\end{figure}

\begin{figure}[H]
  \centering

  \begin{minipage}{1\linewidth}
    \centering
    \includegraphics[width=\linewidth]{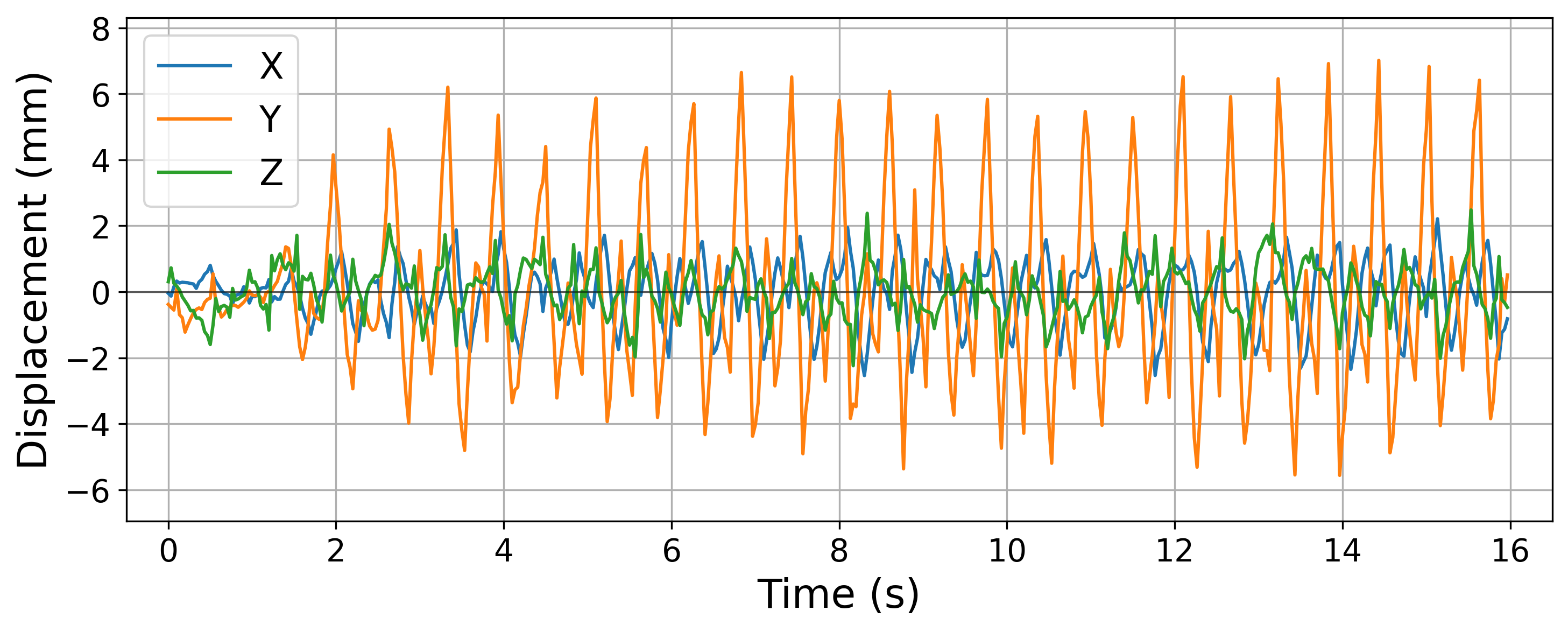}
    \caption*{(a)}
  \end{minipage}


  \begin{minipage}{1\linewidth}
    \centering
    \includegraphics[width=\linewidth]{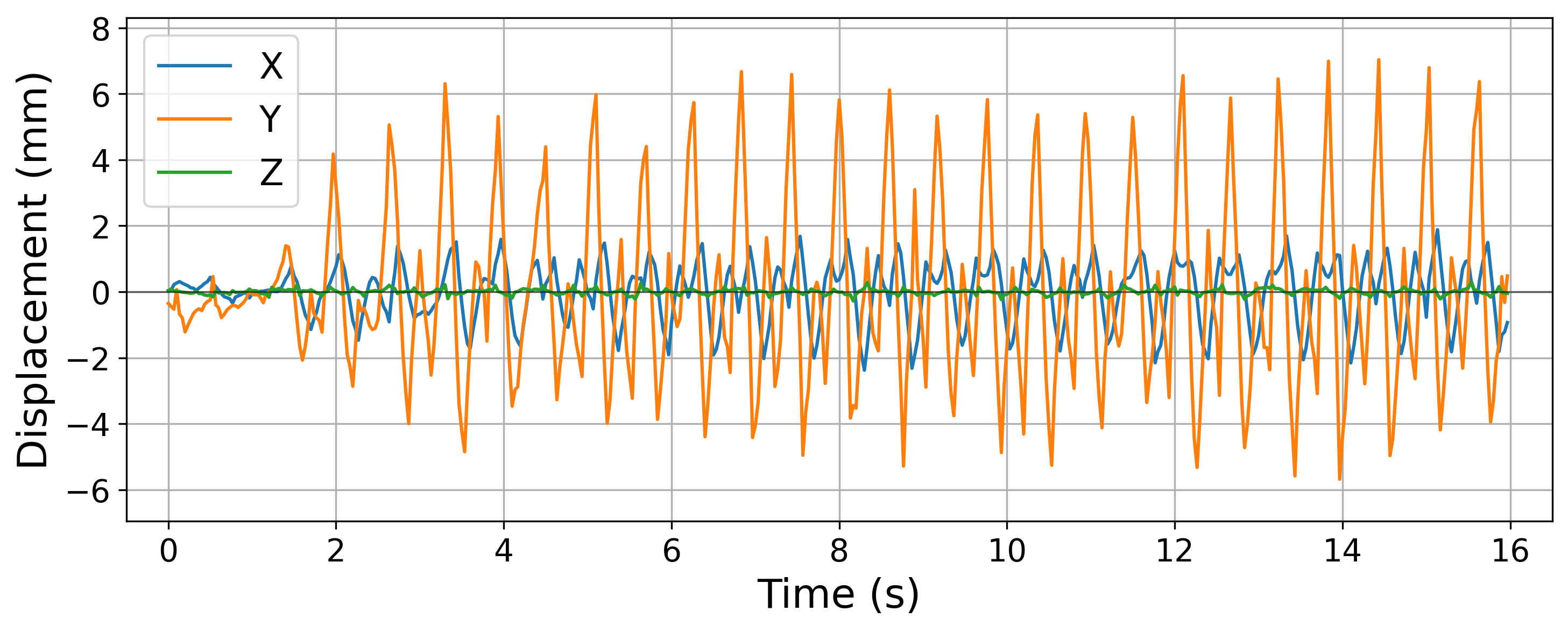}
    \caption*{(b)}
  \end{minipage}


  \caption{
  3D displacement measurement at the 1/4-span location (Data-2) in the structural coordinate system, shown (a) before SGR and (b) after SGR within the proposed VFM-based framework, where X, Y, and Z denote the lateral, vertical, and longitudinal directions of the bridge, respectively, and positive Y values correspond to downward motion.}
  \label{fig:3d_compare_2}
\end{figure}

Fig.~\ref{fig:tracking_compare_2} only compares the pixel-domain tracking trajectories along the $x$-axis from the second video before and after applying structural geometry refinement (SGR), as the $y$-axis trajectories are not changed in the SGR process.
The refined trajectory exhibits a clear reduction in noise, which is consistent with the quantitative improvements reported in Table~\ref{tab:metric_results}.
Finally, Fig.~\ref{fig:3d_compare_2} illustrates the reconstructed 3D displacement time series obtained by the proposed framework, with and without SGR. The SGR effectively suppresses noise in the longitudinal (Z) and lateral (X) directions of the bridge.

\begin{figure}[t]
  \centering
  \includegraphics[width=1\linewidth]{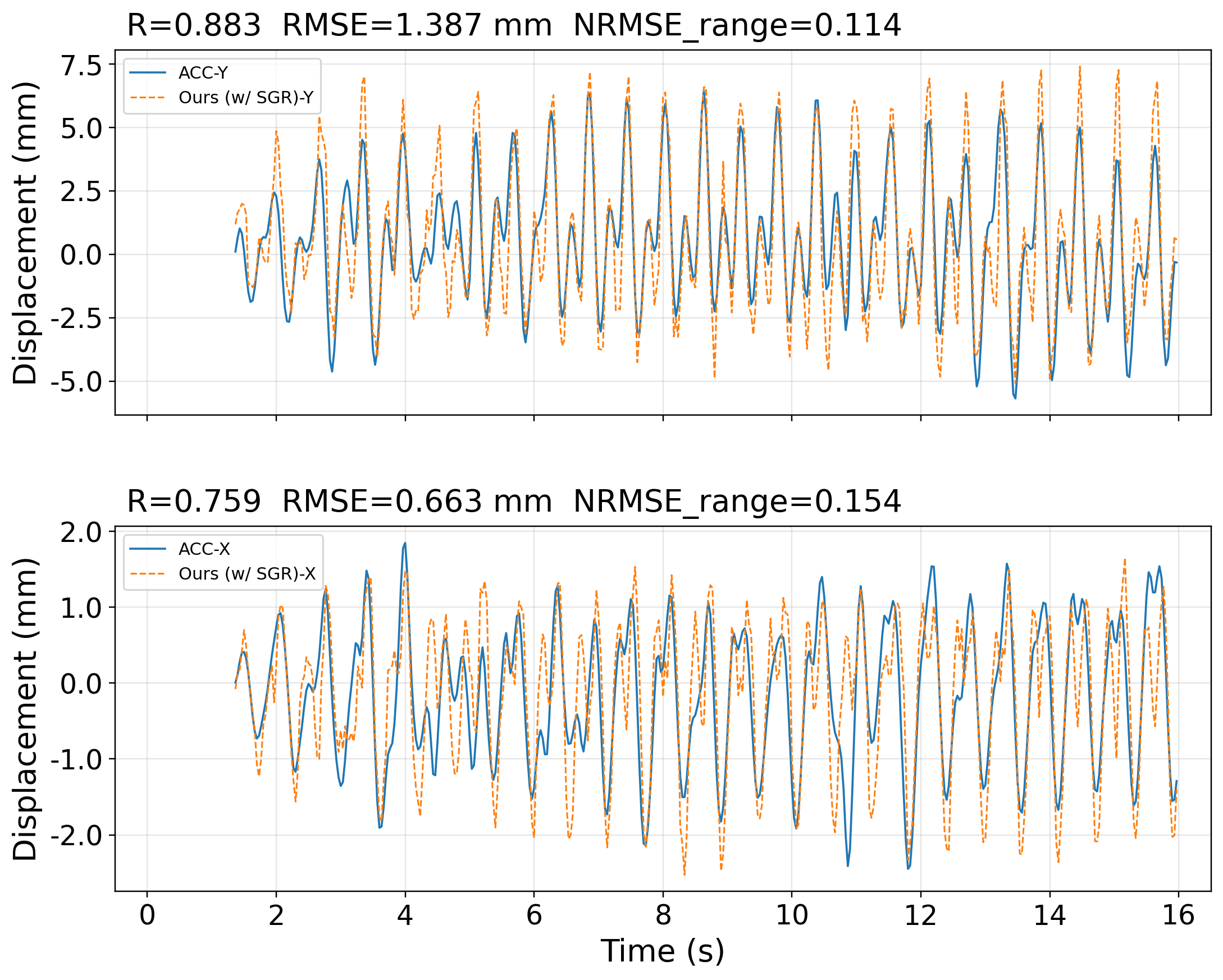}
  \caption{
  Comparison between accelerometer-derived displacement measurements and estimates from the proposed VFM-based framework at the 1/4-span location (Data-2) in the structural coordinate system. The upper plot shows the vertical (Y) displacement, while the lower plot shows the lateral (X) displacement, with positive Y values corresponding to downward motion.}
  \label{fig: acc_vision_2}
\end{figure}

Fig.~\ref{fig: acc_vision_2} compares the accelerometer-derived displacement reference with the displacement estimates obtained by the proposed VFM-based framework with SGR at the 1/4-span location in Data-2. The vision-derived responses closely match the vibration patterns and amplitudes in both the lateral (X) and vertical (Y) directions, with minor deviations mainly caused by video quality variations and residual tracking jitter. These observations are consistent with the quantitative results reported in Table~\ref{tab:metric_results}. 
It should be noted that the accelerometer-derived displacement is obtained by integrating the acceleration signal starting from the detected onset of structural vibration; therefore, since the time axis starts from the beginning of the video recording, a short initial interval without displacement values appears in the figure. For consistency, the displacement estimates from the proposed VFM-based framework are also shown only from this detected vibration onset onward.

\begin{figure}[h]
  \centering
  \includegraphics[width=1\linewidth]{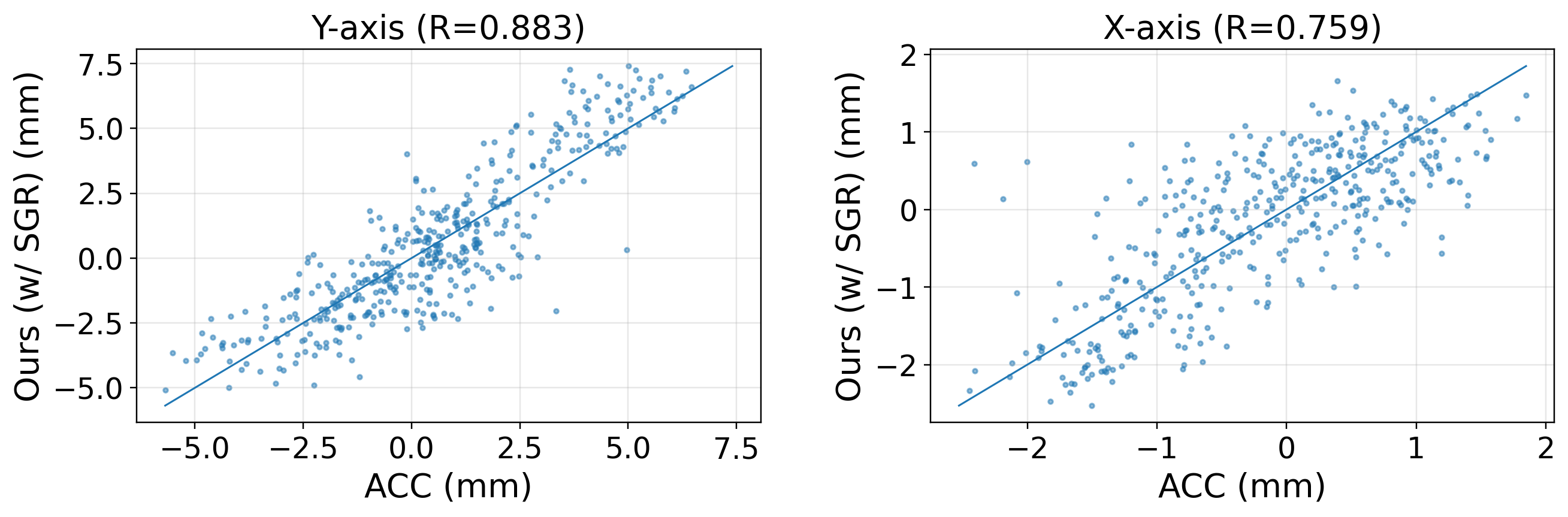}
  \caption{
  Correlation between accelerometer-derived displacement measurements and estimates from the proposed VFM-based framework at the 1/4-span location (Data-2) in the structural coordinate system, where the left plot shows the vertical (Y) direction and the right plot shows the lateral (X) direction.}
  \label{fig: parity_2}
\end{figure}

Fig.~\ref{fig: parity_2} further assesses measurement consistency using scatter plots of accelerometer- and vision-derived displacements. Clear linear relationships are observed along both axes, accompanied by high correlation coefficients (e.g., $R=0.883$ in the vertical (Y) direction), indicating that the proposed framework reliably captures the temporal evolution of the bridge response.

The consistency in both temporal alignment and amplitude demonstrates the validity and robustness of the proposed VFM-based framework under real-world conditions.

\subsection{Computational cost analysis}
We evaluate the computational cost of the core modules in the proposed pipeline, as summarized in Table~\ref{tab:runtime_full}. 
The camera parameter estimation module operates on a two-frame input and introduces negligible overhead when executed on the GPU.
In contrast, point tracking over long image sequences dominates GPU usage, due to the large number of frames and the reliance on dense feature extraction and matching within a streaming framework. In our implementation, tracking is performed on a 480-frame sequence using a streaming strategy, where overlapping chunks (length: 144, overlap: 32) are processed to reduce memory usage while maintaining temporal continuity.
The geometric triangulation itself is highly efficient, requiring only a few milliseconds for a 480-frame sequence when executed on the CPU. 
The triangulation+SGR module is significantly slower due to the iterative nature of the optimization and is also performed on the CPU. 
Specifically, the SGR step relies on a non-linear least-squares solver, which repeatedly evaluates the objective function and performs triangulation over the full sequence at each iteration, leading to increased runtime. 
It is worth noting that the proposed framework is not designed for real-time processing; instead, all data are processed offline after acquisition for accurate analysis.

\begin{table}[h]
\fontsize{8pt}{8pt}\selectfont
\centering
\caption{
Computational cost of the proposed pipeline, including VGGT inference on a GPU and geometric processing on a CPU. 
VGGT inference is performed on an NVIDIA L40 GPU. Camera parameter estimation uses a two-frame input, and point tracking is conducted on a 480-frame sequence with a streaming strategy (chunk length: 144, overlap: 32). 
Triangulation and triangulation with SGR are executed on an Intel i7-12700H CPU. 
Reported runtimes correspond to forward processing only.
}
\label{tab:runtime_full}
\resizebox{\textwidth}{!}{
\begin{tabular}{@{}lcccccc@{}}
\toprule
\textbf{Task} & \textbf{Device} & \textbf{Input Size} & \textbf{Calls} & \textbf{Time (per call)} & \textbf{Total Time} & \textbf{Peak Memory} \\
\midrule

Camera parameter estimation 
& GPU 
& 2 frames 
& 1 
& 177.6 ms 
& 177.6 ms 
& 7.1 GB \\

\midrule

Point tracking (streaming) 
& GPU 
& 144 frames, 1 point 
& 4 
& 23.2 s 
& 93.3 s 
& 41.2 GB \\

\midrule

Triangulation 
& CPU 
& 480 frames, 1 point 
& 1 
& 3.2 ms 
& 3.2 ms 
& 42.5 MB \\

\midrule

Triangulation + SGR 
& CPU 
& 480 frames, 1 point 
& 1 
& 13.2 s 
& 13.2 s 
& 136.0 MB \\

\bottomrule
\end{tabular}
}
\end{table}

\subsection{Sensitivity analysis}
To better interpret the experimental results and evaluate the robustness of the proposed framework, we first examine the characteristics of the three collected vibration sequences. Based on these differences, we further analyze the sensitivity of the proposed framework to two key factors affecting displacement reconstruction accuracy and discuss the role of structural geometry refinement (SGR).

\vspace{4pt}
\paragraph{Dataset characteristics}
Although the dataset used in this study is relatively small in scale, the three collected sequences represent distinct measurement scenarios with different characteristics. As summarized in Table~\ref{tab:data_summary_main}, the sequences were recorded under different environmental conditions and camera configurations, including glare (Data-1), moderate blur (Data-2), and a more challenging combination of blur and low illumination (Data-3). In addition, the vibration amplitudes vary notably across the datasets (Table~\ref{tab:disp_quantities}). Data-2 exhibits relatively large structural responses, particularly in the vertical direction, whereas Data-3 contains much smaller lateral displacements, representing a low-amplitude and noise-sensitive scenario. Data-1 lies between these two cases and includes glare-induced tracking disturbances. Therefore, despite its limited size, the dataset covers several representative practical conditions, ranging from favorable high-amplitude measurements to challenging low-visibility and low-motion cases, providing a meaningful basis for evaluating the robustness of vision-based displacement measurement methods.

\vspace{4pt}
\paragraph{Sensitivity to motion amplitude}
The proposed framework shows lower sensitivity in high-amplitude vibration scenarios. In the field experiments, the bridge exhibits notably larger vibration in the vertical (Y) direction than in the lateral (X) direction. Under the current side-view camera setup, this difference is reflected in the image domain as larger motion along the vertical image direction and smaller motion along the horizontal image direction. When the image motion is sufficiently large, the tracked trajectories remain clearly distinguishable from pixel-level perturbations, leading to stable triangulation and accurate displacement reconstruction. This behavior is evident in the Y-axis measurements of Data-2 and Data-3, where the estimated peak-to-peak amplitudes closely match the accelerometer-derived references (Table~\ref{tab:disp_quantities}) and the corresponding errors remain very small (Table~\ref{tab:metric_results}).
By contrast, the framework becomes more sensitive in small-amplitude motion scenarios, particularly in the X direction, where the structural displacement is significantly smaller. For example, in Data-3, the lateral amplitude is only 1.54~mm, nearly five times smaller than the vertical amplitude. Under such conditions, the effective signal-to-noise ratio decreases, making point tracking more susceptible to pixel-level perturbations. 
Consequently, small tracking errors can propagate through triangulation and lead to larger displacement deviations, as reflected by higher NRMSE$_{\text{range}}$ and RPPAE values and lower correlation coefficients in the X direction of Data-3 (Table~\ref{tab:metric_results}).

\vspace{4pt}
\paragraph{Sensitivity to video quality}
The proposed framework is also sensitive to video degradation. Blur, low illumination, and glare reduce the reliability of point tracking and consequently degrade the accuracy of 3D reconstruction.
This effect is evident in Data-1 and Data-3, where degraded image quality in one view leads to reduced tracking stability and lower reconstruction accuracy. The impact becomes particularly severe when poor video quality coincides with small-amplitude motion, since the effective signal-to-noise ratio is further reduced. These results suggest that the proposed framework is most reliable when the monitored motion is sufficiently visible in stereo views.

\vspace{4pt}
\paragraph{Effect of structural geometry refinement (SGR)}
The quantitative results in Table~\ref{tab:metric_results} show that SGR tends to provide larger benefits in noise-sensitive scenarios rather than uniformly across all conditions.  
Its contribution is most evident in the lateral (X) direction, where structural motion amplitudes are generally smaller and the resulting horizontal image motion is weaker, making point tracking more susceptible to subpixel localization errors and matching noise.
By enforcing geometric constraints on the reconstructed trajectories, SGR suppresses the propagation of tracking errors during triangulation, leading to lower NRMSE$_{\text{range}}$ and RPPAE and higher correlation coefficients in the X direction across all cases (Table~\ref{tab:metric_results}).
By comparison, the improvement is generally smaller in the vertical (Y) direction, where vibration amplitudes are larger, and the resulting vertical image motion is more pronounced, making the baseline tracking already relatively stable. These observations suggest that SGR mainly acts as a refinement mechanism that mitigates error amplification under low-observability and noise-sensitive conditions.

Overall, the sensitivity analysis indicates that the proposed framework performs most reliably when image motion is sufficiently large and video quality remains high, ensuring stable point tracking and accurate stereo triangulation. Conversely, the estimation becomes more susceptible to feature localization and matching errors under small-amplitude motion combined with degraded video quality, due to the reduced signal-to-noise ratio. This behavior explains the strong performance in high-amplitude vertical vibration cases and the reduced robustness in low-amplitude lateral measurements.

\section{Discussion}
\label{sec:discussion}
This section primarily discusses the reliability of camera parameter estimation using VGGT, and the limitations.

\subsection{Reliability of camera parameter estimation using VGGT}
\begin{figure}[t]
  \centering
  \includegraphics[width=0.7\linewidth]{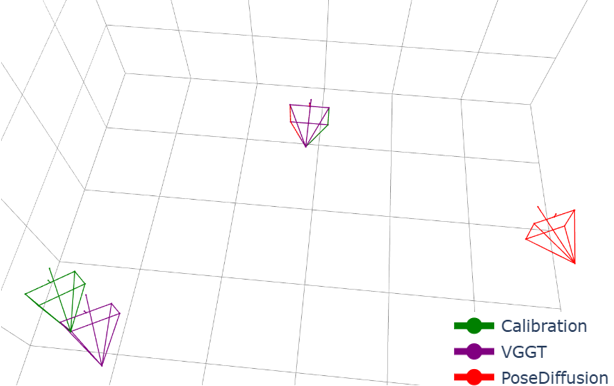}
  \caption{Comparison of camera poses obtained from calibration and estimated by VGGT and PoseDiffusion. The visualization illustrates the relative pose configurations, where translations are shown up to an arbitrary scale to emphasize the alignment between the estimated poses and the calibrated reference. Absolute physical scale is not depicted, as it is independently recovered from the measured baseline.}
  \label{fig: pose}
\end{figure}

For Data-2, a chessboard calibration was performed during data acquisition to obtain reliable reference camera parameters (see \ref{app:calibration}).
Based on the successfully calibrated image pairs, camera poses (rotation and translation) obtained from chessboard calibration (visualized in green) were jointly visualized with the camera poses predicted by VGGT~\cite{wang2025vggt} (visualized in purple) and PoseDiffusion~\cite{wang2023posediffusion} (visualized in red).
As shown in Fig.~\ref{fig: pose}, the camera poses predicted by VGGT exhibit strong consistency with the calibration results, both in terms of camera centers, which reflect translation, and camera orientations, which reflect rotation. By contrast, the poses predicted by PoseDiffusion show noticeable deviations in both position and orientation. This comparison indicates that VGGT is able to provide more accurate camera pose predictions in this challenging real-world scenario, closely matching the calibration-based reference.

We further attempted to estimate camera poses using a traditional SfM pipeline based on COLMAP \cite{sturm2012benchmark, schonberger2016structure} with SuperPoint \cite{detone2018superpoint} and SuperGlue \cite{sarlin2020superglue} features. However, the reconstruction failed, which is consistent with our expectations. SfM-based methods rely heavily on reliable feature detection and matching, and their performance degrades significantly in scenes with blur, defocus, sparse texture, or limited viewpoint diversity, as encountered in our data.

Another critical factor is the role of camera intrinsics. Many classical SfM pipelines and learning-based pose estimation methods either assume known camera intrinsics or treat them as fixed parameters, which limits their ability to reliably infer intrinsics under real-world conditions.
However, accurate camera intrinsics are critical for mapping pixel measurements to camera rays and for recovering physically meaningful scene geometry and metric displacement scale.
In this context, vision foundation models such as VGGT, which are capable of jointly inferring camera intrinsics and extrinsics and exhibit generalization across imaging conditions, are particularly well suited for the proposed framework.

\subsection{Limitations}
Despite the demonstrated effectiveness of the proposed framework, several limitations remain. 

\vspace{4pt}
\paragraph{Manual initialization of the tracked point}
First, the tracked point in the initial frame of one video is currently selected manually. This process is inherently subjective and may not always identify the most informative or stable feature for tracking. Automating this step through feature-quality assessment or learning-based point selection could further improve the robustness and autonomy of the proposed framework.

\vspace{4pt}
\paragraph{Limited dataset scale and diversity}
Second, although the dataset collected in this study is valuable as the first multimodal field dataset for structural displacement measurement, it remains limited in scale and diversity. The data were collected from a single bridge, although under varying conditions, including differences in illumination, video quality, and camera placement. In addition, each sequence has a relatively short duration. Future work will extend data collection to bridges of different types and scales, and increase monitoring duration to capture a broader range of structural responses, enabling more comprehensive benchmarking of vision-based displacement measurement methods.

\vspace{4pt}
\paragraph{Sensitivity to small-amplitude motion and video degradation}
Third, the proposed framework shows limited robustness when monitoring small-amplitude displacements, particularly when video quality is degraded by blur or defocus. Under small structural motions, point tracking becomes more sensitive to feature localization and matching errors, leading to larger tracking errors and reduced triangulation accuracy. This limitation may be mitigated by ensuring accurate camera focusing during data acquisition and by improving point tracking algorithms to better handle small-amplitude motion.

\vspace{4pt}
\paragraph{Dependence on the vision foundation model}
Finally, the overall performance of the system is closely tied to the capabilities of the underlying vision foundation model (VFM). In scenarios where the VFM exhibits limited generalization, the resulting 3D displacement reconstruction may degrade or become unreliable. Nevertheless, the rapid advancement of VFMs suggests that the accuracy and robustness of the proposed framework will continue to improve. As the proposed framework is model-agnostic, it can directly benefit from future state-of-the-art VFMs with stronger tracking and geometric perception, making it a scalable and continuously evolvable solution for 3D structural displacement measurement.

\section{Conclusion}
\label{sec:conclusion}
To address practical deployment challenges in vision-based displacement measurement, we present VFM-SDM, which, to the best of our knowledge, is the first vision foundation model-based framework for task-specific training-free, marker-free, and manual calibration-free structural displacement measurement. 
The main findings and contributions are summarized as follows.

\begin{itemize}
    \item \textbf{Deployment-oriented methodology.} The proposed framework integrates VFM-based visual reasoning with geometry-guided constraints, eliminating the need for artificial markers, cumbersome manual camera calibration, and task-specific model training, thereby enabling a concise and robust pipeline that facilitates practical deployment.
    
    \item \textbf{Field dataset and benchmarking protocol.} To the best of our knowledge, this work presents the first compact field dataset from an in-service pedestrian bridge for structural displacement measurement benchmarking, together with a unified evaluation protocol. The dataset advances the availability of public benchmarks and provides practical evaluation resources for the vision-based structural monitoring community.

    \item \textbf{Field validation and performance.} Evaluation on the field dataset demonstrated robust recovery of structural displacements, with results consistent with accelerometer-derived displacement references. Component-level evaluations further confirm the effectiveness of the proposed framework.
    
    \item \textbf{Key influencing factors.} Displacement estimation accuracy depends on both camera parameter estimation and point tracking performance, the latter being influenced by video quality and structural vibration amplitude. In challenging scenarios, the proposed structural geometry refinement module further stabilizes and regularizes the estimated displacement trajectories.

\end{itemize}

Overall, this work demonstrates that integrating a vision foundation model with geometry-guided structural constraints enables a scalable, deployment-oriented framework for in-situ structural displacement measurement. 
By eliminating structure-specific and task-specific requirements, the proposed framework reduces practical barriers to vision-based structural monitoring and supports data-driven workflows in digital and intelligent monitoring.

Future work will focus on improving the autonomy and robustness of the framework, particularly through automatic tracked-point initialization, improved resilience to small-amplitude motion and video degradation, and validation on larger and more diverse field datasets with longer monitoring durations. 
As the performance of the system is closely tied to the capabilities of the underlying vision foundation model, ongoing advances in foundation models are expected to further improve tracking reliability and 3D geometric perception, thereby enhancing displacement reconstruction accuracy.
The proposed framework therefore facilitates broader deployment in in-situ monitoring of civil infrastructure, including bridges, buildings, towers, and other large-scale structures, providing a flexible and scalable solution for data-driven 3D structural displacement monitoring.

\section*{CRediT authorship contribution statement}
\textbf{Qingyu Xian:} Conceptualization, Data curation, Formal analysis, Investigation, Methodology, Resources, Software, Validation, Visualization, Writing – original draft, Writing – review and editing.
\textbf{Hao Cheng:} Conceptualization, Formal analysis, Supervision, Writing – review and editing.
\textbf{Berend Jan van der Zwaag:} Formal analysis, Funding acquisition, Project administration, Supervision, Writing – review and editing.
\textbf{Rolands Kromanis:} Formal analysis, Data curation, Resources, Writing – review and editing.
\textbf{Ozlem Durmaz Incel:} Formal analysis, Funding acquisition, Project administration, Supervision, Writing – review and editing.

\section*{Declaration of competing interest}
The authors declare that they have no known competing financial interests or personal relationships that could have appeared to influence the work reported in this paper.

\section*{Declaration of Generative AI and AI-assisted technologies in the manuscript preparation process}
During the preparation of this manuscript, the authors used generative AI tools to assist with language editing and improving readability. After using these tools, the authors reviewed and edited the content as needed and take full responsibility for the content of the published article.

\section*{Acknowledgements}
This work is financed by the Dutch Research Council NWO (www.nwo.nl) under the SUBLIME project (KICH1.ST01.20.008) of the NWO research programme KIC. It is also part of the Partnership Program of the Materials Innovation Institute M2i (www.m2i.nl) with project number N21007c, and partially supported by the Sectorplan Beta-II of the Netherlands.

\section*{Data availability}
The data and code that support the findings of this study are available from the corresponding author upon reasonable request.

\bibliographystyle{elsarticle-num-names} 
\bibliography{references}
 
\appendix
\section{Additional smartphone-based vibration sequence}
\label{app:phone}
\begin{figure}[t]
\centering
\begin{subfigure}{\textwidth}
    \centering
    \includegraphics[width=0.48\textwidth]{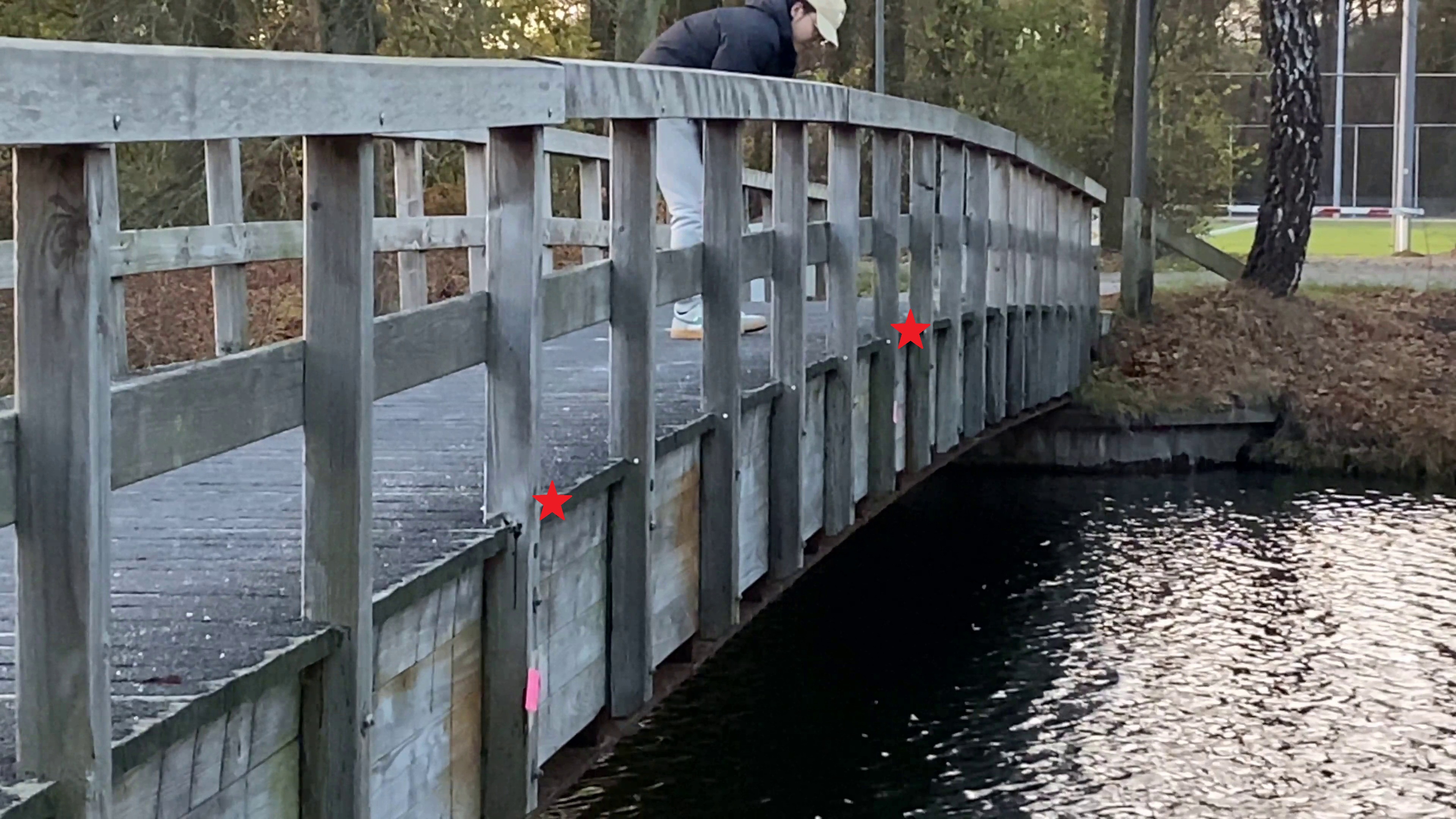}%
    \hfill
    \includegraphics[width=0.48\textwidth]{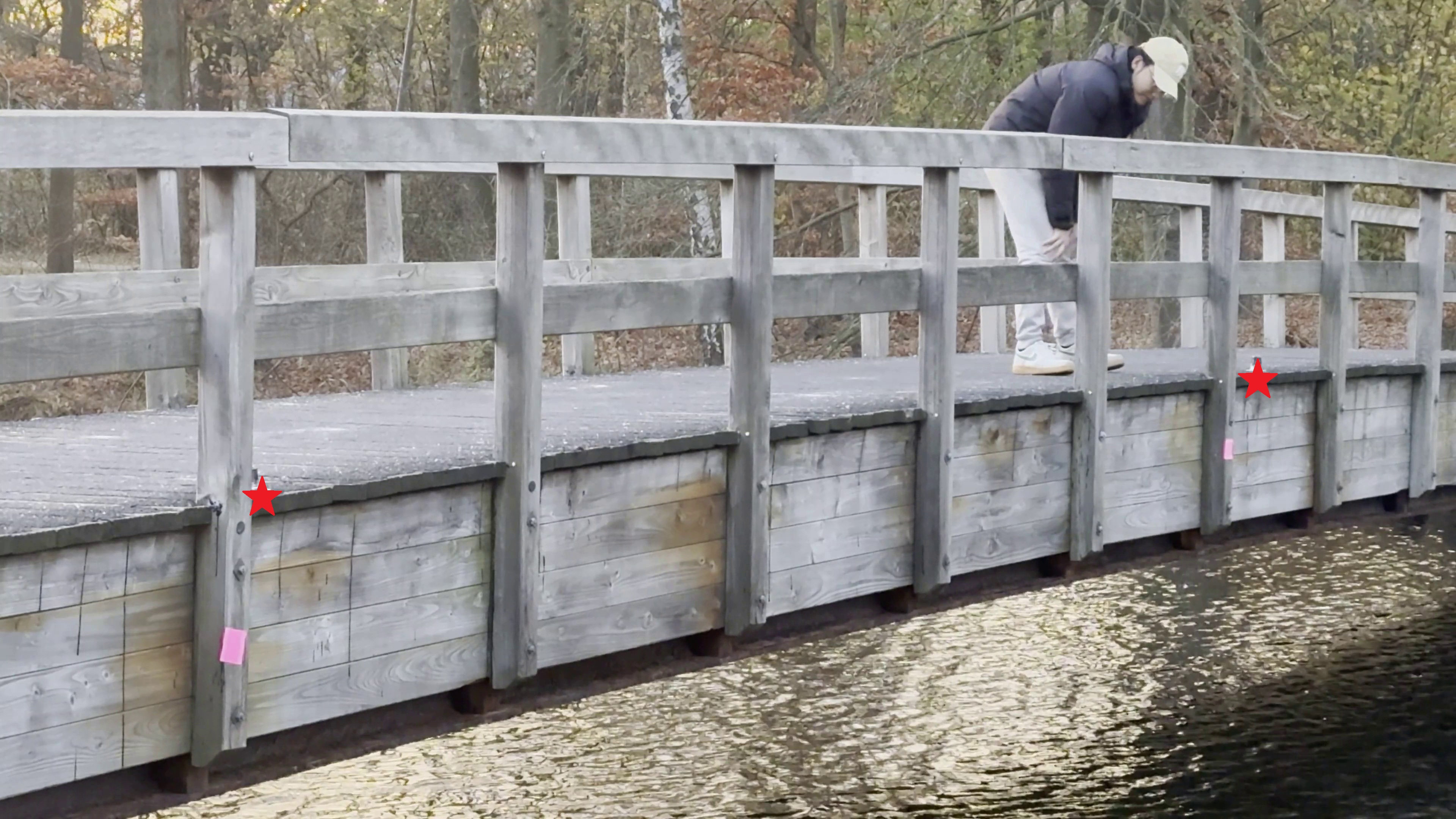}
    \label{fig:data4_stereo}
\end{subfigure}
\caption{Overview of the smartphone-based stereo sequence (Data-4).
The two images show the first frames from the left (camera~1) and right (camera~2) views.
The sticky notes placed on the bridge are used solely to expedite setup by facilitating localization of the monitored region within the FoV, rather than for tracking.
The red stars indicate the locations of the accelerometers.}
\label{fig:dataset_stereo_frames_data4}
\end{figure}

This appendix describes an additional smartphone-based vibration sequence (Data-4), collected using an iPhone~11 and an iPhone~15 Pro Max(see Table~\ref{tab:data_summary_data4} and Fig.~\ref{fig:dataset_stereo_frames_data4}).
It summarizes the characteristics of this vibration sequence, explains its incompatibility with the assumptions of the proposed framework, and outlines its applicability to alternative displacement-measurement approaches.

Due to the characteristics of consumer-grade imaging pipelines, smartphone-based data pose specific challenges to the proposed framework.
In particular, consumer-grade smartphones apply aggressive image stabilization and image-enhancement processes, which violate the brightness-constancy and rigid-projection assumptions required for pixel-based point tracking.
As a result, the 2D trajectories extracted from smartphone videos do not yield physically consistent displacement time series and are therefore excluded from the quantitative and qualitative evaluation of the proposed framework.

\begin{table}[b]
\centering
\fontsize{8pt}{8pt}\selectfont
\caption{Summary of the smartphone-based vibration sequence (Data-4) collected on the same pedestrian bridge.
This vibration sequence is not evaluated under the proposed framework, as it does not conform to its underlying assumptions, but is provided as supplementary material for alternative displacement-measurement approaches.}
\label{tab:data_summary_data4}
\resizebox{\columnwidth}{!}{
\begin{tabular}{lcccc>{\raggedright\arraybackslash}m{4.6cm}}
\toprule
\textbf{Dataset} & \textbf{Date} & \textbf{Monitored Points} & \textbf{Devices} & \textbf{Baseline (cm)} & \textbf{Environmental Condition} \\
\midrule
Data-4 & 2025-11-22 & 1/4 span, 1/2 span & 2 iPhones & 351 & Cloudy; clear imaging conditions \\
\bottomrule
\end{tabular}
}
\end{table}

Nevertheless, we release this smartphone subset (Data-4) as an integral component of the proposed dataset, as it provides synchronized stereo videos and an accelerometer-derived displacement reference under a realistic phone-grade imaging pipeline.
While it is incompatible with the assumptions of the proposed framework, Data-4 remains suitable for validating alternative displacement-measurement paradigms that do not rely on sparse point tracking, such as depth-based 3D reconstruction, dense motion estimation, or learning-based fusion of video and inertial data.

\section{Calibration Procedure}
\label{app:calibration}
\begin{figure}[H]
    \centering
    \begin{subfigure}[t]{0.9\linewidth}
        \centering
        \includegraphics[width=\linewidth]{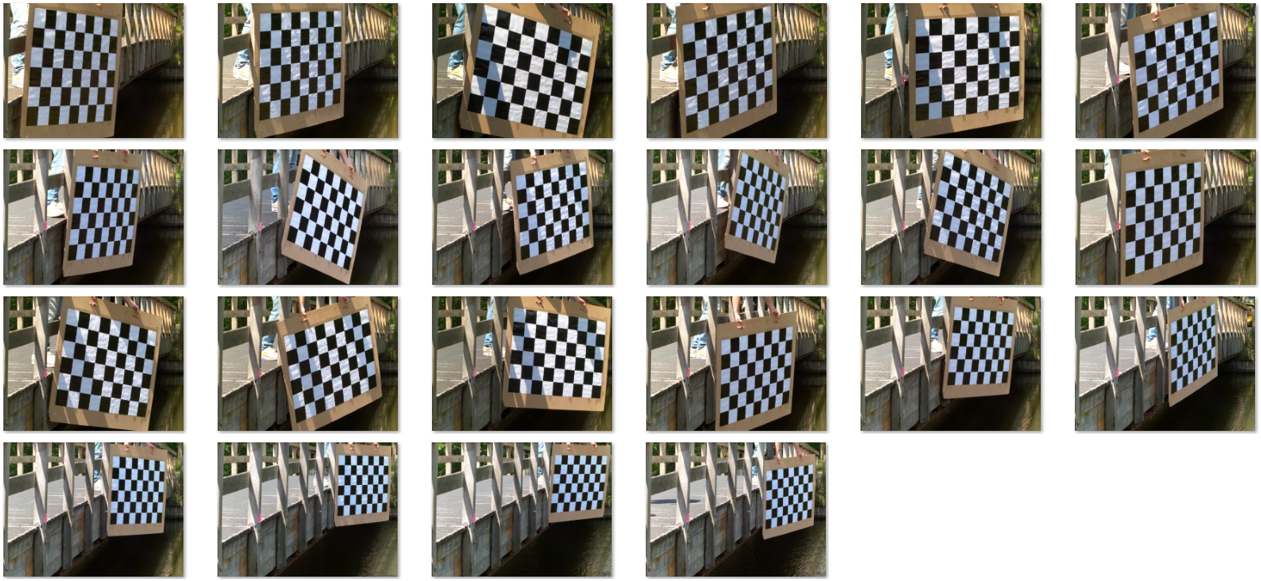}
        \caption{Images captured by Camera~1 during chessboard-based calibration.}
        \label{fig:calib_left}
    \end{subfigure}


    \begin{subfigure}[t]{0.9\linewidth}
        \centering
        \includegraphics[width=\linewidth]{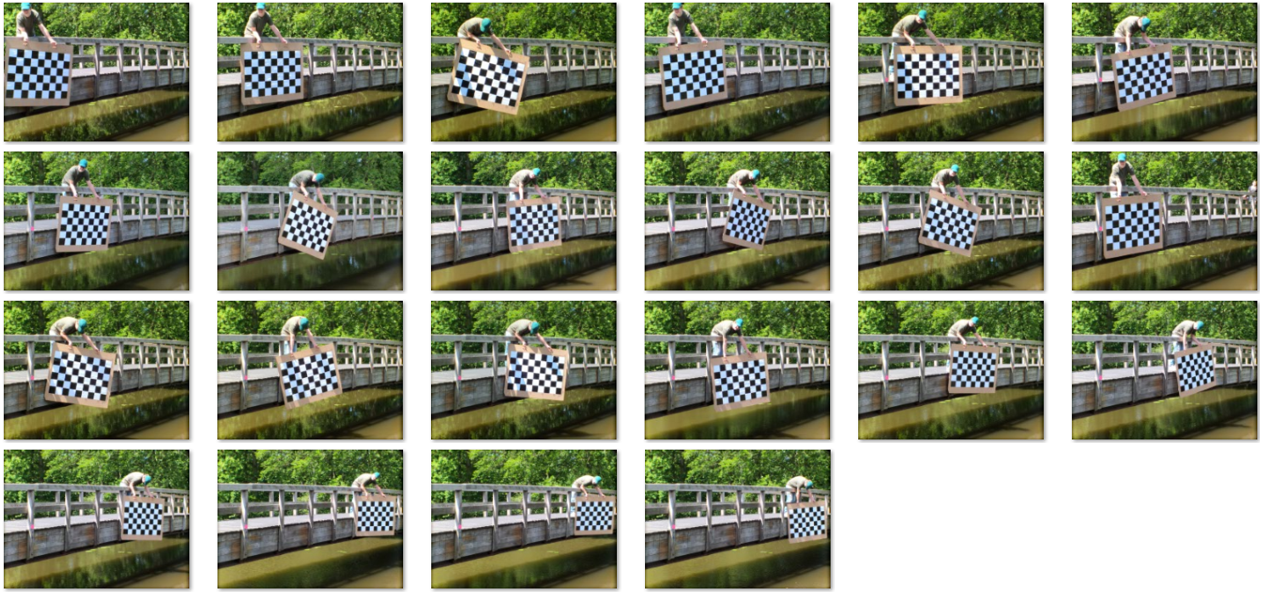}
        \caption{Images captured by Camera~2 during chessboard-based calibration.}
        \label{fig:calib_right}
    \end{subfigure}

    \caption{
    Image pairs captured during chessboard-based camera calibration.}
    \label{fig:calib}
\end{figure}

This appendix details the camera calibration procedure employed during the acquisition of Data-2.
A chessboard-based calibration was conducted to obtain reference camera parameters prior to data collection.
Specifically, a rigid $6 \times 9$ chessboard with a square size of $15\,\mathrm{cm} \times 15\,\mathrm{cm}$ was used, and a total of 22 synchronized image pairs were captured.
During acquisition, the chessboard was deliberately positioned at diverse locations and orientations to ensure adequate coverage of the camera field of view, as illustrated in Fig.~\ref{fig:calib}.

The calibration parameters were estimated using a standard OpenCV-based implementation, and only 12 image pairs yielded sufficiently reliable corner detections for successful calibration during the estimation process.
The remaining image pairs were excluded due to degraded image quality caused by outdoor acquisition conditions, including specular reflections and defocus.
This outcome reflects the practical challenges associated with performing chessboard-based calibration in uncontrolled outdoor environments.

\end{document}